\documentclass{article}

\usepackage[final]{neurips_2024}

\usepackage[utf8]{inputenc} % allow utf-8 input
\usepackage[T1]{fontenc}    % use 8-bit T1 fonts
\usepackage{hyperref}       % hyperlinks
\usepackage{array}
\usepackage{float}
\usepackage{url}            % simple URL typesetting
\usepackage{booktabs}       % professional-quality tables
\usepackage{amsfonts}       % blackboard math symbols
\usepackage{nicefrac}       % compact symbols for 1/2, etc.
\usepackage{microtype}      % microtypography
\usepackage{xcolor}         % colors
\usepackage{graphicx}

\usepackage{xparse}
\usepackage{xspace}
\usepackage{fixltx2e}
\usepackage{amsmath}
\usepackage{amsthm}
\usepackage{tikz}
\usepackage{float}
\usepackage{multirow}
\usepackage{multicol}
\usepackage{colortbl}
\usepackage[textsize=tiny]{todonotes}
\usepackage{mathtools} %\usepackage{dcases}

\usepackage{microtype}
\usepackage{subcaption}
\usepackage{algorithm}
\usepackage[noend]{algpseudocode}
\usepackage{graphicx}
\usepackage{caption}
\usepackage{booktabs} % for professional tables
\usepackage{tabularx}
\usepackage{bbm}
\usepackage[shortlabels]{enumitem}

\usepackage{hyperref,amssymb,enumitem}

\usepackage{wrapfig}

\usepackage{cleveref} % for clever references
\usepackage{arydshln} % for dashed rules in tables https://latex.org/forum/viewtopic.php?t=6175

\usepackage{lscape}

\setlist[itemize]{leftmargin=*}
\setlist[enumerate]{leftmargin=*}

\newcommand\xintu{0.25}
\newcommand\vskipintu{-0.4in}

\renewcommand{\algorithmicrequire}{\textbf{Input: }}

\renewcommand{\algorithmicensure}{\textbf{Output: }}

\newcommand{\ie}{\textit{i.e.,}\@\xspace}
\newcommand{\eg}{\textit{e.g.,}\@\xspace}

\newcommand{\Aug}{\text{Aug}}

\newcommand{\alignexp}[2]%{\mathbb{E}_{{#2}', {#2}'' \sim \Aug(x) } d({#1}({#2}'), {#1}({#2}'')) }
{\underset{ {#2}', {#2}'' \sim \Aug(x) }{\mathbb{E}} [ d\left({#1}({#2}'), {#1}({#2}'')\right) ] }

\newcommand{\alignexpp}[2]%{\mathbb{E}_{{#2}', {#2}'' \sim \Aug(x) } d({#1}({#2}'), {#1}({#2}'')) }
{\underset{ {#2}', {#2}'' \sim \Aug(x) }{\mathbb{E}} d\left({#1}({#2}'), {#1}({#2}'')\right)}

\newcommand{\alignexpone}[2]%{\mathbb{E}_{{#2}', {#2}'' \sim \Aug(x) } d({#1}({#2}'), {#1}({#2}'')) }
{\underset{ {#2} \sim \Aug(x) }{\mathbb{E}} {#1}\left({#2}\right)}

\newcommand{\alignloss}[1]{\mathcal{L}_{\mathcal{A}}(#1,x)}

\newcommand{\expf}{ \underset{f \sim \mathcal{M} \left(\mathcal{D}\right)}{\mathbb{E}} }
\newcommand{\expg}{ \underset{g \sim \mathcal{M}(\mathcal{D} \setminus x)}{\mathbb{E}} }

\newcommand{\sslfff}{\expf\ \alignexpp{f}{x} }
\newcommand{\sslggg}{\expg\ \alignexpp{g}{x} }

\newcommand{\SSLMem}{\texttt{SSLMem}\xspace}
\newcommand{\sslmem}{\texttt{SSLMem}\xspace}

\newcommand{\ssl}{\texttt{SSL}\xspace}

\newcommand{\unitmem}{\texttt{UnitMem}\xspace}
\newcommand{\classmem}{\texttt{ClassMem}\xspace}
\newcommand{\classselectivity}{\texttt{ClassSelectivity}\xspace}
\newcommand{\layermem}{\texttt{LayerMem}\xspace}
\newcommand{\deltamem}{$\Delta$\layermem}
\newcommand{\blockmem}{\texttt{BlockMem}\xspace}
\newcommand{\deltavit}{$\Delta$\blockmem}
\newcommand{\residual}{\texttt{ResBlock}\xspace}

\usepackage{amsmath}

\graphicspath{{images/}}

\newif\ifdraft
\draftfalse

\usepackage{fancyvrb}

\RecustomVerbatimCommand{\VerbatimInput}{VerbatimInput}%
{fontsize=\footnotesize,
 frame=lines,  % top and bottom rule only
 framesep=2em, % separation between frame and text
 commandchars=\|\(\), % escape character and argument delimiters for
 commentchar=*        % comment character
}

\newcommand{\reb}[1]{\textcolor{black}{#1}}

\ifdraft
\newcommand{\franzi}[1]{\textcolor{purple}{[Franzi: #1]}}
\newcommand{\adam}[1]{\textcolor{blue}{[Adam: #1]}}
\newcommand{\wenhao}[1]{\textcolor{green}{[Wenhao: #1]}}

\else
\newcommand{\franzi}[1]{}
\newcommand{\wenhao}[1]{}
\newcommand{\adam}[1]{}
\fi

\title{Localizing Memorization in SSL Vision Encoders}

\author{Wenhao Wang$^{1}$, 
Adam Dziedzic$^{1}$,
\textbf{Michael Backes}$^{1}$, \textbf{Franziska Boenisch}$^{1}$\thanks{Correspondence to boenisch@cispa.de}\\
$^{1}$CISPA, Helmholtz Center for Information Security
}

\begin{document}

\maketitle

\begin{abstract}
\label{sec:abstract}
Recent work on studying memorization in self-supervised learning (SSL) suggests that even though SSL encoders
are trained on millions of images, they still memorize individual data points. While effort has been put into characterizing the memorized data and linking encoder memorization to downstream utility, little is known about where the memorization happens inside SSL encoders. To close this gap, we propose two metrics for localizing memorization in SSL encoders on a per-layer (\layermem) and per-unit basis (\unitmem). Our localization methods are independent of the downstream task, do not require any label information, and can be performed in a forward pass. By localizing memorization in various encoder architectures (convolutional and transformer-based) trained on diverse datasets with contrastive and non-contrastive SSL frameworks, we find that (1)~while SSL memorization increases with layer depth, highly memorizing units are distributed across the entire encoder, (2)~a significant fraction of units in SSL encoders experiences surprisingly high memorization of individual data points, which is in contrast to models trained under supervision, (3)~\textit{atypical} (or outlier) data points cause much higher layer and unit memorization than standard data points, and (4)~in vision transformers, most memorization happens in the fully-connected layers. Finally, we show that localizing memorization in SSL has the potential to improve fine-tuning and to inform pruning strategies.
\end{abstract}
\section{Introduction}\label{sec:intro}

Self-supervised learning (SSL)~(\citep{chen2020simple,simsiam,caron2021dino,bardes2022vicreg,mae,grill2020bootstrap,he2020momentum}) enables pre-training large encoders on unlabeled data to generate feature representations for a multitude of downstream tasks. 
Recently, it was found that, even though their training datasets are large, SSL encoders still memorize individual data points~(\citep{meehan2023ssl,wang2024memorization}).
While prior work characterizes the memorized data and studies the effect of memorization to improve downstream generalization~(\citep{wang2024memorization}), little is known about where in SSL encoders memorization happens.

The few works on localizing memorization are usually confined to supervised learning (SL)~(\citep{baldock2021deep,stephenson2020geometry,maini2023can}), or operate in the language domain~(\citep{zhu2020modifying, meng2022locating,bills2023language,singh2023explaining,chang2023localization}). 
In particular, most results are coarse-grained and localize memorization on a per-layer basis~\citep{baldock2021deep,stephenson2020geometry} and/or require labels~\citep{baldock2021deep,maini2023can}. 

To close the gap, we propose two novel metrics for localizing memorization in SSL encoders in the vision domain. 
Our \layermem localizes memorization of the training data within the SSL encoders on a layer-level.
For a more fine-grained localization, we turn to memorization in individual \textit{units} (\ie neurons in fully-connected layers or channels in convolutional layers). We propose \unitmem which measures memorization of individual training data points through the units' sensitivity to these points.
Both our metrics can be computed independently of a downstream task, in a forward pass without gradient calculation, 
and without labels, which makes them computationally efficient and well-suited for the large SSL encoders pretrained on unlabeled data.
By performing a systematic study on localizing memorization with our two metrics on various encoder architectures (convolutional and transformer-based) trained on diverse vision datasets with contrastive and non-contrastive SSL frameworks, we make the following key discoveries:

\textbf{Memorization happens through the entire SSL encoder.} 
By analyzing our \layermem scores between subsequent layers, we find that the highest memorizing layers in SSL are not necessarily the last ones, which is in line with findings recently reported for SL~\citep{maini2023can}.
While there is a tendency that higher per-layer memorization can be observed in deeper layers, similar to SL~\citep{baldock2021deep,stephenson2020geometry}, our analysis of memorization on a per-unit level highlights that highly memorizing units are distributed across the entire SSL encoder, and can also be found in the first layers.

\textbf{Units in SSL encoders experience high memorization.}
By analyzing SSL encoders with our \unitmem metric, we find that a significant fraction of their units experiences high memorization of individual training data points.
This stands in contrast with models trained using SL for which we observe
high class memorization, measured as the unit's sensitivity to any particular class.
While these results are in line with the two learning paradigms' objectives where SL optimizes to separate different classes whereas SSL optimizes foremost for instance discrimination~\citep{wang2021chaos}, it is a novel discovery that this yields significantly different memorization patterns between SL and SSL down to the level of individual units.

\textbf{Atypical data points cause higher memorization in layers and units.} 
While prior work has shown that SSL encoders overall memorize atypical data points more than standard data points~\citep{wang2024memorization}, our study reveals that the effect is constant throughout \textit{all} encoder layers.
Hence, there are no particular layers responsible for memorizing atypical data points, similarly as observed in SL~\citep{maini2023can}.
Yet, memorization of atypical data points can be attributed on a unit-level where we observe that the highest memorizing units align with the highest memorized (atypical) data points and that overall atypical data points cause higher unit memorization than standard data points.

\textbf{Memorization in vision transformers happens mainly in the fully-connected layers.}
The memorization of transformers~\citep{vaswani2017attention} was primarily investigated in the language domain~\citep{geva2021keyvaluememory,sadrtdinov2021memorization}.
However, the understanding in the vision domain is lacking, and due to the difference in input and output tokens (language transformers operate on discrete tokens while vision transformers operate on continuous ones), the methods for analysis and the findings are not easily transferable. 
Yet, with our methods to localize memorization, we are the first to show that the same trend holds in vision transformers that was previously reported for language transformers, namely that memorization happens in the fully-connected layers.

Finally, we investigate future applications that could benefit from localizing memorization and identify \textit{more efficient fine-tuning} and \textit{memorization-informed pruning strategies} as promising directions.

In summary, we make the following contributions:
\begin{itemize}
    \item We propose \layermem and \unitmem, the first practical metrics to localize memorization in SSL encoders on a per-layer basis and down to the granularity of individual~units.
    \item We perform an extensive experimental evaluation to localize memorization in various encoder architectures trained on diverse vision datasets with different SSL~frameworks.
    \item Through our metrics, we gain new insights 
    into the memorization patterns of SSL encoders and can compare them to the ones of~SL models.
    \item We show that the localization of memorization can yield practical benefits for encoder fine-tuning and pruning.   
\end{itemize}

\section{Related Work}

\paragraph{SSL.}
SSL relies on large amounts of unlabeled data to train encoder models that return representations for a multitude of downstream tasks~\citep{ssl-learning2013}.
Especially in the vision domain, a wide range of SSL frameworks have recently been introduced~\citep{chen2020simple,simsiam,caron2021dino,bardes2022vicreg,mae,grill2020bootstrap,he2020momentum}.
Some of them rely on contrastive loss functions~\citep{chen2020simple,he2020momentum,grill2020bootstrap} whereas others train with non-contrastive objective functions~\citep{pokle22aNonContrastive,simsiam,caron2021dino,mae}.

\paragraph{Memorization in SL.}
Memorization was extensively studied in SL~\citep{zhang2016understanding,arpit2017closer,chatterjee2018learning}. 
In particular, it was shown that it can have a detrimental effect on data privacy, since it enables data extraction attacks~\citep{carlini2019secret, carlini2021extracting, carlini2022privacy}.
At the same time, memorization seems to be required for generalization, in particular for long-tailed data distributions~\citep{feldman2020does, feldman2020neural}.
It was also shown that \textit{harder} or more atypical data points~\citep{arpit2017closer,sadrtdinov2021memorization} experience higher memorization.
While all these works focus on studying memorization from the data perspective and concerning its impact on the learning algorithm, they do not consider where memorization happens.

\paragraph{Memorization in SSL.}
Even though SSL rapidly grew in popularity during recent years, work on studying memorization in SSL is limited.
\citet{meehan2023ssl} proposed to quantify Déjà Vu memorization of SSL encoders with respect to particular data points by comparing the representations of these data points with the representations of a labeled public dataset. 
Data points whose $k$ nearest public neighbors in the representation space are highly consistent in labels are considered to be memorized. 
Since SSL is aimed to train \textit{without labels}, this approach is limited in practical applicability.
More recently,~\citet{wang2024memorization} proposed \sslmem, a definition of memorization for SSL based on the leave-one-out definition from SL~\citep{feldman2020does, feldman2020neural}.
Instead of relying on labels, this definition captures memorization through representation alignment, \ie measuring the distance between representations of a data point's multiple augmentations. 
Since both works rely on the output representations to quantify memorization, neither of them is suitable for performing fine-grained localization of memorization. 
Yet, we use the setup of \sslmem as a building block to design our \layermem metric which localizes memorization per layer.

\paragraph{Localizing Memorization.}
In SL, most work focuses on localizing memorization on a per-layer basis and suggests that memorization happens in the deeper layers~\citep{baldock2021deep,stephenson2020geometry}.
By analyzing which neurons have the biggest impact on \textit{predicting the correct label} of a data point, \citet{maini2023can} were able to study memorization on a per-unit granularity. 
They do so by zero-ing out random units until a label flip occurs.
Their findings suggest that only a few units are responsible for memorizing outlier data points.
Yet, due to the absence of labels in SSL, this approach is inapplicable to our work. 
In the language domain, a significant line of work aims at localizing where semantic facts are stored within large language transformers~\citep{zhu2020modifying, meng2022locating,bills2023language,singh2023explaining}.
\citet{chang2023localization} even proposed \textit{benchmarks} for localization methods in the language domain. In the injection benchmark (INJ Benchmark), they fine-tune a small number of neurons and then assess whether the localization method detects the memorization in these neurons. The deletion benchmark (DEL Benchmark) first performs localization, followed by the deletion of the responsible neurons, and a final assessment of the performance drop on the data points detected as memorized in the identified neurons. 
Since in SSL, performance drop cannot be measured directly due to the absence of a downstream task, the deletion approach is not applicable.
Instead, we verify our \unitmem metric in a similar vein to the INJ Benchmark by fine-tuning a single unit on a data point and localizing memorization as we describe in \Cref{sub:unit_verification}.

\paragraph{Studying Individual Units in ML Models.}
Early work in SL~\citep{erhan2009visualizing} already suggested that units at different model layers fulfill different functions: while units in lower layers are responsible for extracting general features, units in higher layers towards the model output are responsible for very specific features~\citep{zeiler2014visualizing}.
In particular, it was found that units represent different concepts required for the primary task \citep{bau2017network}, where some units focus on single concepts whilst others are responsible for multiple concepts~\citep{morcos2018importance,zhou2018revisiting}.
While these differences have been identified between the units of models trained with SL, we perform a corresponding investigation in the SSL domain through the lens of localizing memorization.

\section{Background and Setup}

\paragraph{SSL and Notation.}
We consider an SSL training framework $\mathcal{M}$. The encoder $f:\mathbb{R}^n \to \mathbb{R}^s$ is pre-trained, or in short \textit{trained}, on the unlabeled dataset $\mathcal{D}$ to output representations of dimensionality $s$. Throughout the training, as the encoder improves, its alignment loss 
$\alignloss{f} = d(f(x'),f(x''))$ 
between the representations of two random augmentations  $x',x''$ of any training data point $x$ decreases with respect to a distance metric $d$ (\eg Euclidean distance). \reb{This effect has also been observed in non-contrastive SSL frameworks~\citep{zhang2022mask}}. 
We denote by $f^l, l \in [1,\dots L] $ the $l^{th}$ layer of encoder $f$. Data points from the test set $\mathcal{\bar{D}}$ are denoted as $\bar{x}$.

\paragraph{Memorized Data.} Prior work in the SL domain usually generates outliers for measuring memorization by flipping the labels of training data points~\citep{feldman2020does,feldman2020neural,maini2023can}. 
This turns these points into outliers that experience a higher level of memorization and leave the strongest possible signal in the model. Yet, such an approach is not suitable in SSL where labels are unavailable.
Therefore, we rely on the \sslmem metric proposed by \citep{wang2024memorization} to identify the most (least) memorized data points for a given encoder.
The findings based on the \sslmem metric indicate that the most memorized data points correspond to atypical and outlier samples.

\paragraph{\sslmem for Quantifying Memorization.}
\sslmem quantifies the memorization of individual data points by SSL encoders. It is, to the best of our knowledge, the only existing method for quantifying memorization in SSL without reliance on downstream labels.
\sslmem for a training data point $x$ is defined as 

\begin{align}
    \ssl_f(x) = \sslfff; &\ \ssl_g(x) = \sslggg \nonumber \\
    \sslmem_{f,g}(x) =& \ \ssl_g(x) - \ssl_f(x) \label{eq:memdefold}
\end{align}

where $f$ and $g$ are two classes of SSL encoders whose training dataset $\mathcal{D}$ differs in data point $x$. $x'$ and $x''$ denote two augmentations randomly drawn from the augmentation set \textit{Aug} that is used during training and $d$ is a distance metric, here $\ell_2$-distance.

\paragraph{Experimental Setup.}
We localize memorization in encoders trained with different SSL frameworks on five common vision datasets, namely CIFAR10, CIFAR100, SVHN, STL10, and ImageNet. We leverage different model architectures from the ResNet family, including ResNet9, ResNet18, ResNet34, and ResNet50. 
We also analyze Vision Transformers (ViTs) using their Tiny and Base versions. Results are reported over three independent trials. 
To identify the most memorized training data points, we rely on the \sslmem metric and follow the setup from~\citep{wang2024memorization}. More details on the experimental setup can be found in \Cref{apendix:experimental-setup}.\footnote{Our code is attached as supplementary material.}
\reb{For the readers' convenience, we include a glossary with short explanations for all concepts and background relevant to this work in \Cref{app:glossary}.}
\section{Layer-Level Localization of Memorization}
\label{sec:per_layer}
In order to localize memorization on a per-layer granularity, we propose a new \layermem metric which relies on the \SSLMem metric, as a building block.
Since the \SSLMem as defined in \Cref{eq:memdefold} is not normalized, we introduce the following normalization to the range $[0,1]$
\begin{equation}
    \sslmem'_{f,g}(x) = \frac{\ssl_g(x) - \ssl_f(x)}{\ssl_f(x) + \ssl_g(x)} \label{eq:memdef}
\end{equation}
such that values close to $0$ denote no memorization while $1$ denotes the highest memorization. This makes the score more interpretable.
While \sslmem returns a memorization score per data point for a given encoder,
\layermem returns a memorization score per encoder layer $l$, measured on a (sub)set $\mathcal{D'} = \{x_1,...,x_{|\mathcal{D'}|}\} \subseteq \mathcal{D}$ of training data $\mathcal{D}$.
Similar to \SSLMem, \layermem makes use of a second encoder $g$ as a reference to detect memorization as
\begin{equation}\label{eq:layermem}
\begin{split}
    \layermem_{\mathcal{D'}}(l) =  \frac{1}{|\mathcal{D'}|}\sum_{i=1}^{|\mathcal{D'}|} \sslmem'_{f^l,g^l}(x_i)\text{.} 
    \end{split}
\end{equation}
$f^l,g^l$ denote the output of encoders $f$ and $g$ after layer $l$, respectively.
Intuitively, our \layermem metric measures the average per-layer memorization over training data points $x_i\in\mathcal{D'}$.
\reb{As our \layermem build on \sslmem$'$, it also inherits the above normalization.} 
Since \Cref{eq:layermem} operates on different layers' outputs which in turn depend on all previous layers, \layermem risks to report accumulated memorization up to layer $l$.
Therefore, we also define $\Delta$ \layermem$_{\mathcal{D'}}(l)$ for all layers $l>1$ as
\begin{equation}\label{eq:deltalayermem}
    \begin{split}
    \Delta\layermem_{\mathcal{D'}}(l) = \layermem_{\mathcal{D'}}(l) - \layermem_{\mathcal{D'}}(l-1).
    \end{split}
\end{equation}
This reports the increase in memorization of layer $l$ with respect to the previous layer $l-1$.
\subsection{Experimental Results and Observations}
\reb{We present our core results and provide additional ablations on our \layermem in \Cref{appendix:layermem-insigths}.}

\begin{wraptable}{r}{0.55\textwidth}
\addtolength{\tabcolsep}{-3.0pt} 
    \centering
    \tiny
        \caption{
    \textbf{Layer-based Memorization Scores.} 
    Res$N$ denotes a residual connection that comes from the previous $N$-th convolutional layer. 
    }
    \label{tab:resnet9_per_layer_our_memorization_score}
    \begin{tabular}{cccccc}
    \toprule 
    Layer & \layermem & $\Delta$\texttt{LM} & \layermem Top50 & $\Delta$\texttt{LM} Top50 & \layermem Least50 \\
    \midrule
         1 &0.091&-&0.144&-&0.003\\
         2 & 0.123&0.032&0.225&0.081&0.012\\
         3 & 0.154&0.031&0.308&0.083&0.022\\
         4 &0.183&0.029&0.402&0.094&0.031 \\
         Res$2$ &0.185&0.002&0.403&0.001&0.041 \\
         5 &0.212 &0.027& 0.479& 0.076 &0.051\\
         6 &0.246&0.034&0.599& 0.120 &0.061\\
         7 & 0.276&0.030&0.697& 0.098 &0.071\\
         8 &0.308&0.032&0.817&0.120 &0.073\\
         Res$6$ &0.311& 0.003& 0.817& 0 &0.086\\
        \bottomrule
    \end{tabular}
\end{wraptable}
\addtolength{\tabcolsep}{3.0pt} 

\paragraph{Memorization Increases but not Monotonically.}
We report the \layermem scores in \Cref{tab:resnet9_per_layer_our_memorization_score} for the ResNet9-based SSL encoder trained with SimCLR on CIFAR10 (further per-layer breakdown and scores for ResNet18, ResNet34, and ResNet50 are presented \reb{in \Cref{tab:resnet18_per_layer_our_memorization_score}, \Cref{tab:resnet34_per_layer_our_memorization_score}, and \Cref{tab:resnet50_per_layer_our_memorization_score}} in \Cref{appendix:per-layer-memorization-resnets}).
We report \layermem across the 100 randomly chosen training data points, their \deltamem (denoted as $\Delta$\texttt{LM}), followed by \layermem for only the Top 50 memorized data points, their \deltamem (denoted as $\Delta$\texttt{LM} Top50), and \layermem for only the Least 50 memorized data points.
The results show that our \layermem indeed increases with layer depth in SSL, similar to the trend observed for SL~\citep{stephenson2020geometry}, \ie deeper layers experience higher memorization than early layers.
However, our \deltamem presents the memorization from a more accurate perspective, where we discard the accumulated memorization from previous layers, including the residual connections. \deltamem indicates that the memorization increases in all the layers but is not monotonic.

We also study the differences in localization of the memorization for most memorized (outliers and atypical examples) vs. least memorized data points (inliers), shown as columns \layermem Top50 and \layermem Least50 in \Cref{tab:resnet9_per_layer_our_memorization_score}, respectively. 
While we observe that the absolute memorization for the most memorized data points is significantly higher than for the least memorized data points, they both follow the same trend of increasing memorization in deeper layers.
The \deltamem for the most memorized points (denoted as $\Delta$\texttt{LM} Top50 in \Cref{tab:resnet9_per_layer_our_memorization_score}) indicates that, following the overall trend, high memorization of the atypical samples is also spread over the entire encoder and not confined to particular layers.

\begin{wraptable}{r}{0.35\textwidth}
\vspace{-0.3cm}
\addtolength{\tabcolsep}{0pt}
    \centering
    \tiny
        \caption{\textbf{Memorization in ViT occurs primarily in the deeper blocks and more in the fully connected than attention layers.} 
    }
    \label{tab:vit-layermem}
    \begin{tabular}{ccc}
    \toprule 
    ViT Block & \layermem & \deltamem \\
    & \multicolumn{2}{c}{\textbf{\textit{Attention Layer}}}  \\ \cline{2-3}
    2 & 0.028& 0.008  \\
    6 & 0.114& 0.009 \\
    12 & 0.281 & 0.010  \\
    \hdashline
    & \multicolumn{2}{c}{\textbf{\textit{Fully-Connected Layer}}}  \\ \cline{2-3}
    2 & 0.039 &0.011 \\
    6 & 0.129 & 0.015 \\
    12 & 0.303 &0.022 \\
    \bottomrule
    \end{tabular}
\vspace{-0.4cm}
\addtolength{\tabcolsep}{0pt}
\end{wraptable}
\paragraph{Memorization in Vision Transformers.}
The memorization of Transformers~\citep{vaswani2017attention} was, so far, primarily investigated in the language domain~\citep{geva2021keyvaluememory,sadrtdinov2021memorization}, however, its understanding in the vision domain is lacking. 
The fully-connected layers in \textit{language transformers} were shown to act as key-value \textit{memories}. Still, findings from language transformers cannot be easily transferred to \textit{vision transformers} (ViTs)~\citep{dosovitskiy2021vit}:
while language transformers operate on the level of \textit{discrete} and interpretable input and output tokens, ViTs operate on \textit{continuous} input image patches and output representations. Through the analysis of our newly proposed metric for memorization in SSL, in \Cref{tab:vit-layermem} (ViT-Tiny trained on CIFAR10 using MAE~\citep{mae}), we are the first to show that memorization in ViTs occurs more in deeper blocks and that within the blocks, \textit{fully-connected layers memorize more than attention layers}.
We present the full set of results for \layermem and \deltamem over \textit{all blocks} in \Cref{tab:vit-layermem-whole}.

\paragraph{Memorization in Different SSL Frameworks.}
We also study the differences in memorization behavior between different SSL frameworks. Therefore, we compare the \layermem score between corresponding layers of a ResNet50 trained on ImageNet with SimCLR~\citep{chen2020simple} and DINO~\citep{caron2021dino}, and of a ViT-Base encoder trained on ImageNet with DINO and MAE~\citep{mae}. 
We ensure by early stopping that the resulting linear probing accuracies of the encoder pairs are similar for better comparability of their memorization.
The ImageNet downstream task performance within both encoder pairs is 66.12\% for SimCLR and 68.44\% for DINO; and 60.43\% for MAE and 60.17\% for DINO.
Our results in \Cref{tab:layermem-fraemworks} show that encoders with the same architecture trained with different SSL frameworks experience a similar memorization pattern, 
\reb{namely that memorization occurs primarily in the deeper blocks/layers.}
In \Cref{fig:comparing_frameworks_unit} in \Cref{app:unitmem}, we additionally show that memorization patterns between different SSL frameworks are similar down to the individual unit level, \reb{\ie the number of highly memorizing units and the magnitude of memorization are roughly the same}.
\reb{We present the full results for \textit{all} ResNet50 layers and \textit{all} ViT blocks in \Cref{tab:resnet50-simclr-dino-full} and \Cref{tab:vit-base-mae-dino-full}, respectively.}

\paragraph{Variability and Consistency of Memorization cross Different Layers.}\label{sec:sonsist}
We use \layermem to analyze the variability and consistency between the samples memorized by different layers in a ResNet9 vision encoder trained with CIFAR10 dataset. The results are shown in \Cref{tab:consistency} and \Cref{fig:consistency} in \cref{app:vi_consist}. The overlap within the 100 most memorized samples between adjacent layers is usually high but decreases the further the layers are separated. Our statistical analysis to compare the similarity of the orderings within different layers’ most memorized samples using the Kendall’s rank correlation coefficient shows that while for closer layers, we manage to reject the null hypothesis (“no correlation”) with high statistical confidence (low p-value) which is not the case for further away layers.

\begin{table}[t]
\small
\centering
\caption{\textbf{Consistency in 100 most memorized samples according to \layermem.} We report the pairwise overlap between the most memorized samples and the consistency in ranking of most memorized samples using the statistical Kendall's Tau test ($\tau$-statistic, $p$-value).
While we observe high overlap and statistical similarity within adjacent layers, especially towards the end of the network, there is low similarity and overlap between early and late layers.
}
\vspace{0.2cm}
\resizebox{\linewidth}{!}{
\begin{tabular}{cccccccccccccccc}
\toprule
Layers & Overlap \% & $\tau,p$ &  &  Layers & Overlap \% & $\tau,p$ &  & Layers & Overlap \% & $\tau,p$ &  & Layers & Overlap \% & $\tau,p$ &  \\
\midrule
\textbf{1 2}  &79  &0.607, 4.07e-29  &  & \textbf{1 3} &52 & 0.505, 9.08e-11&  & \textbf{1 4}  &47 &0.412, 6.07e-7 & & \textbf{1 5} &24 & 0.240, 1.18e-2& \\
\textbf{1 6} & 19 &0.181, 6.01e-2  &  & \textbf{1 7} &18 &0.167, 744e-2 &  &\textbf{1 8}  & 16& 0.104, 1.27e-1& & \textbf{2 3 }&70 & 0.562, 8.48e-19& \\
\textbf{2 4}  & 64 &  0.544, 3.95e-16&  & \textbf{2 5}  & 36& 0.288, 2.08e-4&  & \textbf{2 6}   & 30& 0.249, 3.96e-3& & \textbf{2 7}  & 28&0.241, 9.96e-3 & \\
\textbf{2 8}   & 27 & 0.247, 5.53e-3 &  & \textbf{3 4}  &82 &0.665, 4.41e-42 &  & \textbf{3 5}   & 51& 0.512, 6.67e-11& & \textbf{3 6}  &42 &0.356, 8.31e-5 & \\
\textbf{3 7}  &  39& 0.319, 8.31e-5 &  & \textbf{3 8}  &37 &0.310, 1.09e-4 &  & \textbf{4 5}  &68 & 0.557, 1.61e-18& & \textbf{4 6}  & 54&0.509, 6.31e-11 & \\
\textbf{4 7}  &48  & 0.412, 6.67e-7 &  & \textbf{4 8} &45 &0.396, 2.50e-6 &  & \textbf{5 6}  & 72& 0.559, 4.11e-18& & \textbf{5 7}&61 & 0.531, 4.19e-14& \\
\textbf{5 8}   &58  &  0.527, 1.08e-14&  & \textbf{6 7} & 84&0.657, 1.47e-42 &  & \textbf{6 8}   &79 & 0.644, 4.17e-37& & \textbf{7 8}&94 & 0.837, 9.71e-76& \\
\bottomrule
\end{tabular}
}
\label{tab:consistency}

\end{table}

\begin{wraptable}{r}{0.55\textwidth}
\addtolength{\tabcolsep}{0pt}
\vspace{-0.5cm}
    \centering
    \tiny
        \caption{\textbf{The layer-based memorization is similar across encoders trained with different frameworks.} 
    \texttt{LM}=\layermem, $\Delta$\texttt{LM}=\deltamem.}
    \label{tab:layermem-fraemworks}
    \begin{tabular}{ccccccc}
    \toprule 
    ResNet50 Layer & \multicolumn{2}{c}{\textbf{\textit{SimCLR}}} && \multicolumn{2}{c}{\textbf{\textit{DINO}}} \\ \cline{2-3} \cline{5-6}
    Number & \texttt{LM} & $\Delta$\texttt{LM} && \texttt{LM} & $\Delta$\texttt{LM}  \\[0.0cm]  \midrule
    2 &0.040 &0.003 &&0.041 & 0.002\\
    27 & 0.161&0.005 && 0.165& 0.006\\
    49 & 0.302&0.008 &&0.311 & 0.007\\
    \midrule
    ViT-Base Block & \multicolumn{2}{c}{\textbf{\textit{MAE}}} && \multicolumn{2}{c}{\textbf{\textit{DINO}}} \\ \cline{2-3} \cline{5-6}
    Number & LM & $\Delta$LM && \texttt{LM} & $\Delta$\texttt{LM}  \\[0.0cm]  \midrule
    2 & 0.037&0.010 && 0.036& 0.011\\
    6 &0.120 &0.015 && 0.116& 0.014\\
    12 &0.274 &0.019 && 0.271& 0.019\\
    \bottomrule
    \end{tabular}
\addtolength{\tabcolsep}{0pt}
\end{wraptable}

\subsection{Verification of Layer-Based Memorization}
To analyze whether our \layermem metric and its $\Delta$ variant indeed localize memorization correctly, we first replace different layers of an encoder and then compute linear probing accuracy on various downstream tasks.
Since prior work shows that memorization in SSL is required for downstream generalization~\citep{wang2024memorization}, we expect the highest performance drop when replacing the layers identified as most memorizing.

Our results in \Cref{app:layermemverification} verify this intuition.
They show that by replacing \textit{the most memorizing layers} of an encoder trained on a dataset $A$, \eg CIFAR10, with the equivalent layers of another dataset $B$, \eg STL10, the linear probing accuracy drop for CIFAR10 is significantly larger than when when \textit{replacing random or least memorizing layers}.
Surprisingly, at the same time, the replacement of the most memorizing layers from the CIFAR10 trained encoder with STL10 layers also causes the highest increase in STL10 linear probing accuracy (again in comparison to replacing random or least memorizing layers). 
See a full set of results for replacing any combination of 1, 2, and 3 layers in \Cref{tab:replace-one-layer}, \Cref{tab:replace-two-layers}, and \Cref{tab:replace-three-layers}, respectively.
These results suggest that we might be able to improve standard encoder fine-tuning by localizing the most memorizing layers and fine-tuning these instead of the last layer(s)---currently the standard practice for fine-tuning in SSL.
We verify this assumption in \Cref{tab:better_fine_tuning} and show that fine-tuning the most memorizing layers indeed yields the highest downstream performance on the fine-tuning dataset.
This shows that localizing memorization might have practical application for more efficient fine-tuning in the future.

\begin{table}[t]
    \centering
    \small
        \caption{\reb{\textbf{Fine-tuning most memorizing layers.} We train a ResNet9 encoder with SimCLR on CIFAR10 and fine-tune different (combinations of) layers on the STL10 dataset, resized to 32x32x3. We train a linear layer trained on top of the encoder (HEAD) and report STL10 test accuracy after fine-tuning. Fine-tuning the most memorizing layer(s), in contrast to the last layer(s), yields higher fine-tuning results.}}
    \label{tab:better_fine_tuning}
\vspace{0.2cm}
    \begin{tabular}{cc}
    \toprule
        \textbf{Fine-tuned Layers} & \textbf{Accuracy (\%)} $\uparrow$ \\ \midrule
        None (HEAD) & 48.6\% ± 1.12\% \\ 
        \midrule
        6 (highest \deltamem) + HEAD & \textbf{53.0\% ± 0.86\%} \\
        8 (last layer, highest \layermem) + HEAD & 52.7\% ± 0.97\% \\ 
         \midrule
        6,8 + HEAD & \textbf{56.7\% ± 0.84\%} \\ 
        7,8 + HEAD & 55.3\% ± 0.77\% \\ \midrule
        4,6,8 (highest \deltamem) + HEAD & \textbf{57.9\% ± 0.79\%} \\ 
        6,7,8  + HEAD & 56.5\% ± 0.95\% \\ 
        \bottomrule
    \end{tabular}

\end{table}

\section{Unit-Level Localization of Memorization}
\label{sec:unit}

Experiments from the previous section highlight that we are able to localize the memorization of data points in particular layers of the SSL encoders. 
This raises the even more fundamental question on whether it is possible to trace down SSL memorization to a unit-level.
To answer this question, we design \unitmem, a new metric to localize memorization in individual units of SSL encoders.
We use the term \textit{unit} to refer to both an activation map from a convolutional layer (single-layer output channel) or an individual neuron within a fully connected layer. 
Our \unitmem metric quantifies for every unit $u$ in the SSL encoder how much $u$ is sensitive to, \ie memorizes, any particular training data point.
Therefore, \unitmem relates the \textit{maximum unit activation} that occurs for a data point $x_k$ in the training data (sub)set $\mathcal{D'} \subseteq \mathcal{D}$ with the \textit{mean unit activation} on all other data points in $\mathcal{D'}\setminus \{x_k$\}.

The design of \unitmem is inspired by the class selectivity metric defined for SL by~\citep{smorcos2018selectivity}.
Class selectivity was derived from selectivity indices commonly used in neuroscience~\citep{valois1982neuroscience,britten1992neuroscience,freedman2006neuroscience} and quantifies a unit's discriminability between different classes. It was used as an indicator of good generalization in SL. We provide more background in \Cref{appendix:class-selectivity}. 
To leverage ideas from class selectivity for identifying memorization, we integrate three fundamental changes in our metric in comparison to the class selectivity metric.
While class selectivity is calculated on \textit{classes} of the test set and relies on class \textit{labels}, our \unitmem is (1) \textit{label-agnostic} and (2) computed on individual data points from the \textit{training dataset} to determine their memorization.
Additionally, to account for SSL's strong reliance on augmentations, (3) we calculate \unitmem over the expectation on the augmentation set used during training. 
Research from the privacy community~\citep{yu2021does,liu2021encodermi} suggests that those augmentations leave a stronger signal in ML models than the original data point, \ie relying on the unaugmented point alone might under-report memorization. 
We verify this effect in \Cref{fig:without_with_augmentations} in \Cref{app:unitmem}. 
\reb{We note that through these fundamental changes \unitmem is able to measure memorization of \textit{individual data points} within a class rather than to solely distinguish between classes or concepts like the original class selectivity.
We provide further insights into this difference and perform experimental verification which highlights that \unitmem captures individual data points' memorization rather than capturing classes or concepts in \Cref{appendix:unit-mem-class-paragraphs}.}

To formalize our \unitmem, we first define the mean activation $\mu$ of unit $u$ on a training point $x$ as 
\begin{equation}
\label{eq:augmentations}
\mu_u(x)= {\underset{x' \sim \Aug(x) }{\mathbb{E}}}{\text{activation}}_u{(x')}\text{,}
\end{equation}
where the activation for convolutions feature maps is averaged across all elements of the feature map and for fully connected layers is an output from a single neuron (which is averaged across all patches of $x$ in ViTs).
Further, for the unit $u$, we compute the maximum mean activation $\mu_{max,u}$ across all instances from $\mathcal{D'}$, where $N=|\mathcal{D'}|$, as 
\begin{equation}
    \mu_{max,u} = \text{max}(\{ \mu_u(x_i)\}_{i=1}^{N})\text{.}
\end{equation}

Let $k$ be the index of the maximum mean activation $\mu_{u}(x_k)$, \ie the $argmax$.
Then, we calculate the corresponding \reb{mean} activity $\mu_{-max}$ across all the remaining $N-1$ instances from $\mathcal{D'}$ as
\begin{equation} 
\label{eq:mu_minus_max}
    \mu_{-max,u} = \reb{\text{mean}}(\{\mu_{u}(x_i)\}^{N}_{i=1, i \ne k})\text{.}
\end{equation}
Finally, we define the \unitmem of unit $u$ as
\begin{equation}
\unitmem_{\mathcal{D'}}(u) = \frac{\mu_{max,u} - \mu_{-max,u}}{\mu_{max,u} + \mu_{-max,u}}\text{.}
\label{eq:selectivity}
\end{equation}

The value of the \unitmem metric is bounded between 0 and 1, where 0 indicates that the unit is equally activated by all training data points, while value 1 denotes exclusive memorization, where only a single data point triggers the activation, while all other points leave the unit inactive. 

\subsection{Experimental Results and Observations}
\reb{We present our core results and provide detailed additional ablations on our \unitmem \Cref{app:unitmem}.}

\paragraph{Highly Memorizing Units Occur over Entire Encoder.}

\begin{figure}[t]
\vspace{-0.5em}
\centering
    \begin{subfigure}{0.3\columnwidth}
    \centering
    \includegraphics[width=0.95\linewidth]{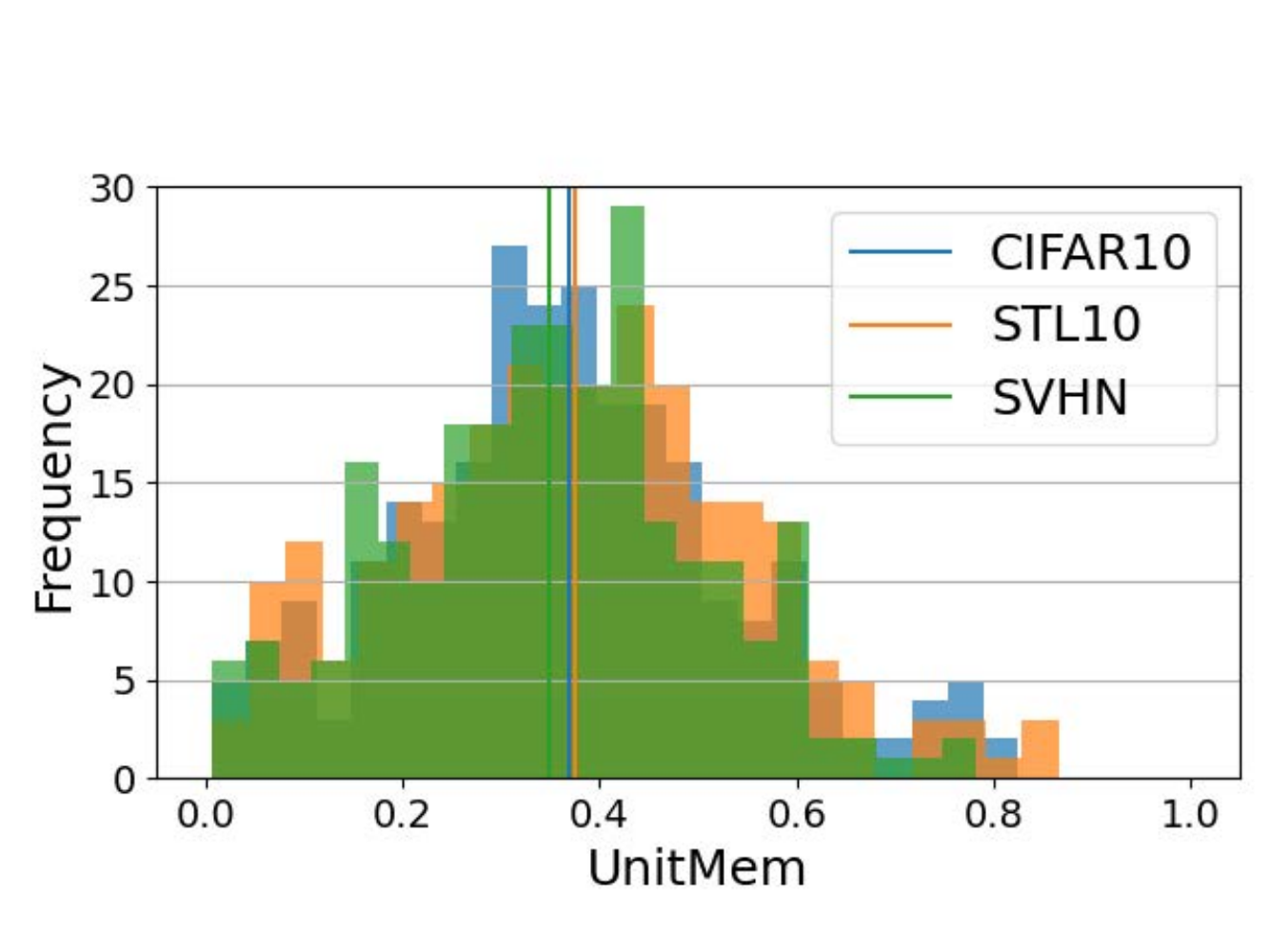}
    \caption{\textbf{Different datasets.}}
    \label{fig:compare_datasets_unitmem}
    \end{subfigure}
       \hfill
    \begin{subfigure}{0.3\columnwidth}
    \centering
    \includegraphics[width=0.95\linewidth]{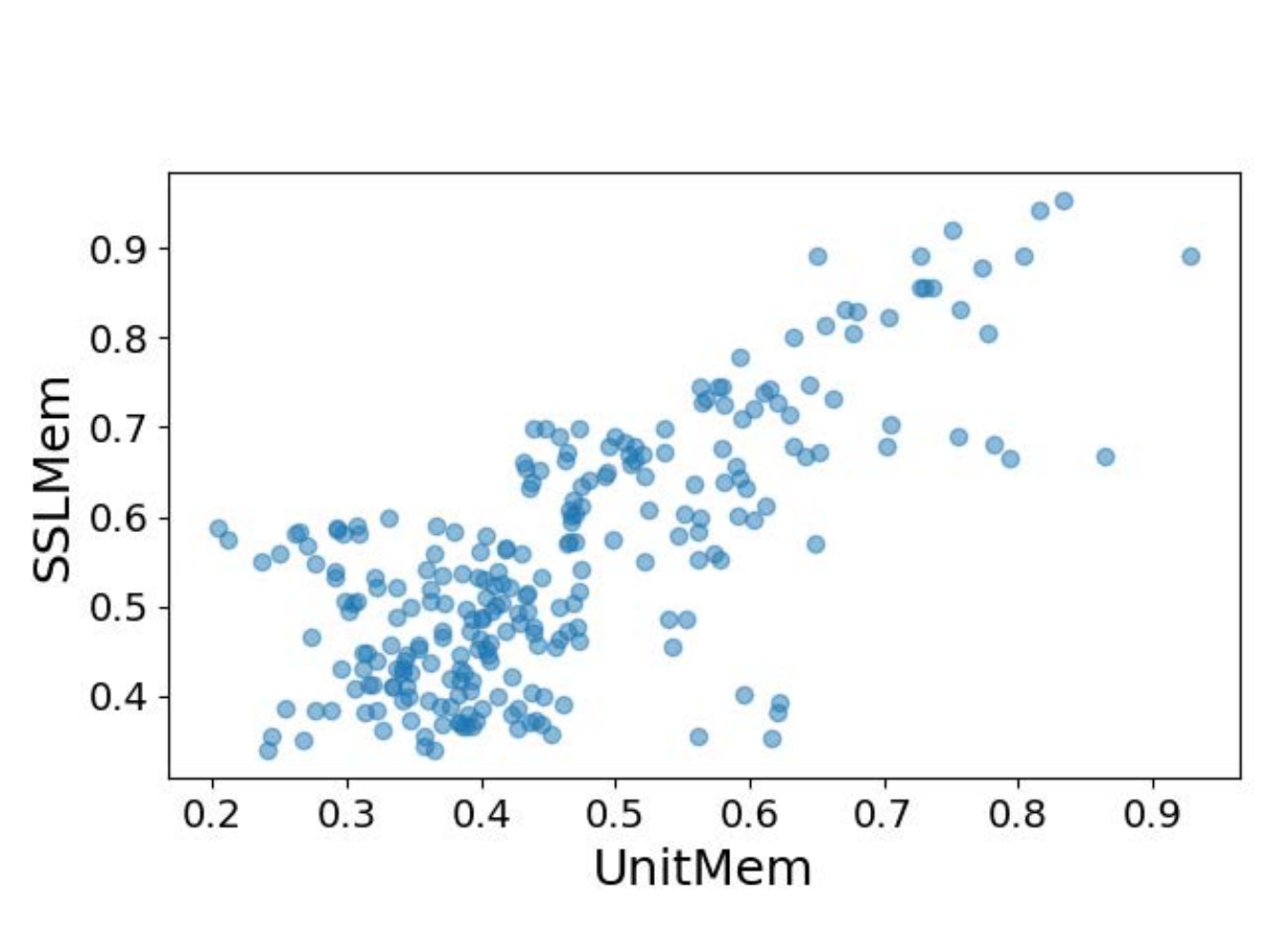}
    \caption{\textbf{\unitmem vs \sslmem.}}  
    \label{fig:sslmem-unitmem-align}
    \end{subfigure}
        \hfill
    \begin{subfigure}{0.3\columnwidth}
    \centering
    \includegraphics[width=0.95\linewidth]{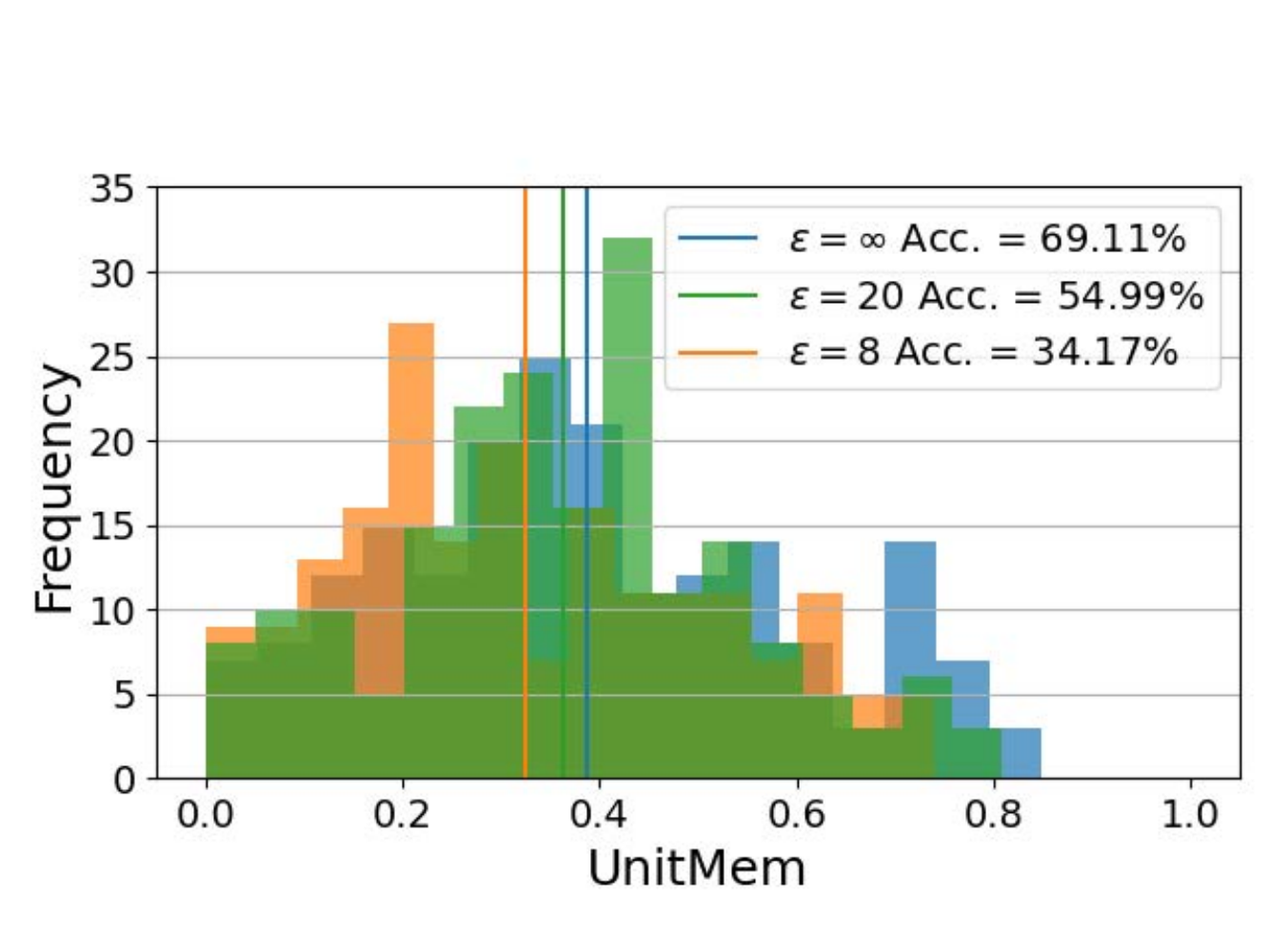}
    \caption{\textbf{\unitmem with DP.}}
    \label{fig:dp}
    \end{subfigure}
    \caption{
    \textbf{Insights into \unitmem.}
    We train a ResNet9 encoder with SimCLR: 
    \textbf{(a)} Different datasets, including SVHN, CIFAR10, and STL10. We report the \unitmem of the last convolutional layer (conv4\_2); 
    \textbf{(b)} Comparing alignment between \sslmem and \unitmem on CIFAR10. Data points with higher general memorization (\sslmem) tend to experience higher \unitmem;  \textbf{(c)} Using different strengths of privacy protection according to DP during training on CIFAR10 and Vit-Base
    }
    \label{fig:unitmem_insights}
    \vspace{-0.5cm}
\end{figure}

Our analysis highlights that over all encoder architectures and SSL training frameworks, highly memorizing units are spread over the entire encoder.
While, on average, earlier layers exhibit lower \unitmem than deeper layers, even the first layer contains highly memorizing units as shown in \Cref{fig:class-vs-unit-mem} (first row).
\Cref{fig:compare_datasets_unitmem} shows that this trend holds over different datasets. Yet, the SVHN dataset, which is visually less complex than the CIFAR10 or STL10 dataset, has the lowest number of highly memorizing units.
This observation motivates us to study the relationship between the highest memorized (atypical or hard to learn) data points and the highest memorizing units. 

\paragraph{Most Memorized Samples and Units Align.}
To draw a connection between data points and unit memorization, we analyze which data points are responsible for the highest $\mu_{max}$ scores. 
This corresponds to a data point which causes the highest activations of a unit, while other points activate the unit only to a small degree or not at all. 
We show the results in \Cref{fig:sslmem-unitmem-align} (also in \Cref{tab:sslmem-unitmem-align} as well as in \Cref{fig:leastmost_sl_ssl} in \Cref{app:unitmem}).
For each unit $u$ in the last convolutional layer of the ResNet9 trained on CIFAR10, we measure its \unitmem score, then we identify which data point is responsible for the unit's $\mu_{max,u}$, and finally measure this point's \sslmem score. We plot the \unitmem and \sslmem scores for each unit and its corresponding point.
Our results highlight that the data points that experience the highest memorization according to the \sslmem score are also the ones memorized in the most memorizing units.
Given the strong memorization in individual units, we next look into two methods to reduce it and analyze their impact.

\paragraph{Differential Privacy reduces Unit Memorization.}
The gold standard to guarantee privacy in ML is Differential Privacy (DP)~\citep{dwork2006differential}.
DP formalizes that any training data point should only have a negligible influence on the final trained ML model. 
To implement this, individual data points' gradients during training are clipped to a pre-defined norm, and controlled amounts of noise are added~\citep{abadi2016deep}. This limits the influence that each training data point can have on the final model.
Building on the DP framework for SSL encoders~\citep{yu2023vip}, we train a ViT-Tiny using MAE on CIFAR10 with three different privacy levels---in DP usually indicated with $\varepsilon$.
We train non-private ($\varepsilon=\infty$), little private ($\varepsilon=20$), and highly private ($\varepsilon=8$) encoders and apply our \unitmem to detect and localize memorization.
Our results in \Cref{fig:dp} highlight that while with increasing privacy levels, the average \unitmem decreases, there are still individual units that experience high memorization.

\paragraph{Data Point vs Class Memorization.}
Since stronger training augmentations yield higher class clustering~\citep{huang2021towards} \reb{(\ie the fact that data points from the same downstream class are close to each other in representation space but distant to data points from other classes)}, we also analyze how the SSL encoders differ from the standard class discriminators, namely SL trained models.
Therefore, we go beyond our previous experiments that measure memorization of units with respect to individual data points and additionally study unit memorization at a class-granularity. 
Therefore, we adjust the class selectivity metric from~\citep{smorcos2018selectivity} to perform on the training dataset rather than on the test data set as the original class selectivity. 
To avoid confusion between the two versions, refer to our adapted metric as \classmem (see \Cref{appendix:class-mem} for an explicit definition).
Equipped with \unitmem and \classmem, we study the behavior of SSL encoders and compare between SSL and~SL.
For our comparison, we train an encoder with SimCLR and a model with SL using the standard cross entropy loss, both on the CIFAR100 dataset using ResNet9. We remove the classification layer from the SL trained model to obtain the same architecture as for the encoder trained with SimCLR.
\begin{wrapfigure}{r}{0.3\textwidth}
\vspace{-0.2cm}
\begin{center}
\centerline{\includegraphics[width=0.3\columnwidth,trim={0.1cm 0.25cm 0 2.5cm},clip]{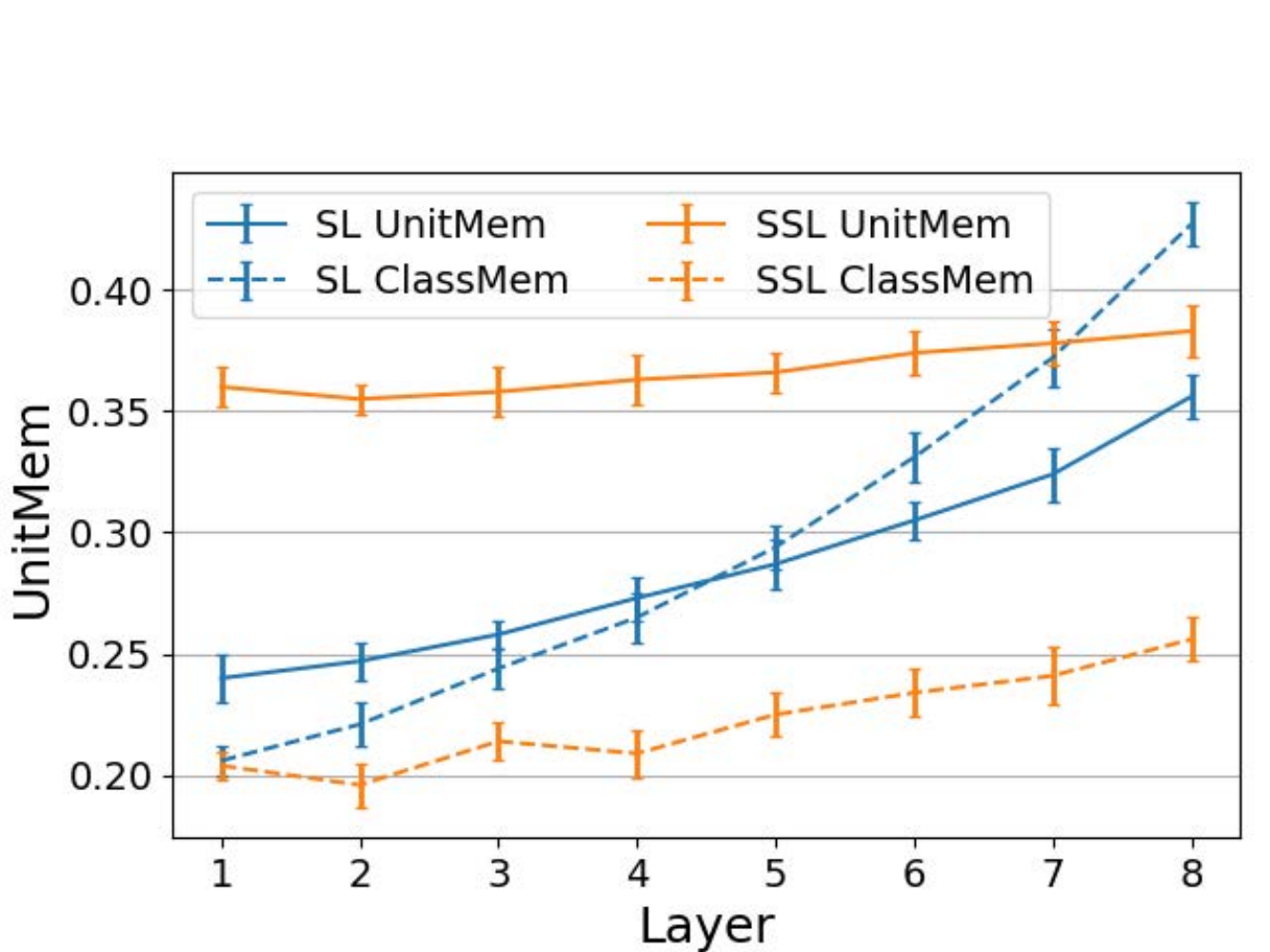}}
\vspace{-0.2cm}
\caption{
\label{fig:UMC_cifar100}
\textbf{\unitmem and \classmem for SL and SSL.
}
}
\end{center}
\vspace{-0.8cm}
\end{wrapfigure}

For comparability, we early stop the SL training once it reaches a comparable performance to the linear probing accuracy on CIFAR100 obtained by the SSL encoder.
Our results in \Cref{fig:UMC_cifar100} show that overall, in SSL throughout all layers, average memorization of individual data points is higher than class memorization, whereas in SL, in deeper layers, the class memorization increases significantly.
We hypothesize that this effect is due to earlier layers in SL learning more general features which are independent of the class whereas later layers learn features that are highly class dependent.
For SSL, such a difference over the network does not seem to exist; both scores increase slightly, however, probably due to the SSL learning paradigm, memorization of individual data points remains higher.

\begin{figure}[t]
    \centering
    \includegraphics[width=1\linewidth]{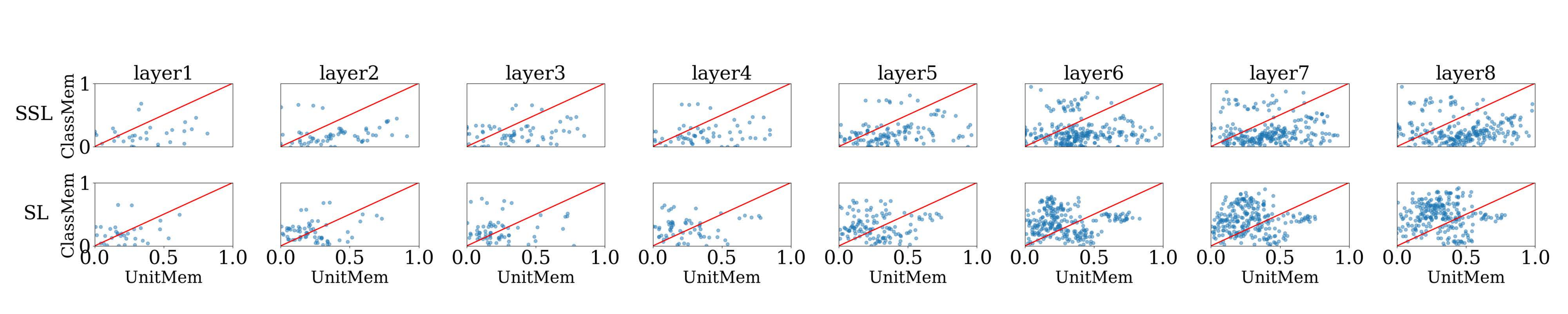}
    \vspace{-2em}
    \caption{\textbf{Significantly more \textit{(less)} units memorize data points rather than classes in SSL \textit{(SL)}.} 
    We measure the \classmem vs \unitmem for 10000 samples from CIFAR100, with 100 random samples per class.
    Each i-th column represents the i-th convolutional layer in ResNet9, with 8 convolution layers, where the 1st row is for SSL while the 2nd row for SL. 
    The red diagonal line denotes $y=x$.
    }
    \label{fig:class-vs-unit-mem}
    \vspace{0em}
\end{figure}

To better understand the memorization of units on the micro level, we investigate further the individual units in each layer. In \Cref{fig:class-vs-unit-mem}, we plot the \classmem vs \unitmem for each unit and in each of the eight encoder layers of ResNet9.
Most units for SSL (row 1) constantly exhibit higher \unitmem than \classmem, \reb{\ie they cluster under the diagonal line}, which suggests that most units memorize individual data points across the whole network. 
Contrary, the initial layers for the model trained with SL have a slight tendency to memorize data points over entire classes whereas in later layers, this trend drastically reverses and most units memorize classes.
\reb{In \Cref{app:ssl_mem_vs_sl_mem}, we investigate how this different memorization behavior between SL and SSL affects downstream generalization.}

\subsection{Verification of Unit-Based Memorization}
\label{sub:unit_verification}
We verify the unit-based localization of memorization with our \unitmem by deliberately inserting memorization of particular units and checking if our \unitmem correctly detects it.
Therefore, we first train a SimSiam-based~\citep{simsiam} ResNet18 encoder trained on the CIFAR10 dataset.
We select SimSiam over SimCLR for this experiment since SimCLR, as a contrastive SSL framework, cannot train on a single data point.
Then, using \layermem, we identity the last convolutional layer in ResNet18 (\ie layer \textit{4.1.conv2}) as the layer with the highest \layermem and \deltamem memorization.
We select the unit from the layer with the highest $\mu_{max}$ and also pick a unit with no activation ($\mu_{max}=0$) for some test data points. 
Then, we fine-tune these units using a single test data point and report the change in \unitmem for the chosen units in \Cref{tab:1filter}.
The results show that our \unitmem correctly detects the increase in memorization in both units. 
Additionally, we analyze the impact of zero-ing out the most or least memorizing vs. random units.
Again with the argument that memorization is required for downstream generalization in SSL~\citep{wang2024memorization}, we expect the highest performance drop when zero-ing out the most memorizing units.
Our results in \Cref{tab:pruning_strategy} in and in \Cref{app:pruning} confirm this hypothesis and show that removing the most memorizing units yields the highest loss in linear probing accuracy on various downstream tasks while pruning the least memorized units preserves better downstream performance than removing random units.
These results suggest that future work may benefit from using our \unitmem metric for finding which units within a network can be pruned while preserving high performance. 

\addtolength{\tabcolsep}{-3.5 pt}
\begin{table}[t]
\vspace{-0.5cm}
    \centering
    \tiny
        \caption{\textbf{Removing the least/most memorized units according to \unitmem preserves most/least linear probing performance}. 
    %We analyze the influence of pruning strategies on the linear probing accuracy. 
    We prune according to units with highest or lowest \unitmem either per layer or for the entire network (total). We also present baselines where we prune randomly selected units.
    \reb{The standard deviation for this baseline is reported over 10 independent trials where different random units were pruned.}
    We train the ResNet9 encoder using CIFAR10 and compute the \unitmem score using 5000 data points from the train set.
    }
    \vspace{0.2cm}
    \label{tab:pruning_strategy}
    \begin{tabular}{ccccc}
    \toprule 
    Pruning & \% of Selected & \multicolumn{3}{c}{\textbf{\textit{Downstream Accuracy (\%) }}} \\ \cline{3-5}
    Strategy & Units & CIFAR10 & SVHN & STL10 \\[0.0cm]  \midrule
    No Pruning & - & 70.44 & 78.22 & 69.12 \\\hdashline    
    Top \unitmem per layer & 10 & 53.04 & 63.84 & 50.94\\
    Random per layer & 10 & 58.09 $\pm$ 1.76 &	67.04 $\pm$ 2.44&55.71 $\pm$ 2.18\\
    Low \unitmem per layer & 10 & 62.58 & 72.26 & 59.26\\ 
    Top \unitmem per layer & 20 & 48.30 & 55.88 & 43.18\\
    Random per layer & 20 & 51.34 $\pm$ 1.21	&58.01$\pm$ 1.34	&46.74 $\pm$ 0.97\\
    Low \unitmem per layer & 20 & 54.84 & 62.60 & 50.02\\\hdashline   
    Top \unitmem total & 10 & 49.16 & 61.28 & 47.30 \\
    Random total & 10 & 56.77 $\pm$ 2.09&	67.09 $\pm$ 1.56&	53.89 $\pm$ 2.33\\
    Low \unitmem total & 10 &  62.62  & 72.28 & 59.30\\ 
    \bottomrule
    \end{tabular}
    \vspace{0.5em}
    \vspace{-3.5em}
\end{table}
\addtolength{\tabcolsep}{3.5 pt}

\vspace{-0.5em}
\section{Conclusions}
\vspace{-0.5em}
We propose the first practical metrics for localizing memorization within SSL encoders on a per-layer and per-unit level.
By analyzing different SSL architectures, frameworks, and datasets using our metrics, we find that while memorization in SSL increases in deeper layers, a significant fraction of highly memorizing units can be encountered over the entire encoder.
Our results also show that SSL encoders significantly differ from SL trained models in their memorization patterns, with the former constantly memorizing data points and the latter increasingly memorizing classes.
Finally, using our metrics for localizing memorization presents itself as an interesting direction towards more efficient encoder fine-tuning and pruning.

%\newpage
%\pagebreak

\subsubsection*{Acknowledgments}
The project on which this paper is based was funded by the Deutsche Forschungsgemeinschaft (DFG, German Research Foundation), Project number 550224287. Additional funding came from the Initiative and Networking Fund of the Helmholtz Association in the framework of the Helmholtz AI project call under the name „PAFMIM“, funding number ZT-I-PF-5-227. Responsibility for the content of this publication lies with the authors.

\bibliography{main}
\bibliographystyle{plainnat}

\newpage
\appendix
\onecolumn
\section{Glossary}
\label{app:glossary}

\reb{For the reader's convenience, we provide a glossary with all important terms and concepts related to our work in \Cref{tab:glossary}.}

\begin{table}[h]
\centering
\small
\caption{\reb{\textbf{Glossary.} We present a concise overview on the concepts relevant to this work.}}
\vspace{0.2cm}
\begin{tabularx}{\linewidth}{@{}>{\bfseries}l@{\hspace{.5em}}X@{}}
\toprule
\textbf{Concept}                   & \textbf{Explanation}                                                                                                                                                                                                         \\ \midrule
Atypical examples         & Data points that are uncommon in the data distribution and different in terms of their features. Examples: Figure 1 from \citep{wang2024memorization}. Sometimes also called “outliers”.                                        \\
Class Selectivity         & A metric proposed by \citep{smorcos2018selectivity} which quantifies a unit’s discriminability between different classes, measured on the test data.                                                                                \\
ClassMem                  & Our adaptation of Class Selectivity measured on the training data.                                                                                                                                                  \\
Downstream Generalization & Expresses how well an encoder is suited to solve some downstream tasks. For classification, it is often measured by linear probing, i.e., training an additional classification layer on top of the encoder output. \\
LayerMem                  & Our proposed metric to quantify memorization of any layer in the SSL encoder.                                                                                                                                       \\
Memorization              & A phenomenon where a machine learning model stores detailed information on its training data.                                                                                                                       \\
Memorized Data Point      & A data point that experiences high memorization by a machine learning model.                                                                                                                                        \\
Memorization Pattern      & A general trend in the low-level memorization of an SSL encoder, i.e., in which layers or units do memorization localize.                                                                                           \\
Unit                      & Term used to refer to an individual neuron in fully connected layers or a channel in convolutional layers.                                                                                                          \\
UnitMem                   & Our proposed metric to quantify the memorization of any unit in the SSL encoder.                     \\
    \bottomrule
\end{tabularx}

\label{tab:glossary}
\end{table}

\section{Experimental Setup}
\label{apendix:experimental-setup}

\paragraph{Datasets.}
We base our experiments on ImageNet ILSVRC-2012~\citep{russakovsky2015imagenet}, CIFAR10~\citep{krizhevsky2009learning}, CIFAR100~\citep{krizhevsky2009learning}, SVHN~\citep{netzer2011reading}, and STL10~\citep{coates2011analysis}. 

\paragraph{Models.}
We use the ResNet family of models~\citep{He2015DeepRL}, including ResNet9, ResNet18, ResNet34, and ResNet50. In \Cref{tab:resnet9_architecture}, we present the detailed architecture of the ResNet9 model.

\addtolength{\tabcolsep}{-3pt} 
\begin{table}[h]
    \centering
    \tiny
        \caption{\textbf{Architecture of ResNet9.} In the \textit{Number of Units} column, we present the number of activation maps (corresponding to individual filters in the filter bank). }
            \vspace{0.2cm}
    \begin{tabular}{cccc}
    \toprule 
    Conv-Layer ID & Layer Name & Number of Units & Number of Parameters\\
    \midrule %Since we have 256 units for layer 6 7 8, 128 for layer5, 64 for layer4 3 2, and 32 units for layer1
         1 & Conv1 & 32 & 896 \\
         - & BN1 & 32 & 64\\
         - & MaxPool1 & 32 & 0 \\
         2 & Conv2-0 & 64 & 18496 \\
         - & BN2-0 & 64 & 128 \\
         - & MaxPool2-0 & 64 & 0 \\
         3 & Conv2-1 & 64 & 36928\\
         - & BN2-1 & 64 & 128 \\
         - & MaxPool2-1 & 64 & 0 \\
         4 & Conv2-2 & 64 & 36928\\
         - & BN2-2 & 64 &  128 \\
         - & MaxPool2-2 & 64 & 0 \\
         5 & Conv3 & 128 &  73856\\
         - & BN3 & 128 & 256 \\
         - & MaxPool3 & 128 & 0 \\
         6 & Conv4-0& 256 & 295168 \\
         - & BN4-0 & 256 & 512 \\
         - & MaxPool4-0 & 256 & 0 \\
         7 & Conv4-1& 256 & 590080 \\
         - & BN4-1 & 256 & 512 \\
         - & MaxPool4-1 & 256 & 0 \\
         8 & Conv4-2& 256 & 590080\\
         - & BN4-2 & 256 & 512 \\
         - & MaxPool4-2 & 256 & 0 \\
       %  \midrule
         %- & \textit{Head} (train) & - \\
         % \midrule
         % - & FC (fine-tune) & - \\
         % - & softmax (fine-tune) & - \\
         
        \bottomrule
    \end{tabular}

    \label{tab:resnet9_architecture}
\end{table}
\addtolength{\tabcolsep}{3pt} 

\paragraph{SSL Frameworks.}
We base our experimentation on four state-of-art SSL encoders: MAE~\citep{mae}, SimCLR~\citep{chen2020simple}, DINO~\citep{caron2021dino}, and SimSiam~\citep{simsiam}.

\paragraph{Training Setup.}
Our experimental setup for training the encoders mainly follows~\citep{wang2024memorization} and we rely on their naming conventions and refer to the data points that are used to train encoder $f$, but not reference encoder $g$ as \textit{candidate} data points.
In total, we use 50000 data points as training samples for CIFAR10, SVHN, and STL10 and 100000 for ImageNet with 5000 candidate data points per dataset.
The encoders evaluated in the paper are trained with batch size 1024, and trained 600 epochs for CIFAR10, SVHN, and STL10, and 300 epochs for ImageNet.
We set the batch size to 1024 for all our experiments and train for 600 epochs on CIFAR10, SVHN, and STL10, and for 300 epochs on ImageNet.
As a distance metric to measure representation alignment, we use the $\ell_2$ distance. 
We repeat all experiments with three independent seeds and report average and standard deviation.
For reproducibility, we detail our full setup in \Cref{tab:settings} with the standard parameters that are used throughout the paper if not explicitly specified otherwise. 

\paragraph{Training Augmentations.}
\label{par:augmentations}
We generate augmentations at random from the following augmentation sets (p indicates augmentation probability):
\begin{itemize}
    \item \textbf{SL, standard, \reb{(referred to as \textit{weak augmentations)}}:} ColorJitter(0.9-0.9-0.9-0.5, p=0.4), RandomHorizontalFlip(p=0.5), RandomGrayscale(p=0.1), RandomResizedCrop(size=32)
    \item \textbf{SSL, standard, \reb{(referred to as \textit{normal augmentations)}}:} ColorJitter(0.8-0.8-0.8-0.2, p=0.8), RandomHorizontalFlip(p=1.0), RandomGrayscale(p=0.2), RandomResizedCrop(size=32)
    \item \textbf{SSL, stronger, \reb{(referred to as \textit{strong augmentations)}}:} ColorJitter(0.8-0.8-0.8-0.2, p=0.9), RandomHorizontalFlip(p=1.0), RandomGrayscale(p=0.5), RandomResizedCrop(size=32), RandomVerticalFlip(p=1.0)
    \item \textbf{SSL (independent):} GaussianBlur(kernel\_size=(4, 4), sigma=(0.1, 5.0), p=0.8), RandomInvert(p=0.2), RandomResizedCrop(size=32), RandomVerticalFlip(p=1.0)
    \item \textbf{Masking (MAE):} 75\% random masking
\end{itemize}

\paragraph{Details on Computing \unitmem.}
Relying on insights from~\citep{encoderMI}, we calculate ~\Cref{eq:augmentations} for the activations within \unitmem over ten augmentations since this has shown to yield a strong signal on the augmented data point.
For convolutional feature maps, the activation of the unit is calculated as the average of all elements in the feature map.
In ViTs, where we measure activation over fully-connected layers, we compute the activation per neuron and average across all patches of a given input. For example, for ViT Tiny encoder pretrained on CIFAR10, the input image of resolution 32x32 is \textit{patchified} into 64 patches, each of size 4x4. Then, each patch is represented by a 192 dimensional embedding.
The classification (CLS) embedding is prepended to the remaining 64 embeddings. Overall, we obtain 65 patches. The last fully connected layer has 192 neurons. For each neuron, we average its activations across the 65 patches. In the case of ViT Base, we have 768-dimensional embeddings and 197 patches for the input image of resolution 224x224.

\paragraph{Details on Fine-Tuning with one Test Data Point}
We provide the exact details of our experiments to verify our \unitmem through deliberate insertion of memorization in \Cref{sub:unit_verification}.
We train a SimSiam-based~\citep{simsiam} ResNet18 encoder on the CIFAR10 dataset and use \layermem to identity layer \textit{4.1.conv2}, \ie the last convolutional layer in ResNet18, as the layer with highest accumulated memorization.
We select the unit from the layer with the highest $\mu_{max}$ and also pick a unit with no activation ($\mu_{max}=0$) for some test data points. 
Then, for compatability with pytorch which does not support individual unit training, we lock all parameters except for the targeted \textit{layer} and train the model with a single sample from the testing dataset.
We choose the sample that achieves the highest activation $\mu_{max}$ on the unit with the highest \unitmem. 
We save the checkpoints after each epoch and test the $\mu_{max}$ for the selected two units. 
Our results in \Cref{tab:1filter} show that the value of $\mu_{max}$ for the selected data point increases in both units and the data point remains the one responsible for the $\mu_{max}$. 

\paragraph{Details on Hardware resources usage}\label{app:hardware}

We finish all our experiments on two devices: a cloud server with four A100 GPUs and a local workstation with Intel 13700k CPU, Nvidia 4090 graphics card and 64GB of RAM

\addtolength{\tabcolsep}{-2.5 pt}
\begin{table}[h]
    \centering
    \scriptsize 
    \setlength{\tabcolsep}{3pt}
    \caption{\textbf{Training Setup for SSL Frameworks and Hyperparameters.} Two numbers denote ImageNet / Others.}    
    \vspace{0.2cm}
\begin{tabular}{cccccccccc}
\toprule
                       & \multicolumn{4}{c}{Model Training}                                       &  & \multicolumn{4}{c}{Linear Probing}                      \\ \cmidrule{2-5} \cmidrule{7-10} 
                       & MAE                         & SimCLR       & DINO         & SimSiam       &  & MAE          & SimCLR       & DINO         & SimSiam       \\ \midrule
Training Epochs         & 300 / 600 & 300 / 600      & 300 / 600      & - / 200     &  & 45 / 90        & 45 / 90        & 45 / 90        &    - / 30    \\
Warm-up Epochs         & 30 / 60                       & 30 / 60        & 30 / 60        &    - / 24    &  & 5 / 10         & 5 / 10         & 5 / 10         &    - / 3      \\
Batch Size             & 2048                        & 4096         & 1024         &     128      &  & 4096         & 4096         & 4096         & 256      \\
Optimizer              & AdamW                      & LARS        & AdamW        &   SGD     &  & LARS         & LARS         & LARS         &   SGD       \\
Learning rate          & 1.2e-3                      & 4.8          & 2e-3         &   2.5e-2       &  & 1.6          & 4.8          & 1.6          &  5e-2         \\
Learning rate Schedule & Cos. Decay                & Cos. Decay & Cos. Decay & Cos. Decay &  & Cos. Decay & Cos. Decay & Cos. Decay & Cos. Decay \\ \bottomrule 
\end{tabular}
    
          \label{tab:settings}
\vspace{-0.5cm}
\end{table}
\addtolength{\tabcolsep}{2.5 pt}

\section{Additional Experiments}

\subsection{Additional Insights into \unitmem}
\label{app:unitmem}

\begin{wrapfigure}{r}{0.3\textwidth}
\vspace{-0.4cm}
\begin{center}
\centerline{\includegraphics[width=\xintu\textwidth,trim={3cm 1.5cm 0 2.5cm}]{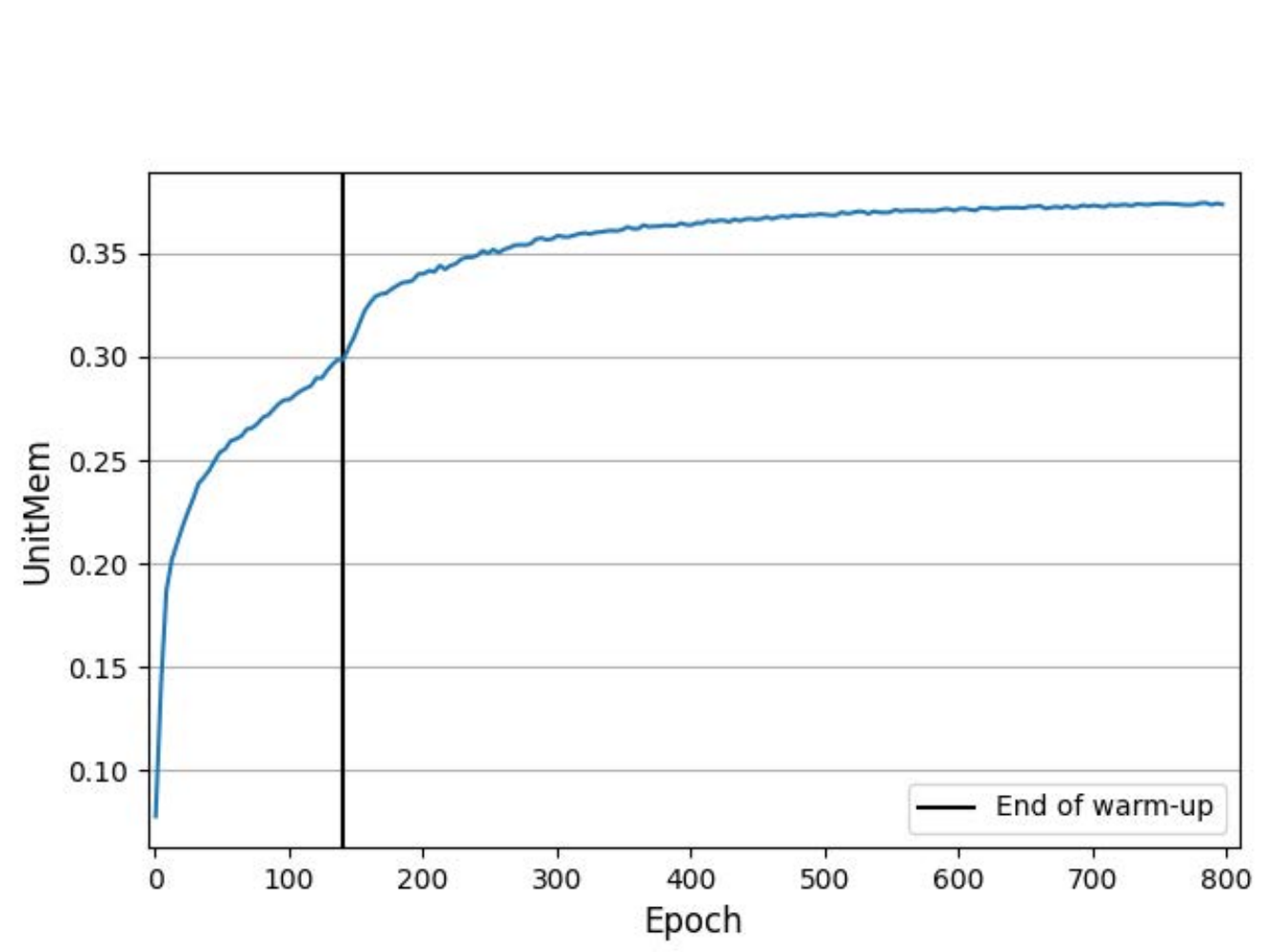}}
\caption{
\label{fig:selectivity_during_training}
\textbf{Average \unitmem of layer 8 over training.
}
}
\end{center}
\vspace{-1.0cm}
\end{wrapfigure}

\textbf{\unitmem increases over training.} First, we assess how \unitmem evolves over training of the SSL encoder. Therefore, we train a ResNet9 encoder using SimCLR on the CIFAR10 dataset for 800 epochs, using 120 warm-up epochs.
Every five epochs, we measure the \unitmem. Our results in \Cref{fig:selectivity_during_training} depict the average \unitmem of the ResNet9's last convolutional layer.

We observe that the unit memorization monotonically increases throughout training and that the increase is particularly high during the first epochs.
After the warm-up, we observe that the increase in unit memorization stagnates until the level of memorization on the unit level converges.
The same trend can be observed over all layers which indicates that SSL encoders increase unit memorization throughout training.

\textbf{Measuring \unitmem without using augmentations leads to an under-reporting of memorization.}

\begin{wrapfigure}{r}{0.3\textwidth}
\vspace{-0.4cm}
\begin{center}
\centerline{\includegraphics[width=\xintu\textwidth,trim={3cm 1.5cm 0 2.5cm}]{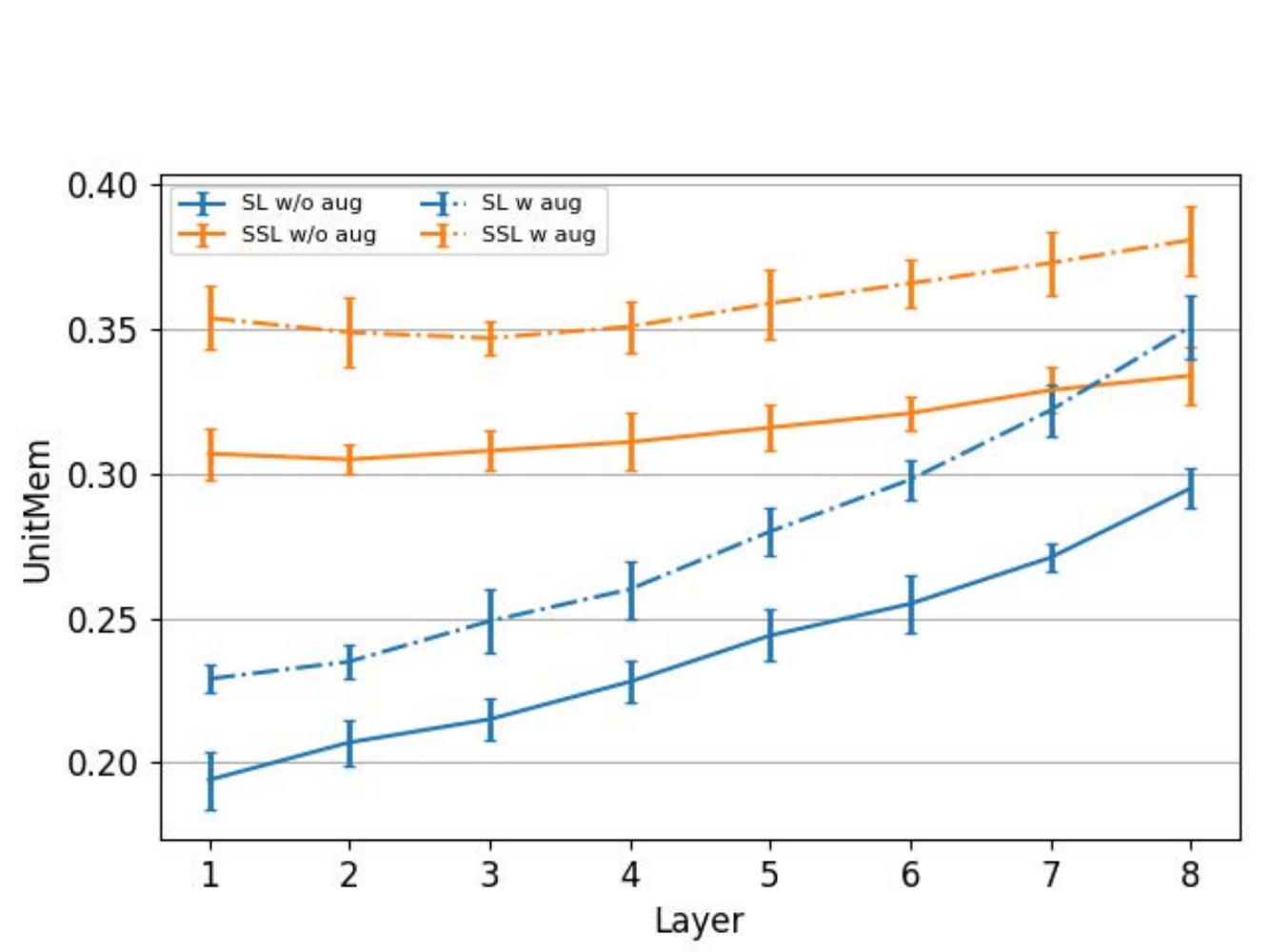}}
\caption{\textbf{\unitmem w \& w/o augmentations.}}
\label{fig:without_with_augmentations}
\end{center}
\vspace{-0.8cm}
\end{wrapfigure}

To assess the impact on using augmentation to implement \Cref{eq:augmentations} for the calculation of our \unitmem has an impact on the reported results, we train two ResNet9 models on the CIFAR100 dataset, one using SimCLR, the other one using standard SL with cross entropy loss.

During training we rely on the standard augmentations for SL and SSL reported above.
To measure memorization, we once use ten augmentations from the training augmentation set, and no augmentations otherwise and report the results in \Cref{fig:without_with_augmentations}.
We find that while the trend of the reported memorization is equal in both settings, the \unitmem measured without augmentations remains constantly lower than when measured with augmentations.
This suggests that when measuring \unitmem, it is important to use augmentations to avoid under-reporting of the memorization.

\begin{wrapfigure}{r}{0.3\textwidth}
\vspace{-0.4cm}
\begin{center}
\centerline{\includegraphics[width=\xintu\textwidth,trim={3cm 1.5cm 0 2.5cm}]{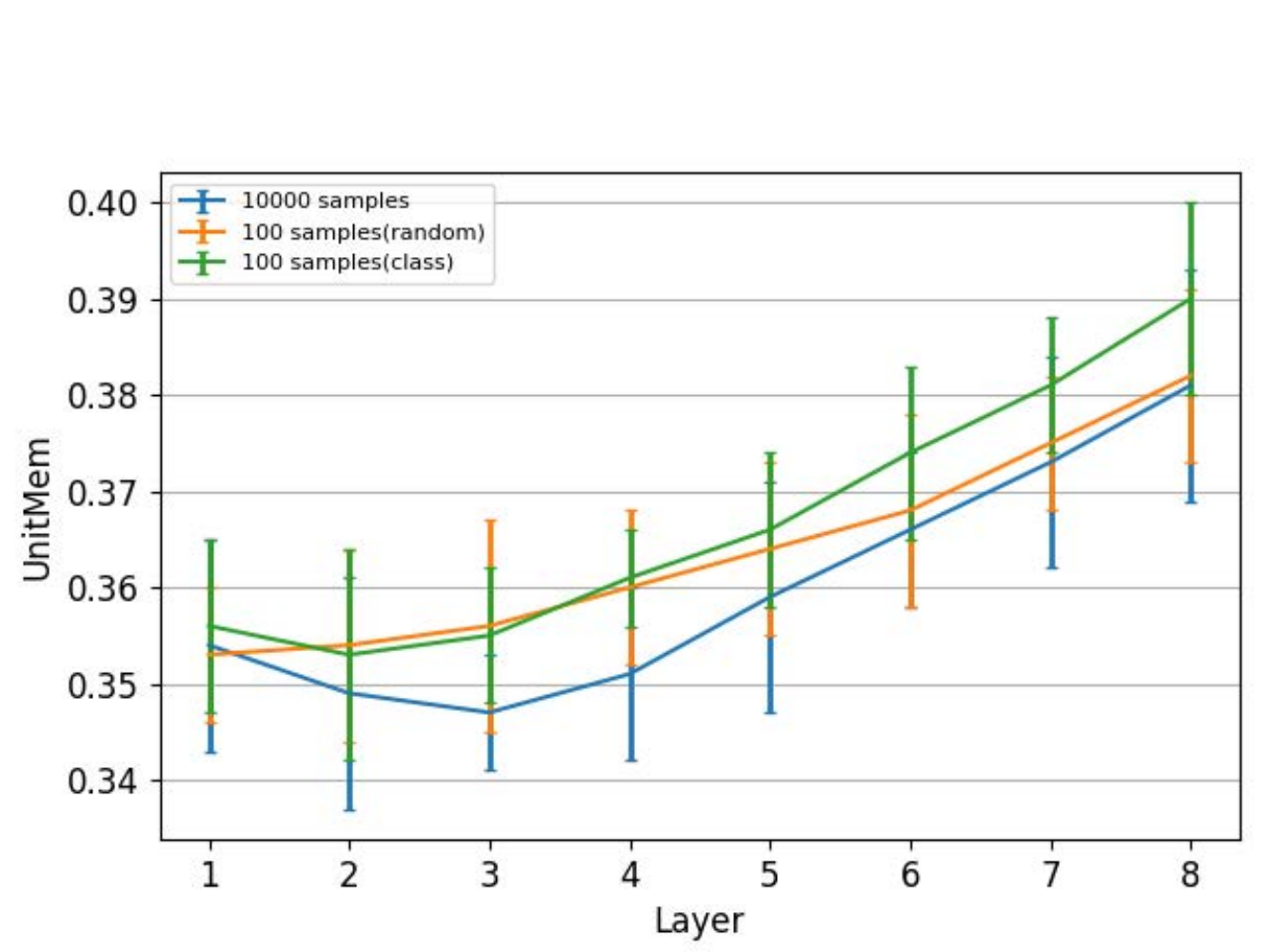}}
\caption{
\label{fig:number_of_samples}
\textbf{Size of $\mathcal{D'}$.}}
\end{center}
\vspace{-0.8cm}
\end{wrapfigure}

\paragraph{The number of data points used to measure \unitmem does not have a significant impact on the reported memorization.}
Using the same ResNet9, trained with SimCLR on CIFAR100, we assess whether the number of data points that we use to calculate \unitmem (the size of $\mathcal{D'}$) has an impact on the reported memorization.
Then, we measure \unitmem using 100 random data point chosen one from each class in CIFAR100, 100 purely randomly chosen data points, and randomly chosen CIFAR100 data points.
We present our findings in \Cref{fig:number_of_samples}. 
Our results highlight that all the lines are within each other's standard deviation, indicating that there is no significant difference in the reported \unitmem, dependent on the make up of the dataset $\mathcal{D'}$.

\begin{wrapfigure}{r}{0.3\textwidth}
\vspace{-0.6cm}
\begin{center}
\centerline{\includegraphics[width=\xintu\textwidth,trim={3cm 1.5cm 0 2.5cm}]{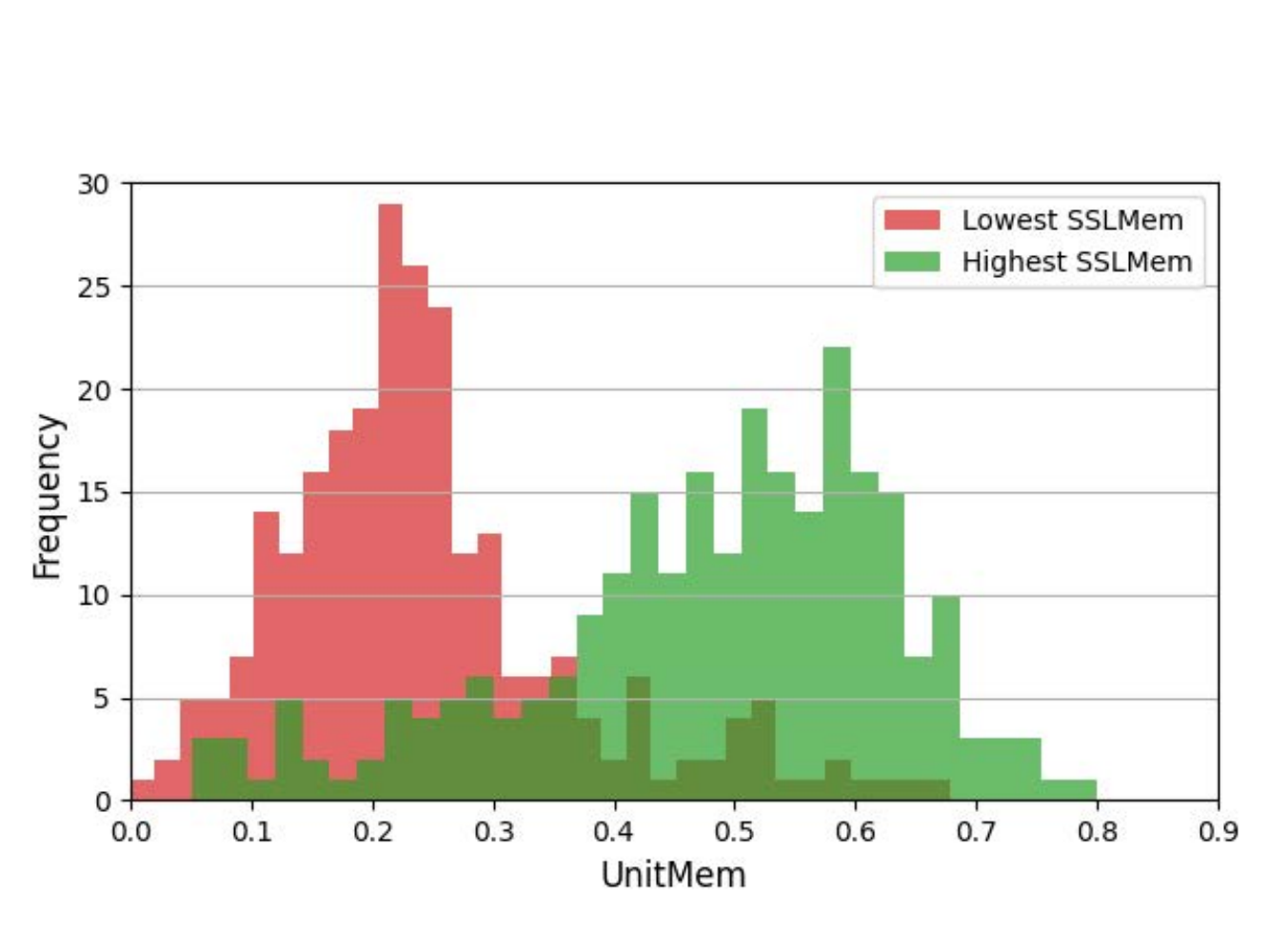}}
\caption{
\label{fig:leastmost_sl_ssl}
\textbf{Least vs most memorized data points.
}
}
\end{center}
\vspace{-0.8cm}
\end{wrapfigure}

\paragraph{Most memorized data points align with the most memorizing units.}
We train a ResNet9 on CIFAR10 using SimCLR and measure \unitmem for the 300 most and 300 least memorized data points identified using \sslmem by \citep{wang2024memorization}.
The measurement of the two sets (most vs least memorized data points) is performed independently.
Our results in \Cref{fig:leastmost_sl_ssl} show that the \unitmem calculated on the most memorized data points is significantly higher than on the least memorized data points (we verify the significance with a statistical $t$-test in \Cref{tab:resnet9_per_layer_our_memorization_score_t_test}.

While also some of the least memorized data points lead to a high activation of the units, highest activation (on average and in particular) can be observed for the most memorized data points.
This underlines the trend observed in \Cref{tab:sslmem-unitmem-align} which shows that highly memorized data points align with the highly memorizing units.

\paragraph{Computing \unitmem based on the median yields similar results to using the mean.}
Our \unitmem metric is inspired by the class selectivity defined for SL by~\citep{smorcos2018selectivity} which quantifies a unit's discriminability between different classes, see \Cref{appendix:class-selectivity}. 
Yet, we calculate the $\mu_{-max,u}$ in ~\Cref{eq:mu_minus_max} using the \textit{median} on the other individual training data points' activations while \classselectivity computes their equivalent of $\mu_{-max,u}$ using the \textit{mean} on all other test classes' activations.

\begin{wrapfigure}{r}{0.6\textwidth}
\vspace{-0.6cm}
    \centering
    \subfloat[Median]{\includegraphics[scale=0.18]{figures/PDF/leastmost_300samples_10class.pdf}
    \vspace{-0.4cm}
    }
    \subfloat[Mean]{\includegraphics[scale=0.18]{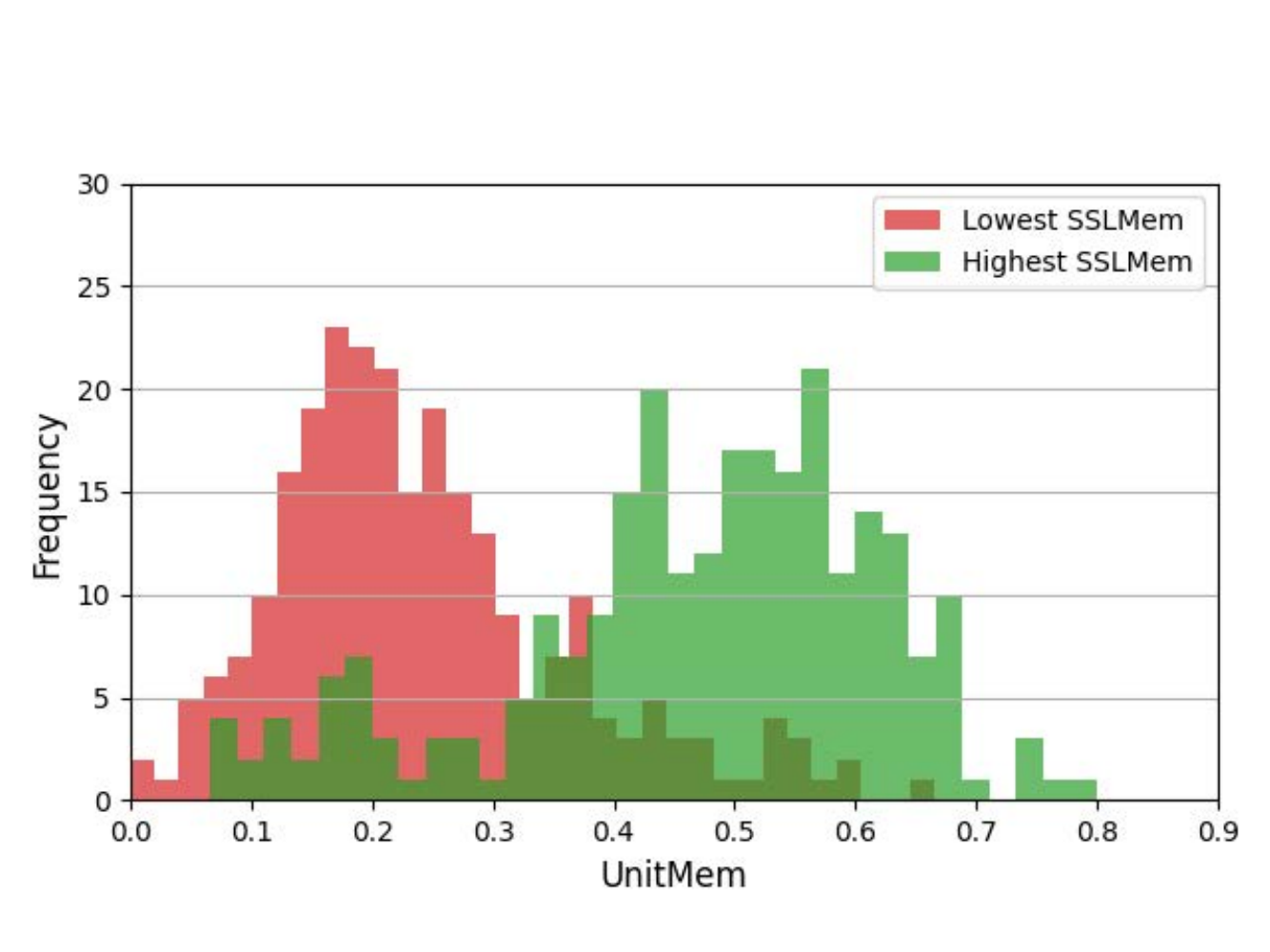}
    \vspace{-0.4cm}
    }
    \vspace{-0.1cm}
    \caption{\textbf{Mean vs Median.} 
    }
    \label{fig:median_mean}
\vspace{-0.8cm}
\end{wrapfigure}

We show in \Cref{fig:median_mean} over the 300 most and least memorized data points for a ResNet9 trained with SimCLR on CIFAR10 that using the median for \unitmem yields very similar results to using the mean. % but is a more robust metric.

\begin{wrapfigure}{r}{0.6\textwidth}
\vspace{-0.6cm}
    \centering
    \subfloat[SSL]{\includegraphics[scale=0.18]{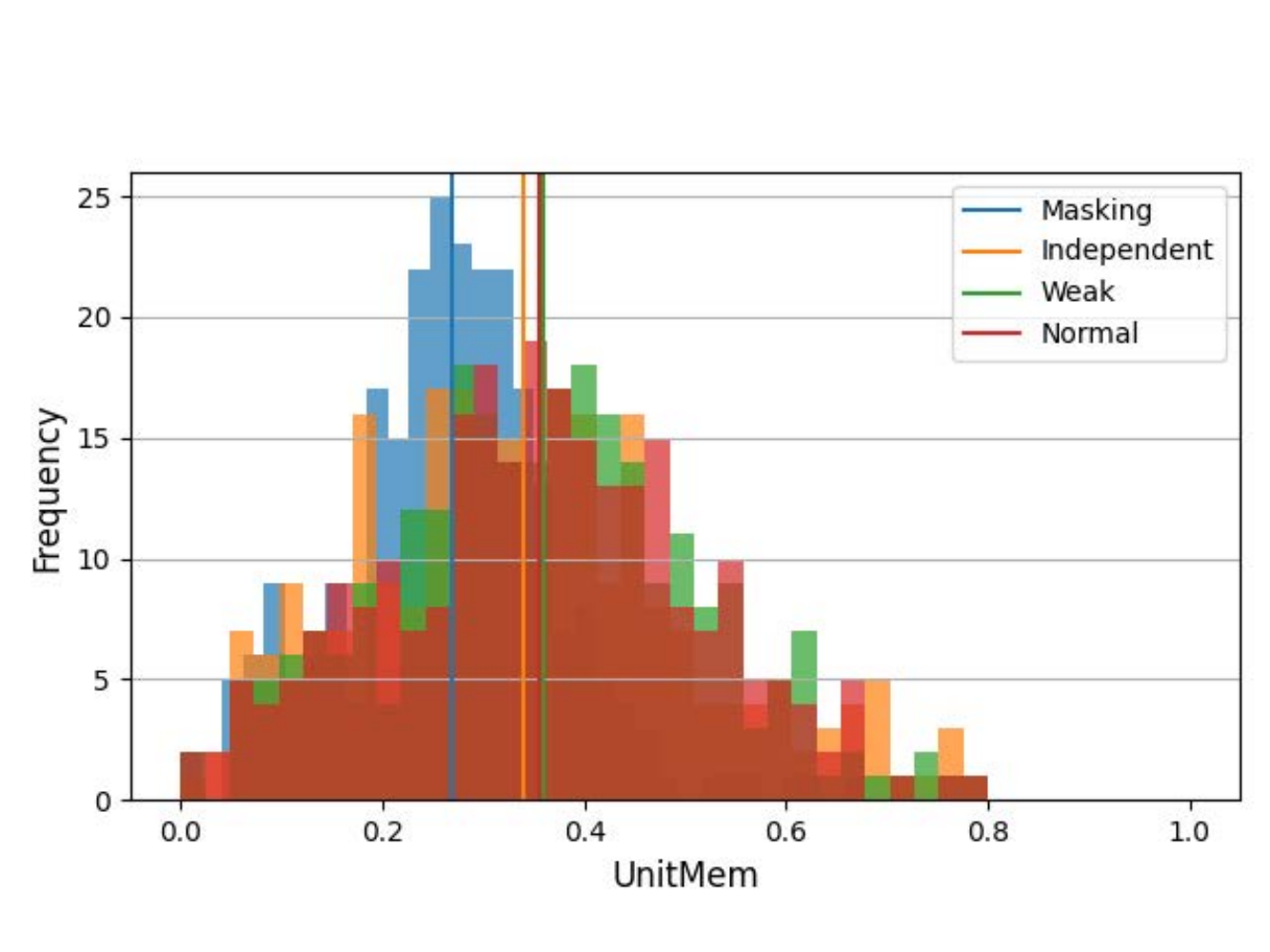}
    \vspace{-0.4cm}
    }
    \subfloat[SL]{\includegraphics[scale=0.18]{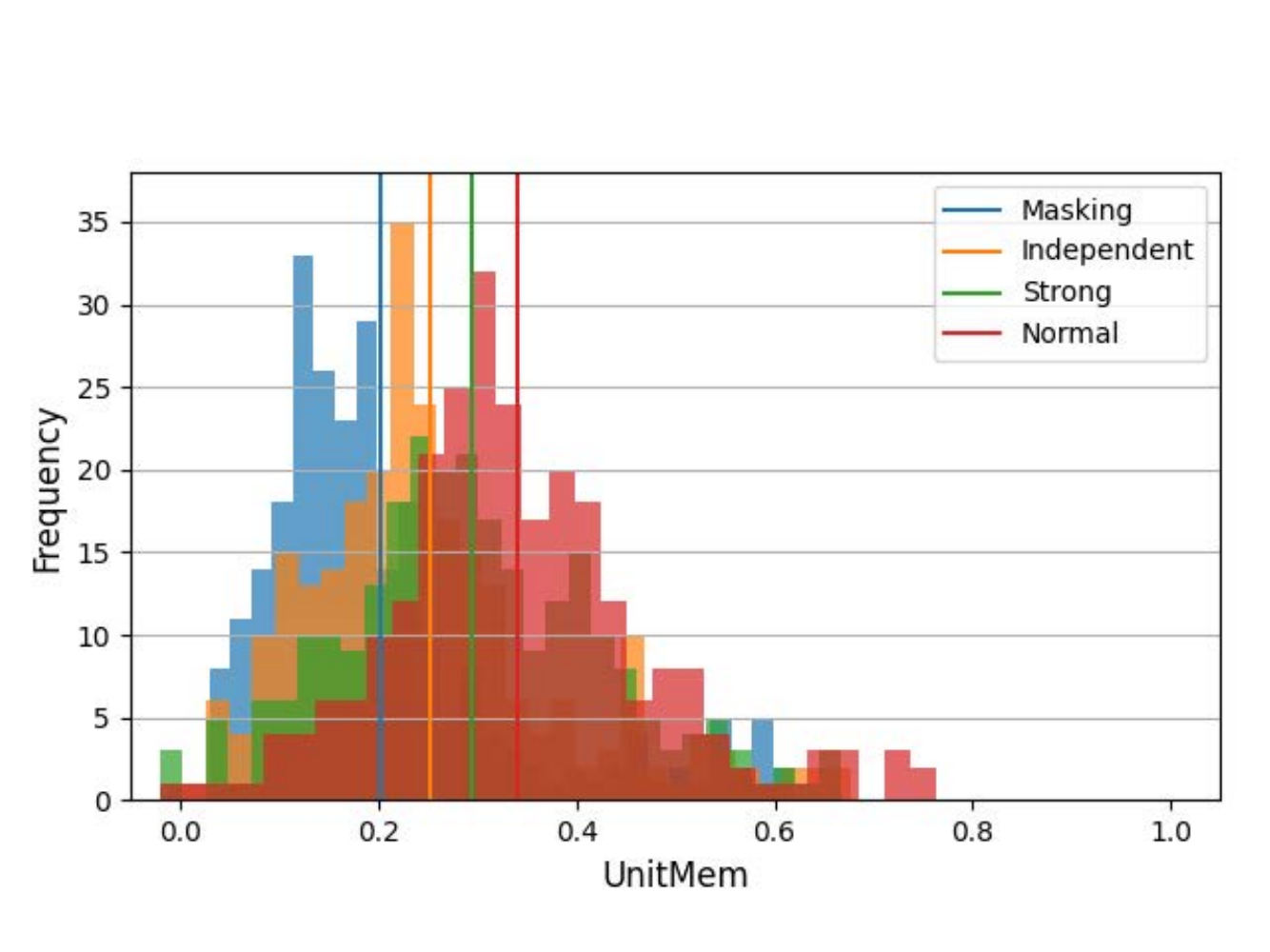}
    \vspace{-0.4cm}
    }
    \caption{\textbf{Different augmentation sets.} 
    }
    \label{fig:aumentation_exp}
\vspace{-0.6cm}
\end{wrapfigure}

\paragraph{For SSL, the concrete augmentation set has no strong impact when measuring \unitmem .}
We additionally set out to study the impact of the augmentation set used to calculate \unitmem. Therefore, we calculate \unitmem on the ResNet9 trained on CIFAR10 using SimCLR using different augmentations sets.
For SSL, we measure once with the standard training augmentations ("Normal"), with an independent set of augmentations of similar strength ("Independent"), with a weaker augmentation set for which we rely on the augmentations used to train the SL model ("Weak"), and an independent very strong set of augmentations modeled after MAE training and using a masking of 75\% of the input image ("Masking"). Our results in
\Cref{fig:aumentation_exp} depict the \unitmem over the last convolutional layer of the ResNet9 encoders.

They highlight that the weak and independent augmentations report extremely similar \unitmem to the original set of training augmentations used. 
For SL, the impact of using different augmentations during training and measuring \unitmem is more expressed. We also measured for a weak augmentation set ("Normal"), an independent weak set ("Independent"), strong augentations for which we relied on the standard SSL augmentations ("Strong"), and the 75\% masking ("Masking"). 
We observe that using the augmentations from training to calculate \unitmem yields the highest localization of memorization.

\begin{wrapfigure}{r}{0.3\textwidth}
\vskip \vskipintu
\vspace{0.55cm}
\begin{center}
\centerline{\includegraphics[width=\xintu\textwidth,trim={3cm 1.5cm 0 2.5cm}]{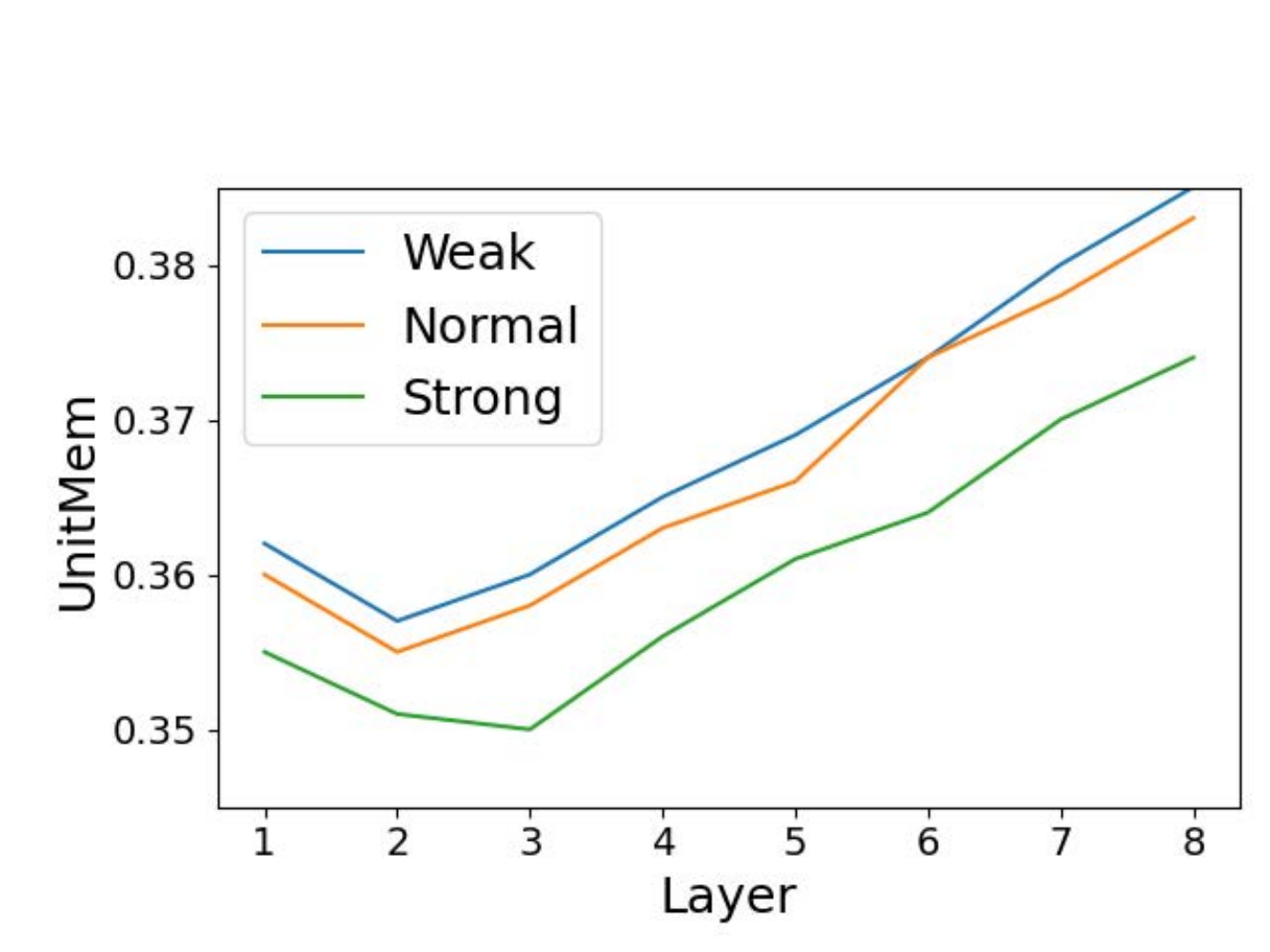}
    }%
    \caption{\textbf{Different augmentation sets.} 
    }
    \label{fig:aug_strength}
\vspace{-1.2cm}
\end{center}
\end{wrapfigure}

\paragraph{Stronger augmentations reduce memorization.}\label{app:aug_unitmem}
We also analyze how the training augmentation strength can impact the final encoder's \unitmem.
\reb{We use ColorJitter, HorizontalFlip, RandomGrayscale, and RandomResizedCrop as augmentations. Their strength is determined by the probability of applying them and their level of distortion. 
In \Cref{par:augmentations}, we present the exact parameters specified for each of them under different strengths.}
Our results in \Cref{fig:aug_strength} suggest that stronger augmentations yield lower per-unit memorization. 
These findings are in line with prior theoretical work on SSL~\citep{wang2021chaos} highlighting that SSL performs foremost the task of instance discrimination (\ie differentiating between individual images), but achieves clustering according to classes due to the augmentations:
with stronger augmentations, multiple data points' augmented views will look extremely similar (\eg the wheels of two different images of cars), such that they eventually activate the same unit.
Thereby, this unit memorizes individual data points less while units trained with weaker augmentations depend on and memorize individual data points more.
Note that we do not observe a strong dependency of our reported \unitmem on the concrete augmentation set used to calculate the metric (see \Cref{eq:augmentations}) as we show in \Cref{fig:aumentation_exp} in \Cref{app:unitmem}.
Yet, using the original set of training augmentations, as we do for our experiments, yields the strongest signal.

\begin{wrapfigure}{r}{0.3\textwidth}
\vskip \vskipintu
\vspace{0.55cm}
\begin{center}
\centerline{\includegraphics[width=\xintu\textwidth,trim={3cm 1.5cm 0 2.5cm}]{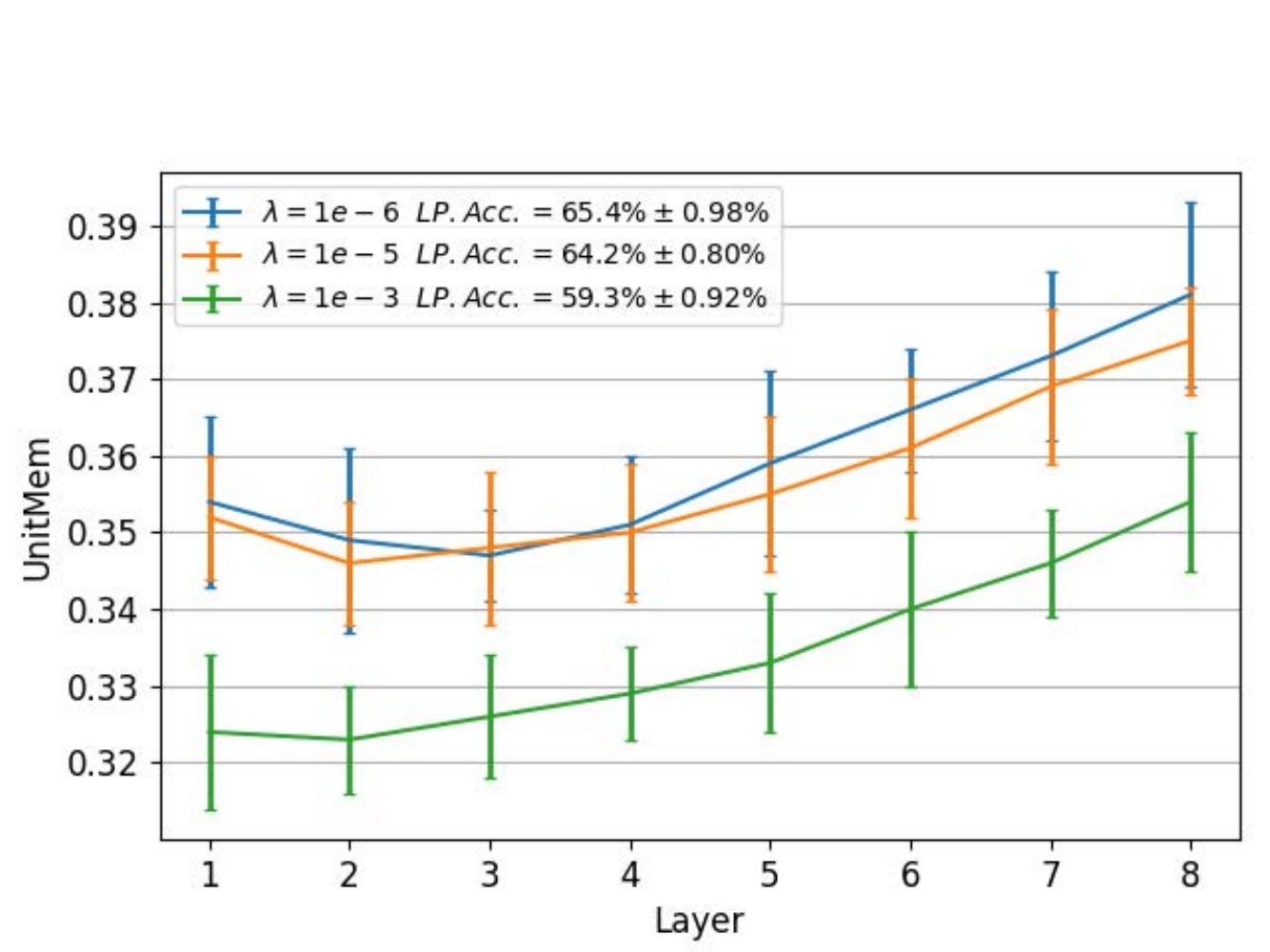}
    }%
    \caption{Different weight decay} 
    \label{fig:weight_decay}
\vspace{-1.2cm}
\end{center}
\end{wrapfigure}

\paragraph{Stronger weight decay reduces memorization.}

To analyze how training weight decay affects the final encoder's \unitmem, we train a ResNet9 using SimCLR on CIFAR10 with three different levels of weight decay. Our results in \Cref{fig:weight_decay} show that stronger weight decay yields lower memorization, yet also decreases linear probing accuracy.

\begin{wrapfigure}{r}{0.6\textwidth}
\vspace{-0.8cm}
    \centering
    \subfloat[SimCLR vs. DINO]{\includegraphics[scale=0.18]{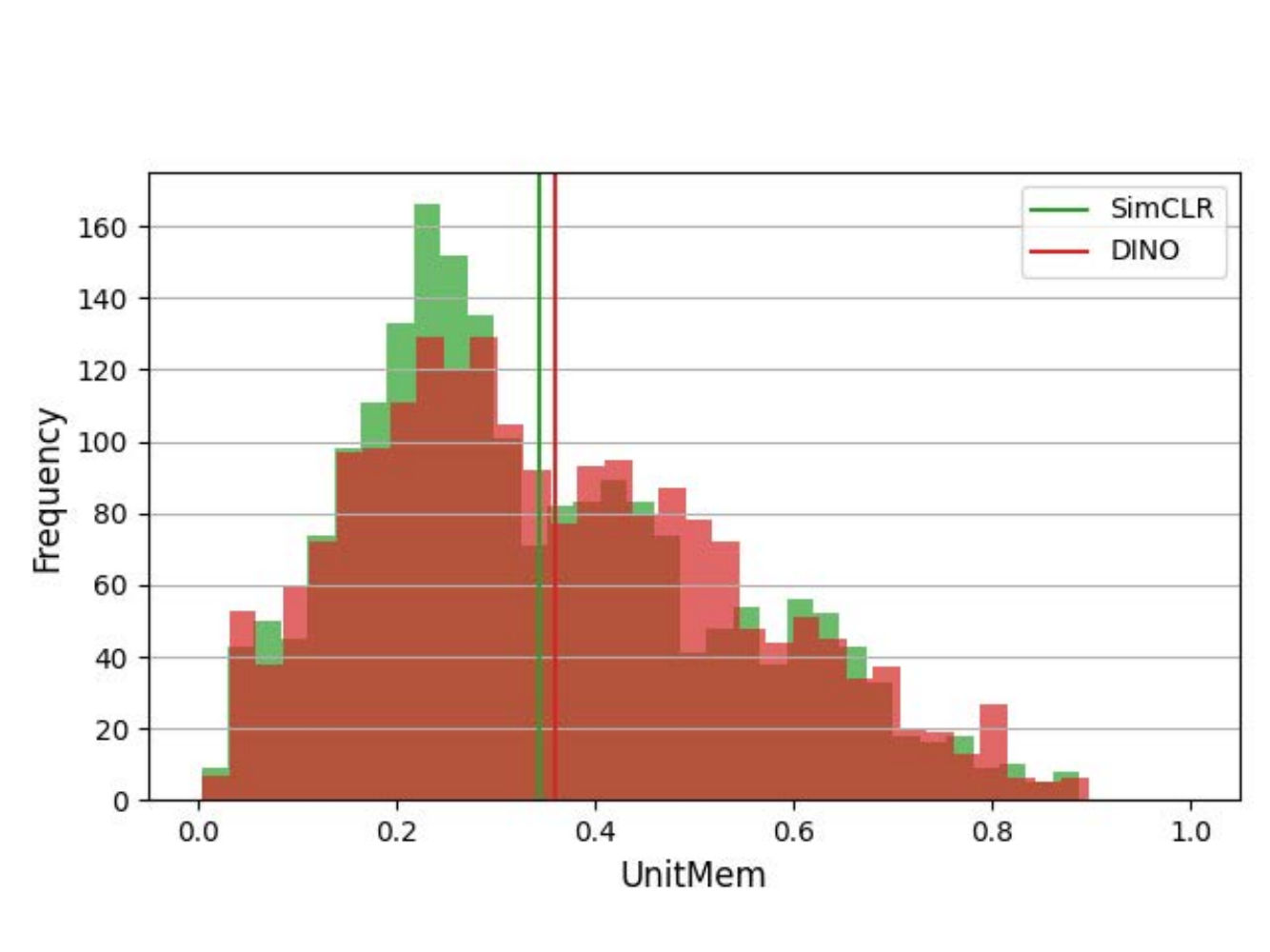}
    \vspace{-0.4cm}
    }
    \subfloat[DINO vs. MAE]{\includegraphics[scale=0.18]{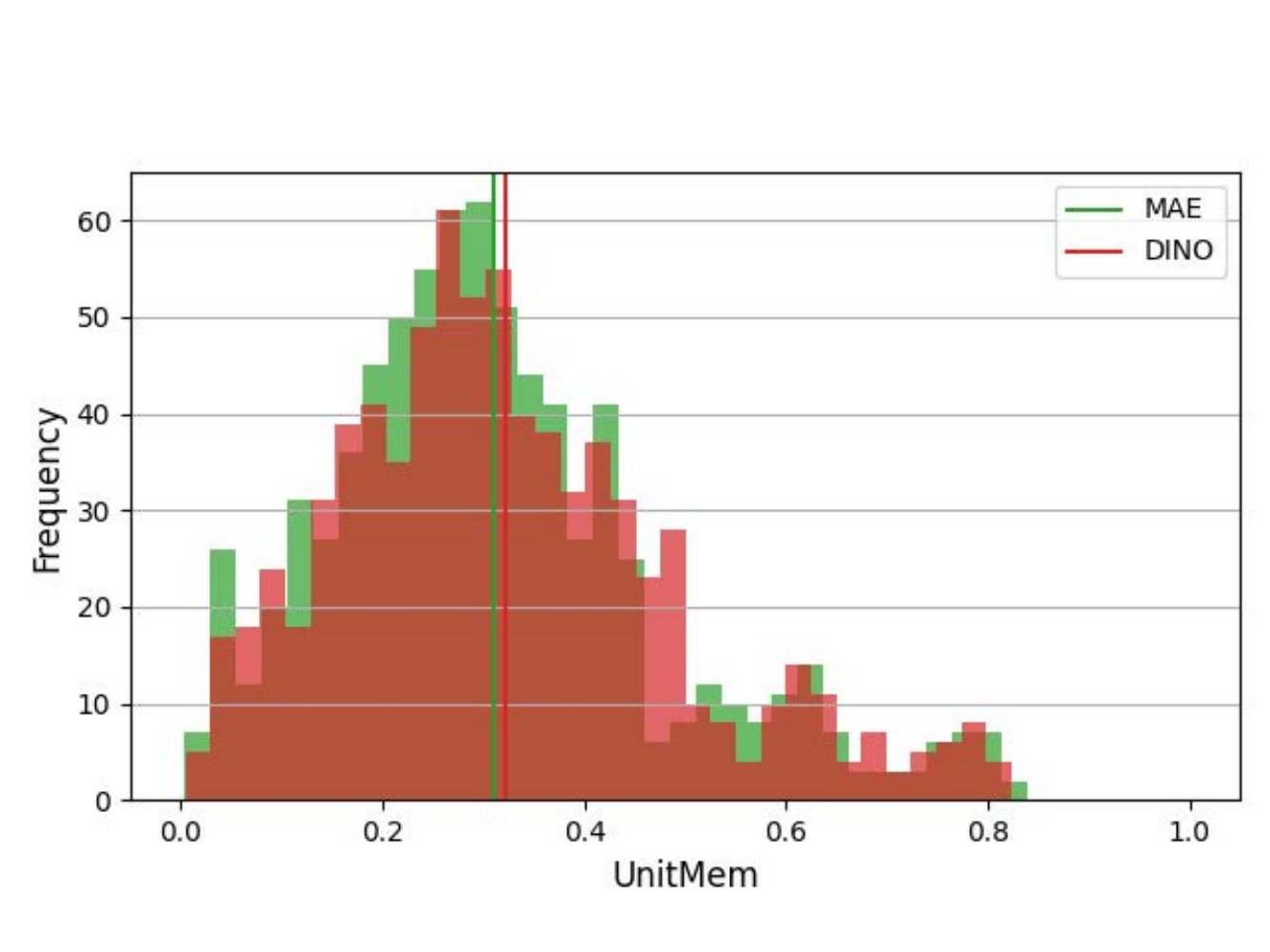}
    \vspace{-0.4cm}
    }%
    \caption{\textbf{Different SSL frameworks.} 
    }
    \label{fig:comparing_frameworks_unit}
\vspace{-0.4cm}
\end{wrapfigure}

\paragraph{Different SSL frameworks yield similar memorization pattern.}
We compare the \unitmem score between corresponding layers of a ResNet50 pre-trained on ImageNet with SimCLR and DINO, as well as for ViT-Base encoders pre-trained on ImageNet with DINO and MAE. 
We ensure that the number of epochs, batch sizes, training dataset sizes, and the resulting linear probing accuracies of the encoders are similar for direct comparability.
Our results in \Cref{fig:comparing_frameworks_unit} depict the \unitmem of the last convolutional layer of the ResNet50, and the final block's fully-connected layer in the ViT.
The plot indicates that the different SSL frameworks applied to the same architecture with the same dataset yield similar memorization pattern.

\addtolength{\tabcolsep}{0pt}
\begin{table}[h]
    \centering
    \tiny
     \caption{\textbf{The memorization in ViT occurs primarily in the deeper blocks and more in the fully-connected than attention layers.} 
    We present the results for ViT Tiny pre-trained on CIFAR10 using MAE.\\
    $\Delta \layermem^{ATT}_N = \layermem^{ATT}_N - \residual^{FC}_{N-1}$, 
    $\Delta \layermem^{FC}_N = \layermem^{FC}_N - \residual^{ATT}_{N}$, 
    $\Delta \blockmem _N = \residual^{FC}_N - \residual^{FC}_{N-1}$.}
    \vspace{0.2cm}
    \begin{tabular}{cccccccccc}
    \toprule 
    ViT Block & \multicolumn{3}{c}{\textbf{\textit{Attention Layer}}} && \multicolumn{3}{c}{\textbf{\textit{Fully-Connected Layer}}} \\ \cline{2-4} \cline{6-9}
    Number & \layermem & \deltamem &\residual && \layermem & \deltamem &\residual &\deltavit \\[0.0cm]  \midrule
         1 &0.006 &-     &0.007 &&0.020 & -    &0.022 &-\\
         2 &0.028 &0.006 &0.028 &&0.039 &0.011 &0.040 &0.018\\ 
         3 &0.046 &0.006 &0.047 &&0.060 &0.013 &0.061 &0.021\\
         4 &0.067 &0.006 &0.067 &&0.083 &0.017 &0.085 &0.024\\
         5 &0.092 &0.007 &0.091 &&0.105 &0.014 &0.106 &0.021\\
         6 &0.114 &0.008 &0.114 &&0.129 &0.015 &0.131 &0.025\\
         7 &0.140 &0.009 &0.139 &&0.155 &0.016 &0.156&0.025\\
         8 &0.164 &0.008 &0.164 &&0.182 &0.018 &0.182 &0.026\\
         9 &0.191 &0.009 &0.190 &&0.210 &0.020 &0.211 &0.029 \\
         10 &0.220 &0.009 &0.220 &&0.240 &0.020 &0.241 &0.030\\
         11 &0.249 &0.008 &0.249 &&0.271 &0.022 &0.271 &0.030\\
         12 &0.280 &0.009 &0.280 &&0.303 &0.023 &0.304 & 0.033\\
    \bottomrule
    \end{tabular}

    \label{tab:vit-layermem-whole}
\end{table}
\addtolength{\tabcolsep}{0pt}

\begin{table}[th]
    \centering
    \tiny    
    \caption{\textbf{Most vs Least Memorized Data Points}.
    We train a ResNet9 using SimCLR on CIFAR10 follwoing the setup by~\citep{wang2024memorization}. We then take the 50 most and 50 least memorized data points according to \sslmem and calculate the \unitmem over for the two sets of points.
    In the table, we report the average per-layer \unitmem of the two sets independently.
    We also perform a statistical $t$-test to find whether the \unitmem scores differ among most and least memorized data points.
    With $p<<0.05$, we are able to reject the null-hypothesis and find that the memorization according to \unitmem differs significantly between the most and least memorized data points.}
    \vspace{0.2cm}
    \begin{tabular}{cccc}
    \toprule 
             Layer & mean \unitmem & mean \unitmem & t-test \\
             Name & most memorized (10\% units) & least memorized (10\% units)  & p-value \\
    \midrule
         conv1& 0.507 & 0.235& 65.89/0.00 \\
         conv2-0& 0.501 & 0.231 & 66.25/0.00  \\
         conv2-1 & 0.503 &0.233 & 65.94/0.00  \\
         conv2-2 & 0.512 & 0.240 & 65.12/0.00 \\
         conv3 & 0.509 & 0.242 & 64.13/0.00 \\
         conv4-0 & 0.514 & 0.246 & 63.67/0.00  \\
         conv4-1 & 0.515 & 0.245 & 64.09/0.00 \\
         conv4-2 & 0.522 & 0.248 & 64.18/0.00 \\
        \bottomrule
    \end{tabular}
    
    \label{tab:resnet9_per_layer_our_memorization_score_t_test}
\end{table}

\begin{table}[t]
\centering
\tiny
\caption{
\textbf{Highly memorized data points align with most memorizing units.}
We select 10\% of the most memorizing units according to \unitmem in the last layer (\textit{conv-4-2}) of the ResNet9 encoder pre-trained on CIFAR10.
The 1st row represents the number of times a given data point was responsible for $\mu_{max}$, the 2nd row counts for how many daat points this applies. 
The last column shows that the highest memorized sample (\sslmem of 0.891) is responsible for the $\mu_{max}$ in the largest number of units (5). 
}
\vspace{0.2cm}
\begin{tabular}{cccccc}
\toprule 
Metric Used$\rightarrow$ & \multicolumn{5}{c}{\textbf{\textit{Average \sslmem Score}}} \\[0.05cm] \cline{2-6}
Frequency$\downarrow$ & 0.694 & 0.813 & 0.833 & 0.857 & 0.891 \\\midrule
\# of times Responsible for $\mu_{max}$ & 1 & 2 & 3 & 4 & 5 \\
\# of Samples & 10 & 2 & 1 & 1 & 1 \\ 
\bottomrule 
\end{tabular}

\label{tab:sslmem-unitmem-align}
\end{table}

\paragraph{Additional Verification of \unitmem.}
We present the additional verification of the \unitmem metric in \Cref{tab:1filter-appendix}.
Therein, we perform two additional experiments to the verification presented in \Cref{sub:unit_verification}.
First, we fine-tune the most memorizing unit and the inactive unit with 300 (instead of 1) data points from the test set (a).
We observe that the data points that experienced the highest memorization for the selected unit remains the highest memorized of the 300 data points.
Additionally, it experiences the highest memorization in the unit that used to be inactive.
Second, we fine-tune the most memorizing unit and the inactive unit with the most memorized data point, but with a batch-size of 300 were we duplicate the data point 300 times (b).
We observe that the effect of the fine-tuning on this point's memorization is far more expressed than when fine-tuning with 300 different data points.

\begin{table}[t]
\centering
\tiny
\caption{\textbf{Verification of the \unitmem metric for memorization in individual units.} 
The SSL model based on SimSiam with ResNet18 architecture and trained on CIFAR10 is fine-tuned on a single data point. 
We select two units with the highest and lowest \unitmem scores. The data point used for fine-tuning achieves $\mu_{max}$ in both units. The \unitmem score increases only for the two selected units while it remains unchanged for the remaining units.}
\vspace{0.2cm}
\begin{tabular}{ccccccc}
\toprule
Targeted & \multicolumn{6}{c}{\textbf{\textit{Number of Fine-Tuning Epochs}}} \\[0.05cm] \cline{2-7}
Unit & 0 & 10 & 50 & 200 & 500 & 1000 \\\midrule
Highest \unitmem & 0.754 & 0.761 & 0.792   & 0.814 & 0.824 & 0.826\\
Lowest \unitmem & 0 & 0 & 0 & 0.0008 & 0.0021 & 0.0109 \\ 
\bottomrule 
\end{tabular}
\label{tab:1filter}
\vspace{0em}
\end{table}

\paragraph{Additional Verification of \unitmem.}
\label{app:pruning}
In \Cref{tab:1filter}, we prune, \ie zero out neurons according to their level of memorization. Our results indicate that by pruning the most memorizing neurons, we cause the highest drop in downstream performance.

\begin{table}[h]
\centering
\caption{\textbf{The $\mu_{max}$ and $\mu_{min}$ after fine-tuning for different number of epochs.}}
\vspace{0.2cm}
\begin{subtable}{0.9\textwidth}
\centering
\small
\caption{\textbf{The $\mu_{max}$ and $\mu_{min}$ after fine-tuning for different numbers of epochs.} This is fine-tuned with 300 data samples from the test dataset. The samples were not seen during the initial training of the encoder, thus only a single filter is affected by them.}
\begin{tabular}{ccccccc}
\toprule

trained nodes & original & 10 epoch &  50 epoch & 200 epoch & 500 epoch & 1000 epoch \\ \midrule
        
\begin{tabular}[c]{@{}c@{}}Most Mem\\ filter $77^{th}$\end{tabular}  & 0.754 & 0.766 & 0.809  & 0.819 & 0.826 & 0.828\\
\begin{tabular}[c]{@{}c@{}}Least Mem\\ filter $459^{th}$\end{tabular} &  0 & 0 & 0.011 & 0.038 & 0.046 & 0.051 \\ 
\bottomrule 
\end{tabular}
\label{tab:1filter_300}
\end{subtable}

\begin{subtable}{0.9\textwidth}
\centering
\small
\caption{\textbf{The $\mu_{max}$ and $\mu_{min}$ after fine-tuning for different number of epochs} This is fine-tuned with only highest $\mu_{max}$ samples while 300 duplication from training datasets.}
\begin{tabular}{ccccccc}
\toprule
trained nodes & original & 10 epoch &  50 epoch & 200 epoch & 500 epoch & 1000 epoch \\ \midrule       
\begin{tabular}[c]{@{}c@{}}Most Mem\\ filter $77^{th}$\end{tabular}  & 0.754 & 0.798 & 0.846  & 0.857 & 0.861 & 0.862\\
\begin{tabular}[c]{@{}c@{}}Least Mem\\ filter $459^{th}$\end{tabular} & 0 & 0.039 & 0.065 & 0.079 & 0.081 & 0.081 \\
\bottomrule 
\end{tabular}
\label{tab:1filter_1_300}
\end{subtable}
\label{tab:1filter-appendix}
\end{table}

\subsection{\unitmem Measures Memorization of Individual Data Points}
\label{appendix:unit-mem-class-paragraphs}
\reb{
To highlight that \unitmem reports memorization of individual data points rather than the a unit's ability to recognize class-wide concepts, we designed an additional experiment.
For the experiment, we rely on the class concept of "wheel" as an example. In the STL10 dataset, three classes have a concept wheel, namely Truck, Plane, Car.
If \unitmem was to report simply a unit's sensitivity to concepts of different classes (rather than individual data points), we would see a drop in \unitmem as we increase the percentage of data points with the concept wheel in the batch used to compute the metric.
This is because then all data points should equally activate the unit, resulting in low memorization according to \Cref{eq:selectivity}.
}

\reb{
We perform the experiment in \Cref{tab:unit_mem_classes_unique}  with 1000 data points chosen from different classes, namely 1) all classes (here 30\% of the data points have wheels), 2) the classes Truck, Plane, Car (close to 100\% of the samples now have the concept of wheels), and 3) purely the class car (close to 100\% wheels). In 2) and 3), close to 100\% of the samples now have the concept of wheels. Thus, if the units were responsible for the concept wheel, they would have a very high activation over all samples and the reported \unitmem should be very low. However, in our results, we see that we do have units with very high \unitmem. These can, in turn not be the units for the class-concept wheel, but must be units that focus on individual characteristics of the individual training images. This means that there must be unique features from the individual images that are still memorized that go beyond the concepts that are the same within a class.
}

\subsection{Additional Insights into \layermem}
\label{appendix:layermem-insigths}
\label{appendix:per-layer-memorization-resnets}

\paragraph{\layermem is not sensitive to the size and composition of the batch.}
\reb{In our ablation study in , we show that \layermem is not sensitive to the size and composition of batch it is computed on. The results can be found in \Cref{tab:layermem_different_candidate_numbers}, where report the \layermem measured for different number of candidate data points.
We pre-trained a ResNet9 using SimCLR on CIFAR10 and determined \layermem on batches of different sizes. For each batch size, we use 3 independent seeds (\ie different batch compositions) and report the average \layermem score and its standard deviation. The results show that the reported \layermem score is, indeed, similar across all setups. This indicates \layermem’s insensitivity to the choice of the batch used to compute it.
}

\paragraph{Full Results with Memorization Scores over all Layers.}
We present the \layermem score for ResNet18 in \Cref{tab:resnet18_per_layer_our_memorization_score}, ResNet34 in \Cref{tab:resnet34_per_layer_our_memorization_score}, and ResNet50 in \Cref{tab:resnet50_per_layer_our_memorization_score}, all trained on CIFAR10 and using the SimCLR framework.

\begin{table}[h]
    \centering
    \tiny
    \caption{\textbf{Full results ResNet18.} We depict our \layermem of the final trained model (at the end of training with CIFAR10, ResNet18 with SimCLR).}
    \vspace{0.2cm}
    \begin{tabular}{cc}
    \toprule 
         Layer& \layermem\\
    \midrule
         conv1 & 0.074 $\pm$ 0.010\\
         max pool & 0.092 $\pm$ 0.007\\
         conv2-1 & 0.101 $\pm$ 0.012\\
         conv2-2 & 0.110 $\pm$ 0.006\\
         conv2-3 & 0.123 $\pm$ 0.013\\
         conv2-4 & 0.134 $\pm$ 0.010\\
         conv3-1& 0.146 $\pm$ 0.008\\
         conv3-2& 0.155 $\pm$ 0.013\\
         conv3-3& 0.166 $\pm$ 0.011\\
         conv3-4& 0.183 $\pm$ 0.007\\
         conv4-1& 0.193 $\pm$ 0.006\\
         conv4-2& 0.206 $\pm$ 0.009\\
         conv4-3& 0.220 $\pm$ 0.011\\
         conv4-4& 0.239 $\pm$ 0.010\\
         conv5-1& 0.246 $\pm$ 0.014\\
         conv5-2& 0.257 $\pm$ 0.007\\
         conv5-3& 0.272 $\pm$ 0.011\\
         conv5-4& 0.295 $\pm$ 0.012\\
         averge-pool& 0.266 $\pm$ 0.009\\
         fully-connected& 0.224 $\pm$ 0.010\\
         softmax& 0.207 $\pm$ 0.007\\
        \bottomrule
    \end{tabular}
    \label{tab:resnet18_per_layer_our_memorization_score}
\end{table}

\begin{table}[h]
    \centering
    \tiny
    \caption{\textbf{Full results ResNet34.} We depict our \layermem  of the final trained model (at the end of training with CIFAR10, ResNet34 with SimCLR).}
            \vspace{0.2cm}
    \begin{tabular}{cc}
    \toprule 
         Layer& \layermem \\
    \midrule
         conv1 & 0.037 $\pm$ 0.008\\
         max pool & 0.069 $\pm$ 0.013\\
         conv2-1 & 0.078 $\pm$ 0.008\\
         conv2-2 & 0.083 $\pm$ 0.007\\
         conv2-3 & 0.091 $\pm$ 0.012\\
         conv2-4 & 0.096 $\pm$ 0.009\\
         conv2-5 & 0.107 $\pm$ 0.016\\
         conv2-6 & 0.115 $\pm$ 0.010\\
         conv3-1& 0.124 $\pm$ 0.011\\
         conv3-2& 0.128 $\pm$ 0.013\\
         conv3-3& 0.131 $\pm$ 0.007\\
         conv3-4& 0.138 $\pm$ 0.008\\
         conv3-5& 0.143 $\pm$ 0.013\\
         conv3-6& 0.149 $\pm$ 0.015\\
         conv3-7& 0.157 $\pm$ 0.013\\
         conv3-8& 0.166 $\pm$ 0.009\\
         conv4-1& 0.172 $\pm$ 0.006\\
         conv4-2& 0.178 $\pm$ 0.010\\
         conv4-3& 0.181 $\pm$ 0.012\\
         conv4-4& 0.186 $\pm$ 0.008\\
         conv4-5& 0.194 $\pm$ 0.013\\
         conv4-6& 0.201 $\pm$ 0.007\\
         conv4-7& 0.205 $\pm$ 0.009\\
         conv4-8& 0.211 $\pm$ 0.011\\
         conv4-9& 0.218 $\pm$ 0.006\\
         conv4-10& 0.227 $\pm$ 0.012\\
         conv4-11& 0.235 $\pm$ 0.010\\
         conv4-12& 0.246 $\pm$ 0.007\\
         conv5-1& 0.257 $\pm$ 0.011\\
         conv5-2& 0.264 $\pm$ 0.014\\
         conv5-3& 0.273 $\pm$ 0.008\\
         conv5-4& 0.285 $\pm$ 0.012\\
         conv5-5& 0.299 $\pm$ 0.011\\
         conv5-6& 0.313 $\pm$ 0.015\\
         averge-pool& 0.297 $\pm$ 0.009\\
         fully-connected& 0.241 $\pm$ 0.013\\
         softmax& 0.233 $\pm$ 0.006\\
        \bottomrule
    \end{tabular}
    \label{tab:resnet34_per_layer_our_memorization_score}
\end{table}

\begin{table}[h]
    \centering
    \tiny
        \caption{\textbf{Full results ResNet50.} We depict our \layermem  of the final trained model (at the end of training with CIFAR10, ResNet50 with SimCLR).}
            \vspace{0.2cm}
    \begin{tabular}{cc}
    \toprule 
         Layer& \layermem \\
    \midrule
        conv1 & 0.046 $\pm$ 0.006\\ 
         max pool & 0.066 $\pm$ 0.012\\
         conv2-1 & 0.071 $\pm$ 0.008\\
         conv2-2 & 0.068 $\pm$ 0.013\\
         conv2-3 & 0.073 $\pm$ 0.012\\
         conv2-4 & 0.079 $\pm$ 0.015\\
         conv2-5 & 0.082 $\pm$ 0.014\\
         conv2-6 & 0.083 $\pm$ 0.010\\
         conv2-7 & 0.088 $\pm$ 0.007\\
         conv2-8 & 0.094 $\pm$ 0.011\\
         conv2-9 & 0.103 $\pm$ 0.014\\
         conv3-1& 0.109 $\pm$ 0.010\\
         conv3-2& 0.112 $\pm$ 0.012\\
         conv3-3& 0.118 $\pm$ 0.009\\
         conv3-4& 0.123 $\pm$ 0.007\\
         conv3-5& 0.127 $\pm$ 0.010\\
         conv3-6& 0.133 $\pm$ 0.011\\
         conv3-7& 0.136 $\pm$ 0.013\\
         conv3-8& 0.140 $\pm$ 0.008\\
         conv3-9& 0.144 $\pm$ 0.005\\
         conv3-10& 0.149 $\pm$ 0.008\\
         conv3-11& 0.156 $\pm$ 0.011\\
         conv3-12& 0.165 $\pm$ 0.007\\
         conv4-1& 0.168 $\pm$ 0.012\\
         conv4-2& 0.175 $\pm$ 0.010\\
         conv4-3& 0.181 $\pm$ 0.006\\
         conv4-4& 0.187 $\pm$ 0.009\\
         conv4-5& 0.192 $\pm$ 0.008\\
         conv4-6& 0.198 $\pm$ 0.014\\
         conv4-7& 0.204 $\pm$ 0.010\\
         conv4-8& 0.211 $\pm$ 0.008\\
         conv4-9& 0.217 $\pm$ 0.011\\
         conv4-10& 0.225 $\pm$ 0.005\\
         conv4-11& 0.231 $\pm$ 0.015\\
         conv4-12& 0.235 $\pm$ 0.011\\
         conv4-13& 0.241 $\pm$ 0.012\\
         conv4-14& 0.248 $\pm$ 0.009\\
         conv4-15& 0.253 $\pm$ 0.011\\
         conv4-16& 0.262 $\pm$ 0.016\\
         conv4-17& 0.268 $\pm$ 0.012\\
         conv4-18& 0.279 $\pm$ 0.011\\
         conv5-1& 0.292  $\pm$ 0.008\\
         conv5-2& 0.293  $\pm$ 0.005\\
         conv5-3& 0.298  $\pm$ 0.012\\
         conv5-4& 0.308  $\pm$ 0.010\\
         conv5-5& 0.316  $\pm$ 0.014\\
         conv5-6& 0.315  $\pm$ 0.012\\
         conv5-7& 0.326  $\pm$ 0.007\\
         conv5-8& 0.327  $\pm$ 0.011\\
         conv5-9& 0.335 $\pm$ 0.013\\
         averge-pool& 0.328 $\pm$ 0.007\\
         fully-connected& 0.266 $\pm$ 0.014\\
         softmax& 0.245 $\pm$ 0.010\\
        \bottomrule
    \end{tabular}
    \label{tab:resnet50_per_layer_our_memorization_score}
\end{table}

We show the further breakdown of the memorization within the layers in \Cref{app:resnet9_per_layer_our_memorization_score}. We observe that the batch normalization layers (denoted as BN) together with the MaxPool layers have a negligible impact on memorization and most of the memorization in each layer is due to the convolutional operations. This is due to the much larger number of parameters in the convolutional filters than in the batch normalization layers and no additional parameters in the MaxPool layers, as shown in \Cref{tab:resnet9_architecture}.
However, the memorization reported per convolutional layer is not correlated with the number of parameters of the layer. For instance, our \deltamem reports the highest memorization for the 6-th layer, while layers 7 and 8 have each twice as many parameters, see \Cref{tab:resnet9_per_layer_our_memorization_score}.

\addtolength{\tabcolsep}{-3.5pt} 
\begin{table}[t]
    \centering
    \tiny
    \caption{
    \textbf{Layer-based Memorization Scores.} 
    We present the layer-wise memorization of an SSL encoder pretrained on CIFAR10 using ResNet9 with SimCLR. The 1st column represents the IDs of convolutional layers and the 2nd column shows the name of the layers. Residual$N$ denotes that the residual connection comes from the previous $N$-th convolutional layer. We report \layermem across the 100 randomly chosen training data points, their \deltamem (denoted as $\Delta$\texttt{LM}), followed by \layermem for only the Top 50 memorized data points, their \deltamem (denoted as $\Delta$Top50), and \layermem for only the Least 50 memorized data points.
    The projection \textit{head} layer (denoted as head) is used only for training.
    }
            \vspace{0.2cm}
    \begin{tabular}{ccccccc}
    \toprule 
    ID & Name& \layermem & $\Delta$\texttt{LM} & \layermem Top50 & $\Delta$Top50& \layermem Least50 \\
    \midrule
         1 & Conv1&0.091&-&0.144&-&0.003\\
         - & BN1 &0.091&0.000&0.144&0&0.004\\
         - & MaxPool &0.097&0.006&0.158&0.014&0.004\\
         2 & Conv2-0&0.123&0.026&0.225&0.067&0.012\\
         - & BN2-0 &0.124&0.001&0.225&0&0.012\\
         - & MaxPool&0.128&0.004&0.236&0.011&0.013\\
         3 & Conv2-1&0.154&0.026&0.308&0.072&0.022\\
         4 & Conv2-2&0.183&0.029&0.402&0.094&0.031 \\
         - & Residual2&0.185&0.002&0.403&0.01&0.041 \\
         5 & Conv3 &0.212 &0.027& 0.479& 0.076&0.051\\
         - & BN3 &0.211&-0.001&0.480&0.001&0.051\\
         - &MaxPool&0.215 &0.004&0.486&0.006&0.050\\
         6 & Conv4-0&0.246&0.031&0.599& 0.113&0.061\\
         - & BN4-0 &0.244&-0.002&0.600&0.001&0.060\\
         - &MaxPool&0.247&0.003&0.603&0.003&0.061\\
         7 & Conv4-1&0.276&0.029&0.697& 0.094&0.071\\
         8 & Conv4-2&0.308&0.032&0.817&0.120 &0.073\\
         - & Residual6&0.311& 0.003& 0.817&0&0.086\\
         \midrule
         - & head &0.319&0.008&0.819&0.002&0.097\\
         \midrule
         - & MaxPool &0.318&-0.001&0.819&0&0.096 \\
         - & FC&0.192&-0.126&0.409&-0.410&0.071\\
        \bottomrule
    \end{tabular}
    \label{app:resnet9_per_layer_our_memorization_score}
\end{table}
\addtolength{\tabcolsep}{3.5pt}

\begin{table}[t]
\centering
\tiny
\caption{\reb{\textbf{\layermem is not sensitive to the number of samples used for its calculation.} We pre-train a ResNet9 using SimCLR on CIFAR10 and determined \layermem on batches of different sizes. For each batch size, we use three independent seeds (\ie different batch compositions) and report the average 
\layermem score and its standard deviation. The results show that the reported \layermem score is, indeed, similar across all setups. This indicates \layermem’s insensitivity to the choice of the batch used to compute it.}}
\vspace{0.2cm}
\begin{tabular}{ccccc}
\toprule
\textbf{Layer}& \textbf{100  samples}& \textbf{500  samples}& \textbf{1000  samples}& \textbf{5000  samples}\\
\midrule
1                                          & 0.092 $\pm$ 8e-4   & 0.093 $\pm$ 7e-4   & 0.089 $\pm$ 7e-4    & 0.091 $\pm$ 8e-4    \\
 2                                          & 0.122 $\pm$ 9e-4   & 0.124 $\pm$ 1e-3   & 0.122 $\pm$ 8e-4    & 0.121 $\pm$ 7e-4    \\
 3                                          & 0.150 $\pm$ 1e-3   & 0.154 $\pm$ 9e-4   & 0.151 $\pm$ 8e-4    & 0.152 $\pm$ 6e-4    \\
 4                                          & 0.181 $\pm$ 1e-3   & 0.182 $\pm$ 8e-4   & 0.180 $\pm$ 8e-4    & 0.181 $\pm$ 8e-4    \\
Res2                            & 0.184 $\pm$ 9e-4   & 0.185 $\pm$ 8e-4   & 0.183 $\pm$ 9e-4    & 0.184 $\pm$ 6e-4    \\
 5                                          & 0.213 $\pm$ 8e-4   & 0.213 $\pm$ 8e-4   & 0.212 $\pm$ 7e-4    & 0.212 $\pm$ 8e-4    \\
 6                                          & 0.246 $\pm$ 1e-3   & 0.249 $\pm$ 7e-4   & 0.247 $\pm$ 8e-4    & 0.245 $\pm$ 9e-4    \\
 7                                          & 0.277 $\pm$ 7e-4   & 0.281 $\pm$ 9e-4   & 0.277 $\pm$ 8e-4    & 0.276 $\pm$ 4e-4    \\
 8                                          & 0.309 $\pm$ 9e-4   & 0.314 $\pm$ 8e-4   & 0.310 $\pm$ 7e-4    & 0.307 $\pm$ 7e-4    \\
Res6                            & 0.310 $\pm$ 1e-3   & 0.316 $\pm$ 1e-3   & 0.313 $\pm$ 8e-4    & 0.309 $\pm$ 9e-4    \\
\bottomrule
\end{tabular}
\label{tab:layermem_different_candidate_numbers}
\end{table}

\reb{
\paragraph{Ablation on \layermem's sensitivity.}
We perform an additional ablation to show that \layermem is not sensitive to the number of samples in the batch used to compute it or the composition of the batch, \ie which samples are chosen.
Our results in \Cref{tab:layermem_different_candidate_numbers} highlight that over different batches with 100, 500, 1000, and 5000 samples, the observed \layermem scores are alike. 
This indicates \layermem’s insensitivity to the choice of the batch used to compute it.
}

\begin{table}
\centering
\tiny
\caption{\reb{\textbf{Comparing the impact of memorization on downstream generalization between SSL and SL.} We train a ResNet9 on CIFAR100 with SSL (pretrained on CIFAR100 using SimCLR and SL (cross-entropy loss, trained until convergence). For the SL model, we remove the classification layer to turn it into an encoder. Then, we report linear probing accuracies on multiple downstream tasks in }}
\vspace{0.2cm}
\begin{tabular}{ccccc}
\toprule
\textbf{Encoder}                                                        & \textbf{CIFAR100}        & \textbf{CIFAR10}         & \textbf{STL10}           & \textbf{SVHN}            \\
\hline
SSL                                    & 65.4\% ± 0.98\% & 57.6\% ± 0.87\% & 48.7\% ± 0.98\% & 59.2\% ± 0.76\% \\
SL (trained until convergence on CIFAR100, last layer removed) & 66.1\% ± 1.12\% & 56.7\% ± 0.83\% & 46.1\% ± 1.04\% & 58.6\% ± 0.82\% \\
\bottomrule
\end{tabular}
\label{tab:ssl_mem_vs_sl_mem}
\end{table}

\subsection{Memorization in SL vs. SSL}
\label{app:ssl_mem_vs_sl_mem}
\reb{We conducted an additional experiment where we trained a ResNet9 on CIFAR100 with SSL (SimCLR) and SL (cross-entropy loss). For the SL model, we remove the classification layer to turn it into an encoder. Then, we report linear probing accuracies on multiple downstream tasks in \Cref{tab:ssl_mem_vs_sl_mem}.
Our results highlight that the SL pretrained encoders exhibit a significantly higher downstream accuracy on their pretraining dataset than the SSL encoder. We assume that this is because of the class memorization. In contrast, the SL pretrained encoders perform significantly worse on other datasets than the SSL pretrained encoders since they might overfit the representations to their classes rather than provide more general (instance-based) representations as the SSL encoders.
Additionally, we note that prior work has shown that the MAE encoder provides the highest performance when a few last layers are fine-tuned. The results in the original MAE paper in Figure 9~\citep{mae} indicate that fine-tuning a few last layers/blocks (e.g., 4 or 6 blocks out of 24 in ViT-Large) can achieve accuracy close to full fine-tuning (when all 24 blocks are fine-tuned). This is in line with our observation that the difference between UnitMem and ClassMem is the highest in the few last layers/blocks. Thus, fine-tuning only these last layers/blocks suffices for good downstream performance.
}

\subsection{Visualization for Variability and Consistency of Memorization cross Different Layers.}\label{app:vi_consist}
\begin{figure}[t]
    \centering
    \includegraphics[width=0.9\linewidth]{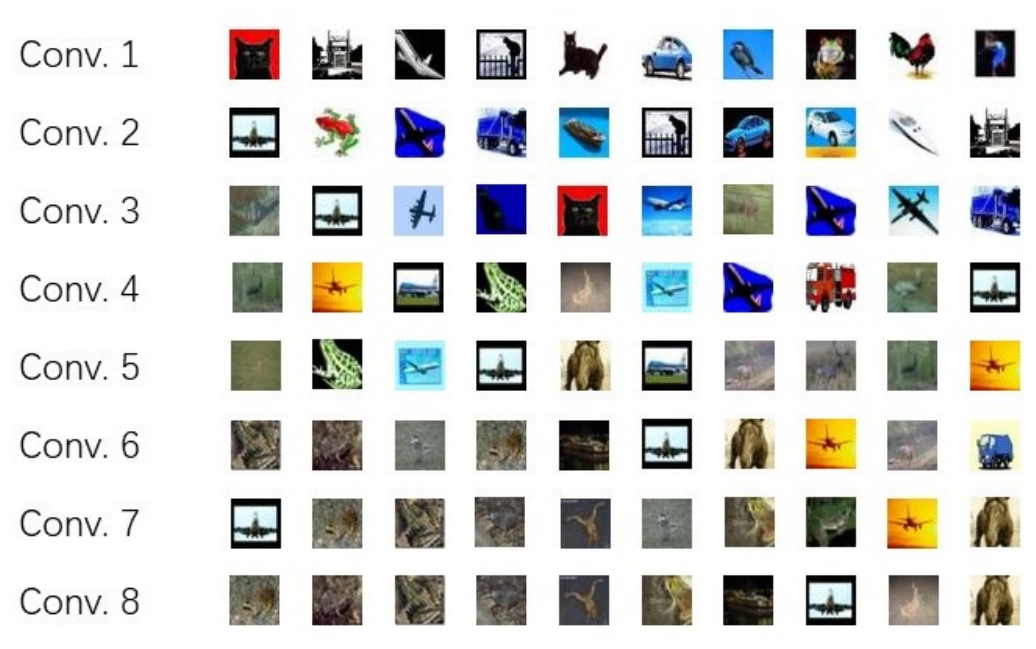}
    \vspace{0em}
    \caption{\textbf{The most memorized samples per layer according to \layermem.}}
    \label{fig:consistency}
    \vspace{0em}
\end{figure}

We present the top 10 most memorized samples of each layer for the ResNet9 vision encoder trained with the CIFAR10 dataset in \Cref{fig:consistency}. The results show that the overlap within the top 10 most memorized samples between adjacent layers is usually high but decreases the further the layers are separated. This aligns with the results of overlap rate and Kendall's Tau test reported in \Cref{tab:consistency}. 

\subsection{Layer-based Memorization Across Different SSL Frameworks and Datasets}
\reb{
We present the full results for the \Cref{tab:layermem-fraemworks}, which show that the layer-based memorization is similar across encoders trained with different SSL frameworks.
The results for the ResNet50 architecture trained with SimCLR and DINO using the ImageNet dataset are presented in \Cref{tab:resnet50-simclr-dino-full}, and the results for the ViT-Base architecture trained with MAE and DINO using the ImageNet dataset are shown in \Cref{tab:vit-base-mae-dino-full}.
}

\subsection{Verification of Layer-Based Memorization}
\label{app:layermemverification}

\addtolength{\tabcolsep}{-3.5 pt}
\begin{table}[t]
\label{tab:resnet9_replace}
    \tiny
    \centering
    \caption{\textbf{Replacing the most/least memorized layers according to $\Delta$\layermem causes the most/least changes in downstream performance.} We study the effect of replacing layers of the ResNet9 encoder trained on CIFAR10 with layers from another ResNet9 encoder trained on STL10 and report the linear probing accuracy on the CIFAR10 and STL10 test sets. %\franzi{The layer naming of the table is no longer consistend with the figures.}
    \reb{Results for the impact of replacing any combination of 1, 2, and 3 layers on downstream accuracy are shown in \Cref{appendix:replace-1-2-3-layers}.}
    }
            \vspace{0.2cm}
    \begin{tabular}{cccc}
    \toprule
    Replacement Criteria & Replaced Layer(s)&  CIFAR10 &  STL10 \\ \midrule
    \textit{None (Baseline)} & \textit{None} & 69.08\% $\pm$ 1.05\%   &  17.81\% $\pm$ 0.92\%  \\
    \hdashline 
    \textbf{Most Memorized} $\Delta$\layermem  & \textbf{4 6 8} & \textbf{36.59\% $\pm$ 1.13\%} & \textbf{32.33\% $\pm$ 0.88\%} \\ 
    Most Memorized \layermem & 6 7 8 & 39.07\% $\pm$ 1.05\% & 29.82\% $\pm$ 0.91\%\\
    Random    & 4 5 7   &43.22\% $\pm$ 1.08\%& 25.89\% $\pm$ 0.93\%\\
    Least Memorized \layermem & 2 3 4 & 49.95\% $\pm$ 1.21\% & 24.71\% $\pm$ 0.99\%\\
    \textbf{Least Memorized} $\Delta$\layermem & \textbf{2 3 5} & \textbf{59.14\% $\pm$ 0.91\%} & \textbf{23.10\% $\pm$ 1.06\%} \\ 
    \bottomrule 
    \end{tabular}
    \label{tab:resnet50_replace_stl10}
\end{table}
\addtolength{\tabcolsep}{3.5 pt}

To analyze whether our \layermem metric and its $\Delta$ variant indeed localize memorization correctly, we first replace different layers of an encoder and then compute linear probing accuracy on various downstream tasks.
Since prior work shows that memorization in SSL is required for downstream generalization~\citep{wang2024memorization}, we expect the highest performance drop when replacing the layers identified as most memorizing.
We verify this hypothesis and train a ResNet9 encoder $f_1$ on the CIFAR10 dataset and compute the \layermem and \deltamem scores per layer. 
Then, we select the three most memorized, random, and least memorized layers and replace them with the corresponding layers from a ResNet9 trained on STL10 ($f_2$). 
Our results in \Cref{tab:resnet50_replace_stl10} show that the highest linear probing accuracy drop on the CIFAR10 test set for $f_1$ is caused by replacing the three most memorized layers from $f_1$ according to the \deltamem score. 
The second biggest drop is observed when replacing according to \layermem, highlighting that indeed our \layermem metric and its $\Delta$ variant identify the most crucial layers in SSL encoders for memorization.
Surprisingly, the replacement of the layers in $f_1$ with the corresponding layers from $f_2$ causes the biggest simultaneous increase in the downstream accuracy for the STL10 dataset. 
We observe the same trends when $f_2$ is trained on SVHN \reb{(\Cref{tab:resnet9_replace_svhn})}, for replacing single layers in ResNet9 \reb{(\Cref{tab:resnet9_replace_onelayer})}, and replacing whole blocks in ResNet50 \reb{(\Cref{tab:resnet50_replace})} instead of only individual layers  as we present in \Cref{appendix:replace-blocks-resnet50}.
The above analysis verifies that the \layermem score and its $\Delta$ variant identify the most crucial layers in SSL encoders. They further strengthen the claims that memorization is required for generalization~\citep{feldman2020does,feldman2020neural,wang2024memorization}.

\paragraph{Replacing layers in ResNet9 for SVHN.} 
\label{appendix:replace-layers-resnet9-svhn}
In \Cref{tab:resnet9_replace_svhn}, we show the effect of replacing the most and least vs random layers of a CIFAR10 trained ResNet9 on the downstream performance.
We replace the layers with the corresponding ones from a ResNet9 encoder trained on SVHN.

\addtolength{\tabcolsep}{-1.8 pt}
\begin{table*}[t]

\label{tab:resnet9_replace_both}
        \scriptsize 
            \caption{\textbf{Evaluating the effect of replacing layers of the ResNet9 encoder pre-trained on CIFAR10 with layers from ResNet9 pre-trained on STL10}. We report the linear probing accuracy of ResNet9 with the replaced layers and tested on the CIFAR10, STL10 test sets.
    }
    \vspace{0.2cm}
\begin{subtable}{0.9\textwidth}
\centering
\caption{CIFAR10 \& STL10}
\begin{tabular}{cccc}
\toprule

Replaced Criterium & Replaced Layer(s)&  CIFAR10 &  STL10 \\ \midrule
        
\textit{None (Baseline)} & \textit{None} & 69.08\% $\pm$ 1.05\%   &  17.81\% $\pm$ 0.92\%  \\
Most Memorized (delta)  & 4 6 8 & 36.59\% $\pm$ 1.13\% &32.33\% $\pm$ 0.88\%\\ 
Most Memorized (absolute) & 6 7 8 & 39.07\% $\pm$ 1.05\% & 29.82\% $\pm$ 0.91\%\\
Random    & 4 5 7   &43.22\% $\pm$ 1.08\%& 25.89\% $\pm$ 0.93\%\\
Least Memorized (delta) & 2 3 5 & 59.14\% $\pm$ 0.91\%& 23.10\% $\pm$ 1.06\%\\ 
Least Memorized (absolute) & 2 3 4 & 49.95\% $\pm$ 1.21\% & 24.71\% $\pm$ 0.99\%\\
\bottomrule 
\end{tabular}
\label{tab:resnet9_replace_stl10}
\end{subtable}

\begin{subtable}{0.9\textwidth}
\centering
\vspace{0.2cm}
\caption{CIFAR10 \& SVHN}
\begin{tabular}{cccccc}
\toprule
Replaced Criterium & Replaced Layer(s)&  CIFAR10 & SVHN \\ \midrule
        
\textit{None (Baseline)} & \textit{None} & 69.08\% $\pm$ 1.05\%   &  19.33\% $\pm$ 0.65\%  \\
Most Memorized (delta) & 4 6 8 & 
33.07\% $\pm$ 1.51\% & 34.05\% $\pm$ 1.01\%\\  
Most Memorized (absolute) & 6 7 8 & 34.97\% $\pm$ 0.84\% & 31.87\% $\pm$ 1.21\%\\
Random    &  4 5 7 & 39.28\% $\pm$ 0.74\%& 26.04\% $\pm$ 0.82\%\\
Least Memorized (delta) & 2 3 5 & 52.81\% $\pm$ 1.03\%& 21.05\% $\pm$ 0.89\%\\
Least Memorized (absolute) & 2 3 4 & 45.39\% $\pm$ 1.10\% & 24.66\% $\pm$ 0.57\%\\
\bottomrule 
\end{tabular}

\label{tab:resnet9_replace_svhn}
\end{subtable}

\end{table*}
\addtolength{\tabcolsep}{1.8 pt}

\addtolength{\tabcolsep}{-1.8 pt}
\begin{table*}[t]
\label{tab:resnet9_replace_onelayer}
        \scriptsize 
         \caption{\textbf{Evaluating the effect of replacing layers of the ResNet9 encoder pre-trained on CIFAR10 with layers from ResNet9 pre-trained on STL10}. We report the linear probing accuracy of ResNet9 with the replaced layers and tested on the CIFAR10, STL10 test sets.
    }
    \vspace{0.2cm}
\begin{subtable}{0.9\textwidth}
\centering
\caption{CIFAR10 \& STL10}
\begin{tabular}{cccc}
\toprule

Replaced Criterium & Replaced Layer(s)&  CIFAR10 &  STL10 \\ \midrule
        
\textit{None (Baseline)} & \textit{None} & 69.08\% $\pm$ 1.05\%   &  17.81\% $\pm$ 0.92\%  \\
Most Memorized (delta)  & 6 & 59.84\% $\pm$ 1.20\% & 21.98\% $\pm$ 0.41\%\\ 
Most Memorized (absolute) & 8 & 60.02\% $\pm$ 0.94\% &  21.67\% $\pm$ 0.72\%\\
Random    & 5 & 62.98\% $\pm$ 0.57\% & 20.44\% $\pm$ 0.85\% \\
Least Memorized (delta) & 2 & 65.52\% $\pm$ 0.74\% & 18.94\% $\pm$ 0.63\%\\ 
Least Memorized (absolute) & 3 & 64.21\% $\pm$ 1.08\% & 18.89\% $\pm$ 0.81\% \\
\bottomrule 
\end{tabular}
\label{tab:resnet9_replace_onelayer_stl10}
\end{subtable}

\begin{subtable}{0.9\textwidth}
\centering
\vspace{0.2cm}
\caption{CIFAR10 \& SVHN}
\begin{tabular}{cccccc}
\toprule
Replaced Criterium & Replaced Layer(s)&  CIFAR10 & SVHN \\ \midrule
        
None &  &  69.08\% $\pm$ 1.05\%  &  19.33\% $\pm$ 0.65\%  \\
Most Memorized (delta) & 6 & 59.22\% $\pm$ 0.97\% & 22.47\% $\pm$ 0.57\% \\  
Most Memorized (absolute) & 8 & 59.69\% $\pm$ 1.04\%  & 21.60\% $\pm$ 0.92\%\\
Random    &  5  &61.07\% $\pm$ 1.12\% & 21.09\% $\pm$ 0.69\% & \\
Least Memorized (delta) & 2  & 62.93\% $\pm$ 1.08\% & 20.44\% $\pm$ 0.71\%\\
Least Memorized (absolute) & 3  & 62.35\% $\pm$ 0.81\% & 20.18\% $\pm$ 0.98\%\\
\bottomrule 
\end{tabular}
\label{tab:resnet9_replace_onelayer_svhn}
\end{subtable}
    \label{tab:one-replacement}
\end{table*}
\addtolength{\tabcolsep}{1.8 pt}

\paragraph{Replacing blocks in ResNet50 trained on CIFAR10 with SimCLR.}
\label{appendix:replace-blocks-resnet50}

We present the results for replacing blocks in ResNet50 trained on CIFAR10 using SimCLR in \Cref{tab:resnet50_replace}.

\addtolength{\tabcolsep}{-1.8 pt}
\begin{table*}[t]
\label{tab:resnet50_replace}
        \scriptsize 
            \caption{\textbf{Evaluating the effect of replacing blocks of ResNet50 pre-trained on CIFAR10 with blocks from ResNet50 pre-trained on STL10 and SVHN}. The accuracy in the table is the linear probing accuracy of ResNet50 on CIFAR10.
    We replace block 3 of conv layers, which was selected according to the biggest $\Delta$ \layermem between two layers (not the absolute value of the \layermem score of the layers).
    }
    \vspace{0.2cm}
\begin{subtable}{0.9\textwidth}
\centering
\caption{CIFAR10 \& STL10}
\begin{tabular}{cccc}
\toprule

Replaced Criterium & Replaced Layer(s)&  CIFAR10 &  STL10 \\ \midrule
        
None & / & 77.12\% $\pm$ 1.42\%   &  18.22\% $\pm$ 0.88\%  \\
Most Memorized  & C4\_B6 C3\_B4 C2\_B3 & 43.66\% $\pm$ 1.20\%&25.78\% $\pm$ 0.95\%\\  
Random    & C3\_B2 C4\_B4 C5\_B2   &51.09\% $\pm$ 1.01\%& 22.55\% $\pm$ 1.17\%\\
Least Memorized & C2\_B1 C2\_B2 C3\_B3 & 57.41\% $\pm$ 0.74\%& 20.10\% $\pm$ 1.11\%\\ \bottomrule 
\end{tabular}
\label{tab:resnet50_replace_stl10_appendix}
\end{subtable}

\begin{subtable}{0.9\textwidth}
\centering
\vspace{0.2cm}
\caption{CIFAR10 \& SVHN}

\begin{tabular}{cccc}
\toprule

Replaced Criterium & Replaced Layer(s)&  CIFAR10 &  SVHN \\ \midrule
None &/ & 77.12\% $\pm$ 1.42\%         & 28.44\% $\pm$ 1.23\% \\
Most Memorized   &C4\_B6 C3\_B4 C2\_B3 & 35.21\% $\pm$ 0.94\%&38.11\% $\pm$ 1.08\%\\  
Random      & C3\_B2 C4\_B4 C5\_B2      &44.19\% $\pm$ 0.97\%& 32.44\% $\pm$ 1.25\%\\
Least Memorized & C2\_B1 C2\_B2 C3\_B3 &49.06\% $\pm$ 1.31\%& 29.98\% $\pm$ 0.85\%\\ \bottomrule 
\end{tabular}
\label{tab:resnet50_replace_svhn}
\end{subtable}
\end{table*}
\addtolength{\tabcolsep}{1.8 pt}

\paragraph{Statistics of Batch-Norm layer for different datasets.}
Batch-norm layers between different datasets might have different statistics. This could impact the downstream performance.
To investigate the changes, we measured the cosine similarity between the weights and biases of the batch-norm layers for two encoders (trained on CIFAR10 and STL10, respectively). The results in \Cref{tab:statistics_BN} show a high per-layer cosine similarity (average over all layers=0.823). This suggests that the statistics are similar, hence, no adjustment is required. We hypothesize that the similarity stems from the fact that the data distributions are similar and that we normalize both input datasets according to the ImageNet normalization parameters.

\begin{table}[t]
    \centering
    \small
\caption{\textbf{Cosine similarities between batch-norm layer outputs for ResNet9 trained on CIFAR10 and STL10.} We normalize the training data according to the ImageNet parameters and train the encoders using SimCLR. We calculate the cosine similarity over the weights ($\gamma$) and the bias ($\beta$) of the respective encoders' trained batch-norm layers.
}
\vspace{0.2cm}
   \begin{tabular}{ccccccccc}
\toprule
Layer  &1&2&3&4&5&6&7&8\\
\midrule
Cosine Similarity &0.797&0.844&0.823&0.811&0.779&0.805&0.847&0.881\\
\bottomrule
\end{tabular}
   \vspace{1pt}

    \label{tab:statistics_BN}
\end{table}

\subsection{\layermem with Different Distance Metrics}\label{sec:distance}

In addition to the $\ell_2$ distance, we also used 3 other distance metrics ($\ell_1$, cosine similarity, and angular distance) to evaluate the stability of \layermem. 
Our results in \Cref{tab:distance} highlight that \textbf{1)} the memorization scores are very similar, independent of the choice of the distance metric, and \textbf{2)} the most memorizing layers according to \large and \deltamem are the same over all metrics. This suggests that our findings are independent of the choice of distance metric.

\begin{table}[t]
    \centering
    \small
\caption{\textbf{\layermem (\texttt{LM}) and $\Delta$-LM under different distance metrics.}
We report for $\ell_1$, $\ell_2$ (see original submission), cosine similarity (Cos. Sim), and angular distance (Ang. Dist). The results highlight that our memorization measure is independent of the underlying metric. (ResNet9, CIFAR10, SimCLR).
}
\vspace{0.2cm}
 \begin{tabular}{ccccccccc}
    \toprule 
    & \multicolumn{2}{c}{$\ell_1$} & \multicolumn{2}{c}{$\ell_2$} & \multicolumn{2}{c}{Cos. Sim.} & \multicolumn{2}{c}{Ang. Dist.}\\
    Layer & \texttt{LM} & $\Delta$\texttt{LM}  & \texttt{LM} & $\Delta$\texttt{LM}  & \texttt{LM} & $\Delta$\texttt{LM}  & \texttt{LM} & $\Delta$\texttt{LM}  \\
    \midrule
         1 &0.099&-&0.091&-&0.104&-&0.096&-\\
         2 &0.128&0.029&0.123&0.032&0.134&0.030&0.128&0.032\\
         3 &0.159&0.031&0.154&0.031&0.163&0.029&0.160&0.032\\
         4 &0.187&0.028&0.183&0.029&0.190&0.027&0.191&0.031\\
         Res$2$ &0.192&0.005&0.185&0.002&0.193&0.003&0.193&0.002\\
         5 &0.221&0.029&0.212&0.027&0.220&0.027&0.222&0.029\\
         6 &0.256&0.035&0.246&0.034&0.256&0.036&0.259&0.037\\
         7 &0.289&0.033&0.276&0.030&0.288&0.032&0.293&0.034\\
         8 &0.325&0.036&0.308&0.032&0.321&0.033&0.328&0.035\\
         Res$6$ &0.329&0.004&0.311&0.003&0.323&0.002&0.329&0.001\\
        \bottomrule
    \end{tabular}

    \label{tab:distance}
\end{table}

\subsection{\layermem with Different Augmentation Strength}\label{sec:aug}
The results, reported in \Cref{tab:aug_layermem} highlight that stronger training augmentations reduce \layermem.

\begin{table}[t]
    \centering
    \small
    \caption{\textbf{\layermem (\texttt{LM}) and $\Delta$-LM under different augmentation sets used during training.}
We use the augmentations defined in \Cref{par:augmentations} during training and metric calculation. The results show that stronger augmentations reduce memorization.
(ResNet9, CIFAR10, SimCLR).
}
\vspace{0.2cm}
       \begin{tabular}{ccccccc}
    \toprule 
    & \multicolumn{2}{c}{weak} & \multicolumn{2}{c}{normal} & \multicolumn{2}{c}{strong} \\
    Layer & \texttt{LM} & $\Delta$\texttt{LM}  & \texttt{LM} & $\Delta$\texttt{LM}  & \texttt{LM} & $\Delta$\texttt{LM}   \\
    \midrule
         1 &0.092&-&0.091&-&0.089&-\\
         2 &0.123&0.031&0.123&0.032&0.120&0.031\\
         3 &0.154&0.031&0.154&0.031&0.150&0.030\\
         4 &0.184&0.030&0.183&0.029&0.178&0.028\\
         Res$2$ &0.187&0.003&0.185&0.002&0.181&0.003\\
         5 &0.215&0.028&0.212&0.027&0.208&0.027\\
         6 &0.249&0.034&0.246&0.034&0.241&0.033\\
         7 &0.280&0.031&0.276&0.030&0.269&0.028\\
         8 &0.313&0.033&0.308&0.032&0.300&0.031\\
         Res$6$ &0.315&0.002&0.311&0.003&0.302&0.002\\
        \bottomrule
    \end{tabular}
\label{tab:aug_layermem}
\end{table}

\subsection{\layermem with Different Initialization of Trainable parameters}

We performed an additional experiment where we trained encoders f and g independently with a different random seed (yielding f’ and g’) to study how random initialization of trainable parameters can affect the memorization of final vision encoder.The results are reported in \Cref{tab:variation}. We compared the overlap in most memorized samples between encoder f (from the paper) and f’. The results (Table 4, attached PDF) show that overlap is overall high (min. 69\% in layer 2) and increases in the later layers (max. 90\%, final layer).

\begin{table}[th]
\small
\centering
\caption{\textbf{Overlap in 100 most memorized samples according to \layermem between 2 different encoders.} We train encoders with different seeds and report the per-layer overlap in their most memorized samples.
We observe an overall high overlap, especially in the last layer. 
}
\vspace{0.2cm}
   \begin{tabular}{ccccccccc}
\toprule
Layer  &1&2&3&4&5&6&7&8\\
\midrule
Overlap \% &73&69&75&89&85&88&86&90\\
\bottomrule
\end{tabular}
\label{tab:variation}
\end{table}

\subsection{Impact of Layer Replacement on Layer Memorization}

\reb{According to the definition of SSLMem \Cref{eq:memdef}, we let the representations of a given input data point $x$ pass through the same (replaced) layer in both f and g. We show the \layermem and \deltamem scores after replacing a single layer in the ResNet9 encoder pre-trained using SimCLR on the CIFAR10 dataset in \Cref{tab:one-replacement} and \Cref{tab:replace-one-layer}. The \layermem score of the replaced layer always drops as expected since this layer does not memorize any original training data points. The decrease in LayerMem between the initial and replaced layers is smaller in the earlier layers (e.g., 1st layer) as compared to the later layers (e.g., 6th layer). This might be because, in general, the earlier layers from different models might be more similar as they are responsible for extracting general features instead of specific ones for a given dataset. The most important take-away from these experiments is that the $\Delta$LayerMem is not affected significantly and its values show the same trends after the layer replacement.
}

\subsection{Layer Replacement for Single, Two, and Three Layers at a Time}
\label{appendix:replace-1-2-3-layers}
\reb{We perform the experiment with the replacement of 1 layer in \Cref{tab:replace-one-layer}, 2 layers \Cref{tab:replace-two-layers}, and 3 layers \Cref{tab:replace-three-layers}. The following results confirm our results from \Cref{tab:resnet50_replace_stl10} in the main paper. When only a single layer is replaced, then the 6th (not the last layer) is the most important one. This layer had the highest \layermem score. Note that the replacement of the 6th layer causes the highest drop in accuracy on the original CIFAR10 dataset and the highest gain in accuracy on STL10. Next, when two layers are replaced, then layers 6th and 8th play the most important roles, where their replacement with layers from the encoder trained on STL10 causes the highest drop on CIFAR10 and the highest performance increase on STL10. This is contrary to the common intuition, which would suggest the replacement of the last two layers instead.}

\section{Additional Setup}

\subsection{Class Selectivity}
\label{appendix:class-selectivity}

We denote the class selectivity metric as \classselectivity. It was proposed by~\citep{smorcos2018selectivity} to quantify a unit's discriminability between different classes and described more in the main part of the paper in \Cref{sec:unit}. 
We derive the basic metric in more detail here.

To compute the \classselectivity metric per unit $u$, first the class-conditional mean activity is calculated for the test dataset $\bar{\mathcal{D}}$. We denote each test data point as $\bar{x}_i \in \bar{\mathcal{D}}$. We assume $M$ classes $C_{j=1}^{M}$, each with its corresponding test data points $\bar{x}_c \in C_j$, where $c={1,2,...,|C_j|}$.

We define the mean activation $\bar{\mu}$ of unit $u$ for class $C_j$ as
\begin{equation}
\bar{\mu}_u(C_j)= \frac{1}{|C_j|} \sum_{\bar{x}_c \in C_j} \text{activation}_{u}(\bar{x}_c)  \text{,}
\end{equation}
where the activation for convolutional feature maps is averaged across all elements of the feature map.
Further, for the unit $u$, we compute the maximum mean activation $\bar{\mu}_{max,u}$ across all classes $C$, where $M=|C|$, as 
\begin{equation}
    \bar{\mu}_{max,u} = \text{max}(\{ \bar{\mu}_u(\bar{x}_i)\}_{i=1}^{M})\text{.}
\end{equation}

Let $p$ be the index location of the maximum mean activation $\bar{\mu}_{u}(\bar{x}_p)$, \ie the $argmax$.
Then, we calculate the corresponding mean activity $\bar{\mu}_{-max,u}$ across all the remaining $M-1$ classes as
\begin{equation} 
    \bar{\mu}_{-max,u} = \frac{1}{M-1} \sum^{M}_{j=1, j \ne p} \bar{\mu}_{u}(C_j)\text{.}
\end{equation}
 
Finally, the class selectivity is then calculated as follows

\begin{equation}
\classselectivity (u) = \frac{\bar{\mu}_{max,u} - \bar{\mu}_{-max,u}}{\bar{\mu}_{max,u} + \bar{\mu}_{-max,u}}\text{,}
\label{eq:class-selectivity}
\end{equation} 
where $\bar{\mu}_{max,u}$ represents the highest class-conditional mean activity and $\bar{\mu}_{-max,u}$ denotes the
mean activity across all other classes (for unit $u$ and computed on the test dataset $\bar{\mathcal{D}}$). 

\subsection{Class Memorization}
\label{appendix:class-mem}

We use a similar definition as \classselectivity for \classmem, which measures how much a given unit is responsible for the memorization of a class. 
While \classselectivity is calculated on the \textit{test set}, we compute \classmem on the \textit{training dataset}.

To compute the \classmem metric per unit $u$, first, the class-conditional mean activity is calculated for the training dataset $\mathcal{D}'$. We denote each train data point as $x_i \in \mathcal{D}'$. We assume $M$ classes $C_{j=1}^{M}$, each with its corresponding train data points $x_c \in C_j$, where $c={1,2,...,|C_j|}$.

We define the mean activation $\Tilde{\mu}$ of unit $u$ for class $C_j$ as
\begin{equation}
\Tilde{\mu}_u(C_j)= \frac{1}{|C_j|} \sum_{x_c \in C_j} \text{activation}_{u}(x_c)  \text{,}
\end{equation}
where the activation for convolutional feature maps is averaged across all elements of the feature map.
Further, for the unit $u$, we compute the maximum mean activation $\Tilde{\mu}_{max,u}$ across all classes $C$, where $M=|C|$, as 
\begin{equation}
    \Tilde{\mu}_{max,u} = \text{max}(\{ \Tilde{\mu}_u(x_i)\}_{i=1}^{M})\text{.}
\end{equation}

Let $p$ be the index location of the maximum mean activation $\Tilde{\mu}_{u}(\Tilde{x}_p)$, \ie the $argmax$.
Then, we calculate the corresponding mean activity $\Tilde{\mu}_{-max,u}$ across all the remaining $M-1$ classes as
\begin{equation} 
    \Tilde{\mu}_{-max,u} = \frac{1}{M-1} \sum^{M}_{j=1, j \ne p} \Tilde{\mu}_{u}(C_j)\text{.}
\end{equation}
 
Finally, the class Memorization is then calculated as follows

\begin{equation}
\classmem (u) = \frac{\Tilde{\mu}_{max,u} - \Tilde{\mu}_{-max,u}}{\Tilde{\mu}_{max,u} + \Tilde{\mu}_{-max,u}}\text{,}
\label{eq:class-selectivity}
\end{equation} 
where $\Tilde{\mu}_{max,u}$ represents the highest class-conditional mean activity and $\Tilde{\mu}_{-max,u}$ denotes the
mean activity across all other classes (for unit $u$ and computed on the train dataset $\mathcal{D}'$). 

\section{Extended Related Work}

\paragraph{Localizing Memorization on the Level of Individual Units.}
In \Cref{sec:unit}, we considered memorization from the perspective of individual units and identified that pruning the least/most memorized units according to \unitmem preserves the least/most performance (as shown in Table \Cref{tab:pruning_strategy}). The work by \citet{maini2023can} characterized individual examples as mislabeled based on the low number of channels or filters that need to be zeroed out to flip the prediction. They observe that significantly more neurons need to be zeroed out to flip clean examples compared to mislabeled ones.

A similar experiment in the SSL domain could potentially reveal a similar trend, where noisy examples are harder to learn and primarily influence a small number of units. However, SSL encoders do not have discrete output changes from zeroing out individual units. One could pre-train the encoder and add linear probing, but this would require labels for the SSL training set, making it inapplicable. Even with labeled data and fine-tuning, identifying noisy SSL examples based on the \sslmem score may not match mislabeled examples in SL. The lack of a discrete oracle and the potential mismatch between noisy SSL and mislabeled SL examples makes it difficult to identify individual units responsible for predictions of selected examples using prior methods.

\section{Impact \& Limitations}
\label{app:limitation}
The fact that memorization can enable privacy attacks, such as data extraction~\citep{carlini2019secret,carlini2021extracting,carlini2023extracting}, has been established in prior work.
Yet, this paper advances the field of machine learning towards a novel fundamental understanding on where in SSL encoders memorization happens, and how memorization differs between standard SL models and SSL encoders.
Our insights hold the potential to yield societal benefits in the form of the design of novel methods to reduce memorization, improve fine-tuning, and yield better model pruning algorithms.

\newpage
% \clearpage
\section{Additional Results}
% \franzi{@Wenhao, here you can put the new results from the rebuttal.
% Add as LaTeX tables and figures. Please try to write the caption "camera-ready", such that we can directly use them.}

\begin{table}[H]
\centering
\tiny
\caption{\textbf{All-layer memorization.} \reb{We train the ResNet50 encoder using  SimCLR and DINO SSL frameworks on the ImageNet dataset. We report the full results with the \layermem and \deltamem scores for each layer.}}
\vspace{0.2cm}
\begin{tabular}{cccccl}
\toprule
ResNet50 Layer & \multicolumn{2}{c}{SimCLR} & \multicolumn{2}{c}{DINO}  &  \\
               & \layermem      & \deltamem & \layermem      & \deltamem &   \\
\midrule
conv1          & 0.038 $\pm$ 0.001  & -         & 0.040 $\pm$ 0.002 & -         &  \\
max pool       & 0.039 $\pm$ 0.002  & 0.001     & 0.040 $\pm$ 0.002 & 0.000     &  \\
conv2-1        & 0.041 $\pm$ 0.002  & 0.002     & 0.043 $\pm$ 0.001 & 0.003     &  \\
conv2-2        & 0.044 $\pm$ 0.002  & 0.003     & 0.045 $\pm$ 0.001 & 0.002     &  \\
conv2-3        & 0.048 $\pm$ 0.001  & 0.004     & 0.048 $\pm$ 0.002 & 0.003     &  \\
conv2-4        & 0.052 $\pm$ 0.002  & 0.004     & 0.052 $\pm$ 0.001 & 0.004     &  \\
conv2-5        & 0.055 $\pm$ 0.001  & 0.003     & 0.056 $\pm$ 0.001 & 0.004     &  \\
conv2-6        & 0.059 $\pm$ 0.002  & 0.004     & 0.060 $\pm$ 0.001 & 0.004     &  \\
conv2-7        & 0.063 $\pm$ 0.001  & 0.004     & 0.065 $\pm$ 0.001 & 0.005     &  \\
conv2-8        & 0.068 $\pm$ 0.001  & 0.005     & 0.069 $\pm$ 0.002 & 0.004     &  \\
conv2-9        & 0.072 $\pm$ 0.002  & 0.004     & 0.073 $\pm$ 0.001 & 0.004     &  \\
conv3-1        & 0.077 $\pm$ 0.002  & 0.005     & 0.078 $\pm$ 0.002 & 0.005     &  \\
conv3-2        & 0.081 $\pm$ 0.003  & 0.004     & 0.083 $\pm$ 0.001 & 0.005     &  \\
conv3-3        & 0.086 $\pm$ 0.002  & 0.005     & 0.088 $\pm$ 0.002 & 0.005     &  \\
conv3-4        & 0.092 $\pm$ 0.001  & 0.006     & 0.094 $\pm$ 0.001 & 0.006     &  \\
conv3-5        & 0.097 $\pm$ 0.001  & 0.005     & 0.099 $\pm$ 0.002 & 0.005     &  \\
conv3-6        & 0.103 $\pm$ 0.002  & 0.006     & 0.104 $\pm$ 0.001 & 0.005     &  \\
conv3-7        & 0.108 $\pm$ 0.002  & 0.005     & 0.110 $\pm$ 0.001 & 0.006     &  \\
conv3-8        & 0.112 $\pm$ 0.002  & 0.004     & 0.115 $\pm$ 0.002 & 0.005     &  \\
conv3-9        & 0.117 $\pm$ 0.001  & 0.005     & 0.120 $\pm$ 0.001 & 0.005     &  \\
conv3-10       & 0.123 $\pm$ 0.002  & 0.006     & 0.126 $\pm$ 0.002 & 0.006     &  \\
conv3-11       & 0.128 $\pm$ 0.001  & 0.005     & 0.131 $\pm$ 0.003 & 0.005     &  \\
conv3-12       & 0.134 $\pm$ 0.002  & 0.006     & 0.136 $\pm$ 0.002 & 0.005     &  \\
conv4-1        & 0.139 $\pm$ 0.002  & 0.005     & 0.142 $\pm$ 0.002 & 0.006     &  \\
conv4-2        & 0.145 $\pm$ 0.002  & 0.006     & 0.148 $\pm$ 0.003 & 0.006     &  \\
conv4-3        & 0.150 $\pm$ 0.003  & 0.005     & 0.153 $\pm$ 0.002 & 0.005     &  \\
conv4-4        & 0.156 $\pm$ 0.003  & 0.006     & 0.159 $\pm$ 0.003 & 0.006     &  \\
conv4-5        & 0.161 $\pm$ 0.002  & 0.005     & 0.164 $\pm$ 0.003 & 0.005     &  \\
conv4-6        & 0.166 $\pm$ 0.003  & 0.005     & 0.169 $\pm$ 0.004 & 0.005     &  \\
conv4-7        & 0.172 $\pm$ 0.004  & 0.006     & 0.175 $\pm$ 0.002 & 0.006     &  \\
conv4-8        & 0.178 $\pm$ 0.003  & 0.006     & 0.181 $\pm$ 0.003 & 0.006     &  \\
conv4-9        & 0.183 $\pm$ 0.002  & 0.005     & 0.186 $\pm$ 0.002 & 0.005     &  \\
conv4-10       & 0.189 $\pm$ 0.003  & 0.006     & 0.192 $\pm$ 0.003 & 0.006     &  \\
conv4-11       & 0.194 $\pm$ 0.004  & 0.005     & 0.198 $\pm$ 0.004 & 0.006     &  \\
conv4-12       & 0.200 $\pm$ 0.003  & 0.006     & 0.203 $\pm$ 0.005 & 0.005     &  \\
conv4-13       & 0.207 $\pm$ 0.006  & 0.007     & 0.210 $\pm$ 0.003 & 0.007     &  \\
conv4-14       & 0.212 $\pm$ 0.002  & 0.005     & 0.216 $\pm$ 0.004 & 0.006     &  \\
conv4-15       & 0.218 $\pm$ 0.003  & 0.006     & 0.221 $\pm$ 0.003 & 0.005     &  \\
conv4-16       & 0.224 $\pm$ 0.004  & 0.006     & 0.228 $\pm$ 0.005 & 0.007     &  \\
conv4-17       & 0.229 $\pm$ 0.003  & 0.005     & 0.234 $\pm$ 0.003 & 0.006     &  \\
conv4-18       & 0.235 $\pm$ 0.005  & 0.006     & 0.240 $\pm$ 0.003 & 0.006     &  \\
conv5-1        & 0.242 $\pm$ 0.003  & 0.007     & 0.247 $\pm$ 0.004 & 0.007     &  \\
conv5-2        & 0.249 $\pm$ 0.003  & 0.007     & 0.254 $\pm$ 0.005 & 0.007     &  \\
conv5-3        & 0.257 $\pm$ 0.002  & 0.008     & 0.261 $\pm$ 0.004 & 0.007     &  \\
conv5-4        & 0.265 $\pm$ 0.002  & 0.008     & 0.269 $\pm$ 0.005 & 0.008     &  \\
conv5-5        & 0.272 $\pm$ 0.004  & 0.007     & 0.276 $\pm$ 0.004 & 0.007     &  \\
conv5-6        & 0.279 $\pm$ 0.003  & 0.007     & 0.283 $\pm$ 0.003 & 0.007     &  \\
conv5-7        & 0.287 $\pm$ 0.003  & 0.008     & 0.292 $\pm$ 0.006 & 0.009     &  \\
conv5-8        & 0.295 $\pm$ 0.003  & 0.008     & 0.300 $\pm$ 0.004 & 0.008     &  \\
conv5-9        & 0.304 $\pm$ 0.005  & 0.009     & 0.309 $\pm$ 0.004 & 0.009     & \\
\bottomrule
\end{tabular}
\label{tab:resnet50-simclr-dino-full}
\end{table}

\begin{table}[H]
\centering
\tiny
    \caption{\reb{\textbf{Replace a single layer.} We follow the settings from the \Cref{tab:resnet50_replace_stl10} (same encoder) and replace a single layer at a time.}}
        \vspace{0.2cm}
    \begin{tabular}{ccccc}
    \toprule
        Replaced Layers & CIFAR10 & STL10 & ~ & ~ \\ \hline
        None & 69.37\%±1.07\% & 20.09\%±0.64\% & ~ & ~ \\ 
        1 & 63.07\%±0.93\% & 18.79\%±0.62\% & ~ & ~ \\ 
        2 & 65.19\%±0.87\% & 19.01\%±0.77\% & ~ & ~ \\ 
        3 & 64.47\%±1.15\% & 18.99\%±0.81\% & ~ & ~ \\ 
        4 & 60.29\%±0.74\% & 20.44\%±0.66\% & ~ & ~ \\ 
        5 & 62.74\%±0.82\% & 19.93\%±0.59\% & ~ & ~ \\ 
        6 & 59.91\%±1.09\% & 21.92\%±0.67\% & ~ & ~ \\ 
        7 & 60.77\%±0.75\% & 20.97\%±0.52\% & ~ & ~ \\ 
        8 & 60.04\%±0.90\% & 21.71\%±0.58\% & ~ & ~ \\ 
        \hline
        max(=layerX) & 65.19\%±0.87\%(2) & 21.92\%±0.67\%(6) & ~ & ~ \\ 
        min(=layerX) & 59.91\%±1.09\%(6) & 18.79\%±0.62\%(2) & ~ & ~ \\ 
        \bottomrule
    \end{tabular}
\label{tab:replace-one-layer}
\end{table}

\begin{table}[H]
\centering
\tiny
\caption{
%\reb{ViT-Base MAE, DINO full on ImageNet}
\textbf{All-layer memorization.}
\reb{We train the ViT-Base encoder using MAE and DINO SSL frameworks on the ImageNet dataset. We report the full results with the \layermem and \deltamem scores for each block.}
}
\vspace{0.2cm}
\begin{tabular}{cccccc}
\toprule
ViT-Base     & \multicolumn{2}{c}{MAE}                 & \multicolumn{2}{c}{DINO}       &           \\
Block Number & \layermem      & \deltamem & \layermem      & \deltamem &  \\
\midrule
1            & 0.019$\pm$0.001   & -         & 0.019$\pm$0.001   & -         &           \\
2            & 0.037$\pm$0.001   & 0.011     & 0.036$\pm$0.002   & 0.012     &           \\
3            & 0.055$\pm$0.002   & 0.013     & 0.056$\pm$0.003   & 0.012     &           \\
4            & 0.075$\pm$0.002   & 0.013     & 0.077$\pm$0.002   & 0.014     &           \\
5            & 0.095$\pm$0.002   & 0.012     & 0.096$\pm$0.004   & 0.013     &           \\
6            & 0.118$\pm$0.004   & 0.016     & 0.119$\pm$0.006   & 0.015     &           \\
7            & 0.139$\pm$0.003   & 0.015     & 0.142$\pm$0.004   & 0.014     &           \\
8            & 0.163$\pm$0.005   & 0.018     & 0.168$\pm$0.005   & 0.017     &           \\
9            & 0.188$\pm$0.004   & 0.017     & 0.193$\pm$0.003   & 0.018     &           \\
10           & 0.215$\pm$0.006   & 0.018     & 0.219$\pm$0.006   & 0.019     &           \\
11           & 0.243$\pm$0.005   & 0.021     & 0.247$\pm$0.005   & 0.020     &           \\
12           & 0.271$\pm$0.003   & 0.020     & 0.275$\pm$0.004   & 0.019     &           \\
\bottomrule
\end{tabular}
\label{tab:vit-base-mae-dino-full}
\end{table}

\begin{table}[H]
\centering
\tiny
    \caption{\reb{\textbf{Replace two layers.} We follow the setting from the \Cref{tab:resnet50_replace_stl10} (same encoder) and replace two layers at a time.}}
        \vspace{0.2cm}
    \begin{tabular}{ccccc}
    \toprule
        Replaced Layers & CIFAR10 & STL10 & ~ & ~ \\ \hline
        None & 69.37\% ± 1.07\% & 18.44\% ± 0.64\% \\ 
        1 2 & 53.44\% ± 0.90\% & 20.31\% ± 0.51\% \\ 
        1 3 & 52.51\% ± 0.83\% & 20.80\% ± 0.60\% \\ 
        1 4 & 50.77\% ± 0.78\% & 22.16\% ± 0.66\% \\ 
        1 5 & 50.98\% ± 0.97\% & 22.02\% ± 0.59\% \\ 
        1 6 & 46.09\% ± 0.77\% & 21.46\% ± 0.69\% \\ 
        1 7 & 49.59\% ± 0.89\% & 24.87\% ± 0.66\% \\ 
        1 8 & 45.96\% ± 0.94\% & 21.66\% ± 0.71\% \\ 
        2 3 & 55.44\% ± 0.73\% & 19.31\% ± 0.72\% \\ 
        2 4 & 53.61\% ± 0.92\% & 24.18\% ± 0.68\% \\ 
        2 5 & 53.34\% ± 1.06\% & 20.39\% ± 0.57\% \\ 
        2 6 & 48.59\% ± 0.81\% & 23.32\% ± 0.70\% \\ 
        2 7 & 51.07\% ± 1.13\% & 21.97\% ± 0.52\% \\ 
        2 8 & 50.15\% ± 0.82\% & 22.57\% ± 0.61\% \\ 
        3 4 & 52.99\% ± 1.01\% & 21.09\% ± 0.73\% \\ 
        3 5 & 52.67\% ± 0.90\% & 21.00\% ± 0.81\% \\ 
        3 6 & 48.22\% ± 0.79\% & 23.48\% ± 0.62\% \\ 
        3 7 & 50.81\% ± 0.86\% & 22.09\% ± 0.70\% \\ 
        3 8 & 49.07\% ± 0.92\% & 23.19\% ± 0.67\% \\ 
        4 5 & 50.49\% ± 0.96\% & 22.41\% ± 0.55\% \\ 
        4 6 & 44.88\% ± 0.91\% & 23.61\% ± 0.49\% \\ 
        4 7 & 46.38\% ± 1.13\% & 24.04\% ± 0.62\% \\ 
        4 8 & 45.09\% ± 0.75\% & 24.11\% ± 0.66\% \\ 
        5 6 & 46.02\% ± 1.07\% & 24.29\% ± 0.81\% \\ 
        5 7 & 49.21\% ± 1.00\% & 22.99\% ± 0.80\% \\ 
        5 8 & 45.71\% ± 0.94\% & 24.33\% ± 0.73\% \\ 
        6 7 & 44.76\% ± 0.88\% & 24.90\% ± 0.54\% \\ 
        6 8 & 44.13\% ± 1.01\% & 25.08\% ± 0.64\% \\ 
        7 8 & 44.98\% ± 0.94\% & 24.91\% ± 0.81\% \\ 
        \hline
        max (=layerX) & 55.44\% ± 0.73\% (2 3) & 25.08\% ± 0.64\% (6 8) \\ 
        min (=layerX) & 44.13\% ± 1.01\% (6 8) & 19.31\% ± 0.72\% (2 3) \\ 
                \bottomrule
    \end{tabular}
\label{tab:replace-two-layers}
\end{table}

\begin{table}[H]
\centering
\tiny
    \caption{\reb{\textbf{Replace three layers.} We follow the settings from the \Cref{tab:resnet50_replace_stl10} (same encoder) and replace three layers at a time.}}
        \vspace{0.2cm}
    \begin{tabular}{ccccc}
    \toprule
        Replaced Layers & CIFAR10 & STL10 & ~ & ~ \\ \hline
             None & 69.37\% ± 1.07\% & 18.44\% ± 0.64\% \\ 
        1 2 3 & 49.77\% ± 0.66\% & 21.95\% ± 0.42\% \\ 
        1 2 4 & 48.33\% ± 0.65\% & 23.28\% ± 0.77\% \\ 
        1 2 5 & 49.51\% ± 0.70\% & 22.18\% ± 0.57\% \\ 
        1 2 6 & 45.31\% ± 0.59\% & 26.41\% ± 0.53\% \\ 
        1 2 7 & 48.99\% ± 0.84\% & 22.54\% ± 0.57\% \\ 
        1 2 8 & 46.29\% ± 0.48\% & 25.74\% ± 0.62\% \\ 
        1 3 4 & 47.66\% ± 0.52\% & 24.37\% ± 0.42\% \\ 
        1 3 5 & 49.04\% ± 0.65\% & 22.20\% ± 0.66\% \\ 
        1 3 6 & 45.19\% ± 0.72\% & 26.57\% ± 0.61\% \\ 
        1 3 7 & 48.19\% ± 0.58\% & 23.76\% ± 0.82\% \\ 
        1 3 8 & 46.20\% ± 0.83\% & 25.67\% ± 0.60\% \\ 
        1 4 5 & 47.35\% ± 0.60\% & 24.99\% ± 0.58\% \\ 
        1 4 6 & 43.89\% ± 0.70\% & 29.57\% ± 0.67\% \\ 
        1 4 7 & 44.53\% ± 0.66\% & 27.55\% ± 0.71\% \\ 
        1 4 8 & 43.94\% ± 0.78\% & 29.26\% ± 0.52\% \\ 
        1 5 6 & 44.23\% ± 0.70\% & 27.89\% ± 0.52\% \\ 
        1 5 7 & 48.03\% ± 0.55\% & 23.30\% ± 0.63\% \\ 
        1 5 8 & 44.71\% ± 0.71\% & 26.99\% ± 0.58\% \\ 
        1 6 7 & 43.30\% ± 0.44\% & 29.81\% ± 0.57\% \\ 
        1 6 8 & 41.72\% ± 0.70\% & 30.71\% ± 0.67\% \\ 
        1 7 8 & 44.59\% ± 0.83\% & 28.06\% ± 0.47\% \\ 
        2 3 4 & 48.89\% ± 0.38\% & 22.76\% ± 0.38\% \\ 
        2 3 5 & 50.48\% ± 0.67\% & 21.71\% ± 0.60\% \\ 
        2 3 6 & 46.98\% ± 0.57\% & 25.68\% ± 0.46\% \\ 
        2 3 7 & 49.81\% ± 0.62\% & 22.31\% ± 0.49\% \\ 
        2 3 8 & 48.07\% ± 0.93\% & 23.74\% ± 0.70\% \\ 
        2 4 5 & 48.55\% ± 0.79\% & 23.90\% ± 0.82\% \\ 
        2 4 6 & 44.99\% ± 0.58\% & 27.88\% ± 0.57\% \\ 
        2 4 7 & 47.78\% ± 0.68\% & 24.87\% ± 0.75\% \\ 
        2 4 8 & 45.41\% ± 0.86\% & 26.98\% ± 0.51\% \\ 
        2 5 6 & 45.91\% ± 0.44\% & 26.47\% ± 0.60\% \\ 
        2 5 7 & 48.37\% ± 0.55\% & 22.90\% ± 0.50\% \\ 
        2 5 8 & 47.18\% ± 0.52\% & 25.57\% ± 0.81\% \\ 
        2 6 7 & 45.62\% ± 0.69\% & 26.78\% ± 0.48\% \\ 
        2 6 8 & 42.99\% ± 0.63\% & 29.41\% ± 0.43\% \\ 
        2 7 8 & 46.89\% ± 0.93\% & 25.77\% ± 0.63\% \\ 
        3 4 5 & 47.90\% ± 0.56\% & 24.74\% ± 0.48\% \\ 
        3 4 6 & 43.32\% ± 0.58\% & 28.73\% ± 0.60\% \\ 
        3 4 7 & 45.80\% ± 0.57\% & 26.59\% ± 0.65\% \\ 
        3 4 8 & 44.49\% ± 0.55\% & 28.36\% ± 0.46\% \\ 
        3 5 6 & 45.49\% ± 0.71\% & 26.89\% ± 0.90\% \\ 
        3 5 7 & 48.14\% ± 0.73\% & 23.66\% ± 0.58\% \\ 
        3 5 8 & 46.61\% ± 0.69\% & 25.90\% ± 0.39\% \\ 
        3 6 7 & 44.01\% ± 0.72\% & 28.20\% ± 0.58\% \\ 
        3 6 8 & 42.00\% ± 0.65\% & 29.93\% ± 0.33\% \\ 
        3 7 8 & 45.21\% ± 0.43\% & 27.26\% ± 0.41\% \\ 
        4 5 6 & 41.84\% ± 0.76\% & 30.20\% ± 0.53\% \\ 
        4 5 7 & 44.20\% ± 0.72\% & 27.64\% ± 0.48\% \\ 
        4 5 8 & 42.31\% ± 0.82\% & 29.81\% ± 0.33\% \\ 
        4 6 7 & 40.02\% ± 0.71\% & 32.55\% ± 0.58\% \\ 
        4 6 8 & 37.66\% ± 0.49\% & 31.94\% ± 0.38\% \\ 
        4 7 8 & 40.96\% ± 0.62\% & 30.78\% ± 0.56\% \\ 
        5 6 7 & 42.77\% ± 0.60\% & 29.66\% ± 0.69\% \\ 
        5 6 8 & 40.55\% ± 0.68\% & 31.02\% ± 0.47\% \\ 
        5 7 8 & 43.79\% ± 0.91\% & 28.54\% ± 0.55\% \\ 
        6 7 8 & 38.95\% ± 0.57\% & 31.59\% ± 0.66\% \\ 
        max (=layerX) & 50.48\% ± 0.67\% (2 3 5) & 31.94\% ± 0.38\% (4 6 8) \\ 
        min (=layerX) & 37.66\% ± 0.49\% (4 6 8) & 21.71\% ± 0.60\% (2 3 5) \\ 
                        \bottomrule
    \end{tabular}
\label{tab:replace-three-layers}
\end{table}

\begin{table}[H]
    \centering
    \tiny
    \caption{\reb{
    \textbf{\unitmem distinguishes between individual examples within a class.}
    We use 1000 samples for each experiment to compute the UnitMem score. \textbf{All}: denotes all classes, \textbf{TPC}: stands for the 3 following classes Truck, Plance, and Car classes, while \textbf{Car}: is simply the car class.
    }}
            \vspace{0.2cm}
    \setlength{\tabcolsep}{5pt}
    \begin{tabular}{ccccccccccc}
    \toprule
        \textbf{Layer} & \textbf{All} & \textbf{All} & \textbf{All} & \textbf{TPC} & \textbf{TPC} & \textbf{TPC} & \textbf{Car} & \textbf{Car} & \textbf{Car} \\ 
        \textbf{Number} & \textbf{min} & \textbf{max} & \textbf{avg} & \textbf{min} & \textbf{max} & \textbf{avg} & \textbf{min} & \textbf{max} & \textbf{avg} \\
        \hline
        \textbf{Layer 1} & 0±0 & 0.845±0.014 & 0.366±0.011 & 0.007±1e-4 & 0.801±0.018 & 0.357±0.009 & 0.011±1e-4 & 0.829±0.015 & 0.360±0.010 \\ 
        \textbf{Layer 2} & 0.006±9e-5 & 0.832±0.016 & 0.352±0.010 & 0±0 & 0.789±0.015 & 0.350±0.013 & 0.009±8e-5 & 0.810±0.011 & 0.351±0.013 \\ 
        \textbf{Layer 3} & 0±0 & 0.841±0.017 & 0.363±0.008 & 0±0 & 0.800±0.014 & 0.355±0.010 & 0.010±9e-5 & 0.825±0.009 & 0.356±0.008 \\ 
        \textbf{Layer 4} & 0±0 & 0.871±0.012 & 0.377±0.009 & 0.004±1e-4 & 0.833±0.019 & 0.373±0.012 & 0.015±2e-4 & 0.844±0.014 & 0.371±0.009 \\ 
        \textbf{Layer 5} & 0.010±2e-4 & 0.859±0.016 & 0.381±0.008 & 0.016±3e-4 & 0.810±0.013 & 0.375±0.011 & 0.013±1e-4 & 0.837±0.012 & 0.380±0.010 \\ 
        \textbf{Layer 6} & 0.020±4e-4 & 0.905±0.018 & 0.403±0.011 & 0.022±3e-4 & 0.868±0.019 & 0.381±0.008 & 0.030±5e-4 & 0.879±0.014 & 0.394±0.007 \\ 
        \textbf{Layer 7} & 0.019±3e-4 & 0.894±0.013 & 0.398±0.009 & 0.021±3e-4 & 0.859±0.014 & 0.380±0.013 & 0.019±3e-4 & 0.861±0.017 & 0.387±0.011 \\ 
        \textbf{Layer 8} & 0.017±2e-4 & 0.905±0.013 & 0.409±0.010 & 0.25±4e-4 & 0.863±0.017 & 0.385±0.010 & 0.024±4e-4 & 0.870±0.013 & 0.397±0.012 \\ 
        \bottomrule
    \end{tabular}
    \label{tab:unit_mem_classes_unique}
\end{table}

\newpage
\newpage
\section*{NeurIPS Paper Checklist}

\begin{enumerate}

\item {\bf Claims}
    \item[] Question: Do the main claims made in the abstract and introduction accurately reflect the paper's contributions and scope?
    \item[] Answer: \answerYes{} % Replace by \answerYes{}, \answerNo{}, or \answerNA{}.
    \item[] Justification: We introduce our main contributions and key findings from line 5 to line 19 in the abstract and line 73 to line 80 in \Cref{sec:intro}

\item {\bf Limitations}
    \item[] Question: Does the paper discuss the limitations of the work performed by the authors?
    \item[] Answer: \answerYes{} % Replace by \answerYes{}, \answerNo{}, or \answerNA{}.
    \item[] Justification: We discuss our Limitations in \Cref{app:limitation}

\item {\bf Theory Assumptions and Proofs}
    \item[] Question: For each theoretical result, does the paper provide the full set of assumptions and a complete (and correct) proof?
    \item[] Answer: \answerYes{} % Replace by \answerYes{}, \answerNo{}, or \answerNA{}.
    \item[] Justification: All related results are clearly stated and referenced in either the main paper or the appendix.

    \item {\bf Experimental Result Reproducibility}
    \item[] Question: Does the paper fully disclose all the information needed to reproduce the main experimental results of the paper to the extent that it affects the main claims and/or conclusions of the paper (regardless of whether the code and data are provided or not)?
    \item[] Answer: \answerYes{} % Replace by \answerYes{}, \answerNo{}, or \answerNA{}.
    \item[] Justification: The detailed experimental setup is introduced in \Cref{apendix:experimental-setup} and related source code is uploaded to open-review.
    \item[] Guidelines:
    \begin{itemize}
        \item The answer NA means that the paper does not include experiments.
        \item If the paper includes experiments, a No answer to this question will not be perceived well by the reviewers: Making the paper reproducible is important, regardless of whether the code and data are provided or not.
        \item If the contribution is a dataset and/or model, the authors should describe the steps taken to make their results reproducible or verifiable. 
        %\item Depending on the contribution, reproducibility can be accomplished in various ways. For example, if the contribution is a novel architecture, describing the architecture fully might suffice, or if the contribution is a specific model and empirical evaluation, it may be necessary to either make it possible for others to replicate the model with the same dataset, or provide access to the model. In general. releasing code and data is often one good way to accomplish this, but reproducibility can also be provided via detailed instructions for how to replicate the results, access to a hosted model (e.g., in the case of a large language model), releasing of a model checkpoint, or other means that are appropriate to the research performed.
        %\item While NeurIPS does not require releasing code, the conference does require all submissions to provide some reasonable avenue for reproducibility, which may depend on the nature of the contribution. For example
        \begin{enumerate}
            \item If the contribution is primarily a new algorithm, the paper should make it clear how to reproduce that algorithm.
            \item If the contribution is primarily a new model architecture, the paper should describe the architecture clearly and fully.
            \item If the contribution is a new model (e.g., a large language model), then there should either be a way to access this model for reproducing the results or a way to reproduce the model (e.g., with an open-source dataset or instructions for how to construct the dataset).
            \item We recognize that reproducibility may be tricky in some cases, in which case authors are welcome to describe the particular way they provide for reproducibility. In the case of closed-source models, it may be that access to the model is limited in some way (e.g., to registered users), but it should be possible for other researchers to have some path to reproducing or verifying the results.
        \end{enumerate}
    \end{itemize}

\item {\bf Open access to data and code}
    \item[] Question: Does the paper provide open access to the data and code, with sufficient instructions to faithfully reproduce the main experimental results, as described in supplemental material?
    \item[] Answer: \answerYes{} % Replace by \answerYes{}, \answerNo{}, or \answerNA{}.
    \item[] Justification: All related source code is uploaded to open-review and experiments are conducted on open source datasets.

\item {\bf Experimental Setting/Details}
    \item[] Question: Does the paper specify all the training and test details (e.g., data splits, hyperparameters, how they were chosen, type of optimizer, etc.) necessary to understand the results?
    \item[] Answer: \answerYes{} % Replace by \answerYes{}, \answerNo{}, or \answerNA{}.
    \item[] Justification: The detailed experimental setup is introduced in \Cref{apendix:experimental-setup}.

\item {\bf Experiment Statistical Significance}
    \item[] Question: Does the paper report error bars suitably and correctly defined or other appropriate information about the statistical significance of the experiments?
    \item[] Answer: \answerYes{} % Replace by \answerYes{}, \answerNo{}, or \answerNA{}.
    \item[] Justification: For the average results of all multiple experiments in the paper, we report the standard deviation in the tables and draw the error bar used to represent the standard deviation in the figures.
    \item[] Guidelines:
    \begin{itemize}
        \item The answer NA means that the paper does not include experiments.
        \item The authors should answer "Yes" if the results are accompanied by error bars, confidence intervals, or statistical significance tests, at least for the experiments that support the main claims of the paper.
        \item The factors of variability that the error bars are capturing should be clearly stated (for example, train/test split, initialization, random drawing of some parameter, or overall run with given experimental conditions).
        %\item The method for calculating the error bars should be explained (closed form formula, call to a library function, bootstrap, etc.)
        \item The assumptions made should be given (e.g., Normally distributed errors).
        \item It should be clear whether the error bar is the standard deviation or the standard error of the mean.
        \item It is OK to report 1-sigma error bars, but one should state it. The authors should preferably report a 2-sigma error bar than state that they have a 96\% CI, if the hypothesis of Normality of errors is not verified.
        \item For asymmetric distributions, the authors should be careful not to show in tables or figures symmetric error bars that would yield results that are out of range (e.g. negative error rates).
        \item If error bars are reported in tables or plots, The authors should explain in the text how they were calculated and reference the corresponding figures or tables in the text.
    \end{itemize}

\item {\bf Experiments Compute Resources}
    \item[] Question: For each experiment, does the paper provide sufficient information on the computer resources (type of compute workers, memory, time of execution) needed to reproduce the experiments?
    \item[] Answer: \answerYes{} % Replace by \answerYes{}, \answerNo{}, or \answerNA{}.
    \item[] Justification: The hardware usage is introduced in \Cref{app:hardware}
    
\item {\bf Code Of Ethics}
    \item[] Question: Does the research conducted in the paper conform, in every respect, with the NeurIPS Code of Ethics \url{https://neurips.cc/public/EthicsGuidelines}?
    \item[] Answer: \answerYes{} % Replace by \answerYes{}, \answerNo{}, or \answerNA{}.
    %\item[] Justification: \justificationTODO{}

\item {\bf Broader Impacts}
    \item[] Question: Does the paper discuss both potential positive societal impacts and negative societal impacts of the work performed?
    \item[] Answer: \answerNA{} % Replace by \answerYes{}, \answerNo{}, or \answerNA{}.
    \item[] Justification: In this work, we introduce two metrics for locating memorization in SSL vision encoders, as well as some key findings  that result from experimenting with our metrics. No potential direct social impact is expected. 
    
\item {\bf Safeguards}
    \item[] Question: Does the paper describe safeguards that have been put in place for responsible release of data or models that have a high risk for misuse (e.g., pretrained language models, image generators, or scraped datasets)?
    \item[] Answer: \answerNA{} % Replace by \answerYes{}, \answerNo{}, or \answerNA{}.
    \item[] Justification: All datasets used to train models for this paper are safe public datasets, including ImageNet ILSVRC-2012~\citep{russakovsky2015imagenet}, CIFAR10~\citep{krizhevsky2009learning}, CIFAR100~\citep{krizhevsky2009learning}, SVHN~\citep{netzer2011reading}, and STL10~\citep{coates2011analysis} (This is also introduced in \Cref{apendix:experimental-setup}).  

\item {\bf Licenses for existing assets}
    \item[] Question: Are the creators or original owners of assets (e.g., code, data, models), used in the paper, properly credited and are the license and terms of use explicitly mentioned and properly respected?
    \item[] Answer: \answerYes{} % Replace by \answerYes{}, \answerNo{}, or \answerNA{}.
    \item[] Justification: All models and datasets usage is detailed introduced in \Cref{apendix:experimental-setup}. We totally abbey the terms of use of these public sources. All other code works are done by ourselves with no potential risks of licenses. 

\item {\bf New Assets}
    \item[] Question: Are new assets introduced in the paper well documented and is the documentation provided alongside the assets?
    \item[] Answer: \answerYes{} % Replace by \answerYes{}, \answerNo{}, or \answerNA{}.
    \item[] Justification: We provide all our code for metrics mentioned in our paper to open review and write a detailed Readme file for the usage of our code. 

\item {\bf Crowdsourcing and Research with Human Subjects}
    \item[] Question: For crowdsourcing experiments and research with human subjects, does the paper include the full text of instructions given to participants and screenshots, if applicable, as well as details about compensation (if any)? 
    \item[] Answer: \answerNA{} % Replace by \answerYes{}, \answerNo{}, or \answerNA{}.
    \item[] Justification: 

\item {\bf Institutional Review Board (IRB) Approvals or Equivalent for Research with Human Subjects}
    \item[] Question: Does the paper describe potential risks incurred by study participants, whether such risks were disclosed to the subjects, and whether Institutional Review Board (IRB) approvals (or an equivalent approval/review based on the requirements of your country or institution) were obtained?
    \item[] Answer: \answerNA{} % Replace by \answerYes{}, \answerNo{}, or \answerNA{}.
    %\item[] Justification: \justificationTODO{}

\end{enumerate}

\end{document}